\documentclass[11pt]{style/prism-princeton}

\usepackage{marvosym}

\makeatletter
\newcommand{\longdash}[1][2em]{%
  \makebox[#1]{$\m@th\smash-\mkern-7mu\cleaders\hbox{$\mkern-2mu\smash-\mkern-2mu$}\hfill\mkern-7mu\smash-$}}
\makeatother
\newcommand{\omitskip}{\kern-\arraycolsep}

\author[\heartsuit]{Kevin Qinghong Lin}
\author[\spadesuit]{Batu El}
\author[\heartsuit]{Yuhong Shi}
\author[\spadesuit]{Pan Lu}
\author[\diamondsuit]{Juil Sock}
\author[\spadesuit]{Djordje Padejski}
\author[\heartsuit]{Philip Torr\textsuperscript{\Letter}}
\author[\spadesuit]{James Zou\textsuperscript{\Letter}}

\affiliation[\heartsuit]{University of Oxford}
\affiliation[\spadesuit]{Stanford University}
\affiliation[\diamondsuit]{BBC R\&D}

\usepackage{wrapfig}
\usepackage[most]{tcolorbox}
\usepackage{xcolor}
\usepackage{arydshln, subcaption}
\usepackage{epigraph}
\usepackage{fontawesome5, wasysym, textcomp}

\usepackage[most]{tcolorbox}
\definecolor{citecolor}{HTML}{0071bc}
\hypersetup{
    citecolor=citecolor,
}

\definecolor{polaris-bg-elevated}{HTML}{f5f5f5}
\definecolor{polaris-border-subtle}{HTML}{e5e5e5}

\definecolor{filter-bg}{HTML}{ebebeb}
\definecolor{filter-border}{HTML}{c8c8c8}

\begin{document}
\newcommand{\method}{Data Journalist Agent}
\newcommand{\methodshort}{Data2Story}
\newcommand{\eg}{\textit{e.g.,}}
\newcommand{\ie}{\textit{i.e.,}}

\definecolor{googlered}{HTML}{EA4335}
\newcommand{\anno}[1]{\textcolor{googlered}{#1}}
\definecolor{okgreen}{HTML}{2E8B57}
\definecolor{nored}{HTML}{C0504D}
\definecolor{midgray}{HTML}{9A9A9A}
\definecolor{lightgreen}{RGB}{240, 251, 237}
\definecolor{grouprow}{HTML}{EDEDED}
\definecolor{ourrow}{HTML}{FFF4D6}
\newcommand{\yes}{\textcolor{okgreen}{\ding{51}}}
\newcommand{\no}{\textcolor{nored}{\ding{55}}}
\newcommand{\yespart}{\textcolor{midgray}{\ding{51}}}

\abstract{
Data tells stories that shape society, and the data journalist's job is to turn raw information into a story that non-expert audiences can understand and trust through to the end.
A high-quality news feature routinely takes a newsroom team weeks, including hunting for context, running statistics, choosing an angle, and designing visuals.
Recent agents are capable at individual steps: automated data-science agents close the analysis loop, while design agents can synthesize beautiful websites.
\emph{But can an agent serve as an assistant to data journalists?}
We introduce \textbf{\method\ (\methodshort)}, a multi-agent framework that orchestrates specialised roles into a single virtual newsroom. \methodshort\ highlights two innovations over prior approaches.
\textbf{(i) Claims are evidence-grounded and verifiable.} We introduce an “Inspector”, which links the intermediate results produced by individual roles to their sources so that the numbers, angles, and assets are grounded in data, code, or a reference (\eg~an external URL).
\textbf{(ii) Articles are multimodally generative.}
Rather than defaulting to plain text and static charts, \methodshort\ reasons about what its readers will want to read visually, then deploys multimodal tools so that the article fits both the data topic and the intended audience (\eg~an interactive map with zoom for a geography piece, or an audio clip for a music piece), making the result readable and engaging.
We evaluate \method\ on 18 articles from diverse topics and publication sources, each paired with the originally published expert-written piece, along four axes:
\textbf{(a) Human--agent angle coverage}, measuring the overlap and complementarity of angles between \methodshort\ and human-authored articles, to characterize what each side covers;
\textbf{(b) Rubric evaluation with a human study} across 53 human participants, with the rubric covering visual design, narrative pacing, data transparency, claim-data alignment, and insight value;
\textbf{(c) Computer-use agents as judge}: as an automatic cost-saving proxy for how real-world users navigate and interact with the article, we employ computer-use agents that fully perceive the interface through actions such as clicking and scrolling;
and lastly, \textbf{(d) Verifiability}, where a coding verifier re-executes every statement against the data and checks that the claims are verifiable or can be grounded in a reference.
Our central finding is that \methodshort{} produces competitive and evidence-traceable multimedia stories, with particularly strong performance on transparency and auditability dimensions.
However, human-authored articles retain a clear edge in editorial angle, creative design, and informative presentation.
 \methodshort{} is not intended to replace journalists. Rather, it serves as a solution to support story development, enabling reporting that is more evidence-based, transparent, and verifiable. 
}

\title{
\raisebox{-0.25\height}{\includegraphics[height=1.3em]{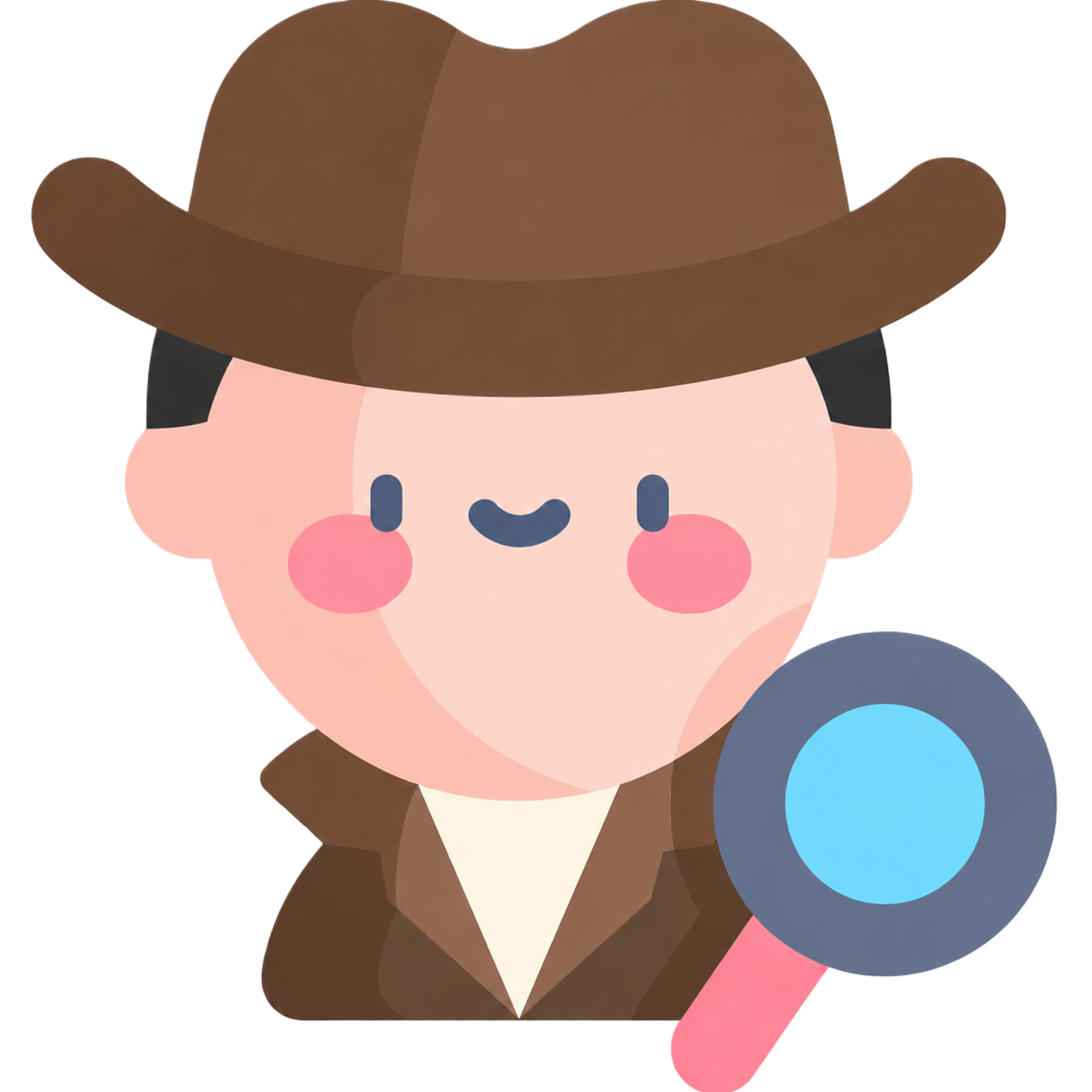}}
\method: Transforming Data into Verifiable Multimodal Stories
}

\website{https://data2story.github.io}{https://data2story.github.io}
\code{https://github.com/QinghongLin/data2story-skill}{https://github.com/QinghongLin/data2story-skill}
\correspondence{\email{philip.torr@eng.ox.ac.uk},\quad \email{jamesz@stanford.edu}}

\maketitle

\begin{figure}[!t]
    \centering
    \includegraphics[width=\columnwidth]{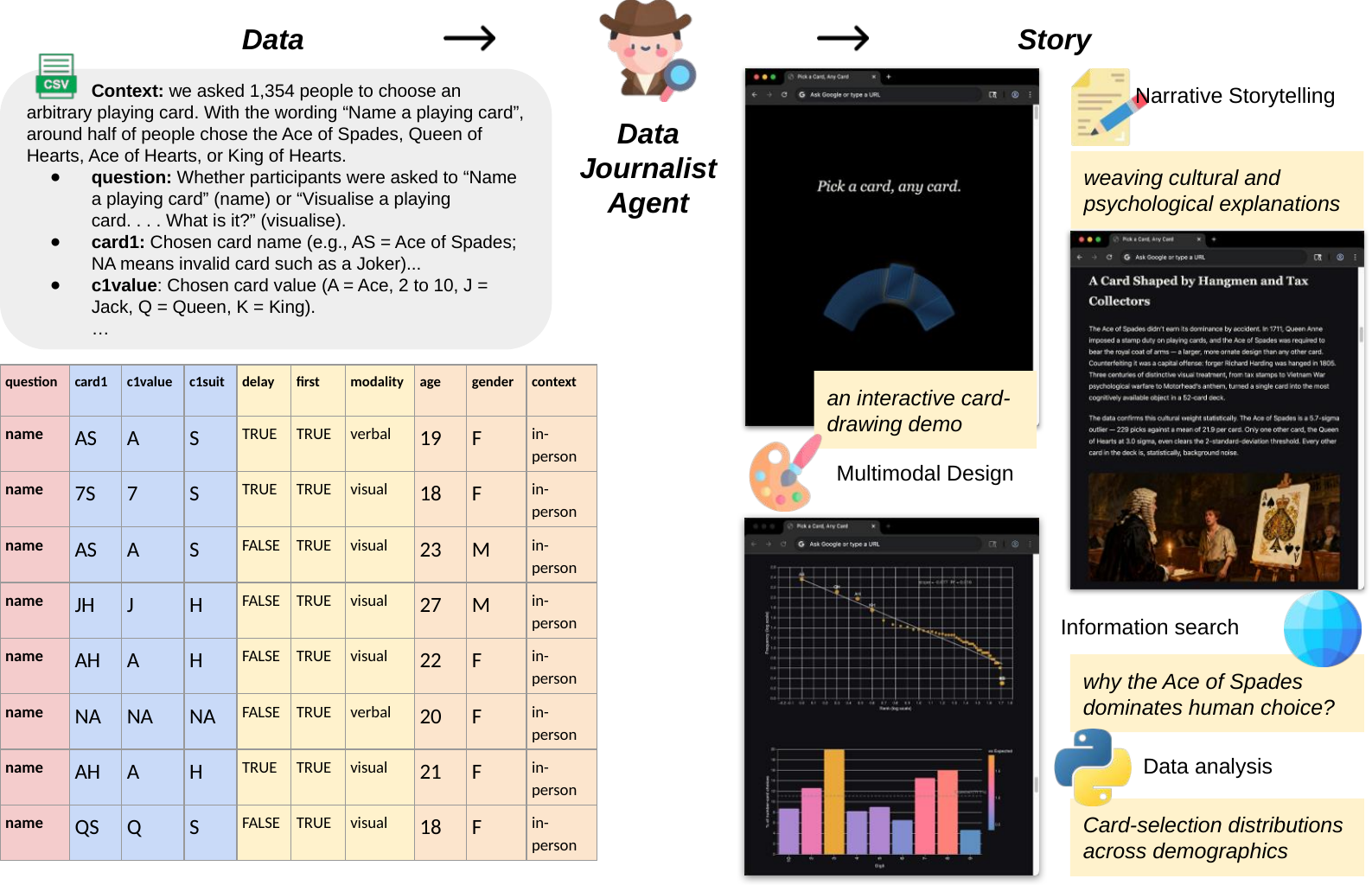}
\caption{
\textbf{\methodshort~turns a raw dataset (\eg~a CSV) into a verifiable, multimodal article (\ie~a website).} 
We use a \href{https://osf.io/534g2/overview}{“Pick a card” dataset} as an illustration.
This transformation involves information seeking (\eg~``why the Ace of Spades dominates human choice?''), data analysis via programming (\eg~computing card-selection distributions across demographics), narrative storytelling (\eg~weaving cultural and psychological explanations into a cohesive article), and multimodal design (\eg~an interactive card-drawing demo).
}
\vspace{-1em}
\label{fig:teaser}
\end{figure}

\section{Introduction}
\label{sec:introduction}
Data journalists turn raw data into stories like {``How has the way pop singers use their voice changed across generations?''} that everyday readers can follow, helping the public understand what lies behind the data -- yet a small newsroom team can spend weeks on a single high-quality article.
Recent agents are individually capable at each of these steps: automated data-science agents~\citep{dsbench,scienceagentbench,mlebench,mlagentbench} can profile a dataset, run the right statistics, and return defensible results with reproducible code.
Visualization agents~\cite{matplotagent,lida,coda,design2code} generate visual artifacts (such as websites) from a language instruction. 
\textit{But can an agent serve as an assistant to data journalists? taking raw data all the way to a story readers actually want to finish and can trust?}

However, building such an end to end agentic journalist system is non-trivial. 
Behind each finished article is a long process: gathering background, running careful statistics, choosing an angle, designing assets, building an appealing page, and several rounds of editing. 
The task is fundamentally \textbf{multi-disciplinary}, 
demanding the simultaneous exercise of multiple skills that rarely co-exist in a single contributor, which is why news is typically the product of a coordinated newsroom team. 

Companies such as \href{https://citizenportal.ai/home}{CitizenPortal} and \href{https://www.locunity.com/#preview}{Locunity} are already deploying AI agents to produce news articles at scale, signalling that AI-enabled journalism is no longer hypothetical.
However, a critical challenge shared by these systems is the lack of \textbf{verification and traceability} (as highlighted by the recent discussion~\cite{rusch2025aicitycouncil}): 
readers and editors have no reliable way to confirm where a number came from, whether a chart accurately reflects the underlying data, or whether a claim was inferred or hallucinated.
This is a particularly demanding requirement for language agents, which are prone to hallucination~\cite{ji2023survey}.
\methodshort\ directly addresses this gap: nearly every statistic, visual asset, and factual claim is grounded in executable code or a verifiable source URL, making the full reasoning chain auditable end to end.



Motivated by this, we introduce \textbf{\method\ (\methodshort)}, a multi-agent framework that orchestrates seven specialised roles into a virtual newsroom: a Detective for context hunting, an Analyst for running statistics, an Editor for narrative framing, a Designer for visual assets, a Programmer for website creation, an Auditor for reviewing the Programmer's output and offering suggestions for revision, and, most notably, an “Inspector” that traces elements of the final article back to its upstream evidence.
As illustrated in Figure~\ref{fig:teaser}, \method\ takes any data source as input and emits a generative multimedia article. Its key contributions are as follows:
\textbf{(i) Claims are evidence-grounded.} To ensure the output is grounded in verifiable evidence, we introduce a dedicated agent that links most elements of the published article (\ie~numbers, quotes, and visual assets) back to their provenance (\ie~a specific line of code, a data source, or an external URL). This makes the resulting article \textit{verifiable and auditable}.
\textbf{(ii) Articles are multimodally generative.} Rather than formatting articles as plain text or static documents, we argue that an article should be multimedia-rich (e.g., interactive charts, images, video, and audio).
We let a Designer reason about the topic and what readers will want to see and interact with.
For example, as shown in Figure~\ref{fig:teaser}, for an article on card-game outcome statistics, we add a playable starter so that readers can interact with this game directly.

To validate the effectiveness of \method, we first showcase \methodshort\ on the newest datasets that few humans have yet written up (\eg~the 2026 World Cup schedule), where it discovers original findings of its own, such as an interactive map that ties venue geography to weather and highlights the matches at greatest high-temperature risk. This demonstrates its value for discovery and display via a user-friendly medium.
Moreover, we collect 18 data samples from three representative publication sources, each paired with the  expert-written piece.
For a comprehensive assessment, we design metrics along four complementary axes. 
\textbf{(a) Human--agent angle coverage} extracts the factual claims from articles and reports similarity-matched coverage between human and agent.
\textbf{(b) Rubric evaluation with human judges} asks 53 participants to score agent-generated or human-written articles blind on five rubric dimensions covering visual design, narrative pacing, data transparency, claim-data alignment, and insight value, and pick the preferred one overall. 
\textbf{(c) Computer-use agent as judge}:
we explore a cost-saving automatic proxy for how real-world users navigate and interact with an article, employing computer-use agents that perceive the rendered interface through actions such as clicking and scrolling;
\textbf{(d) Verifiability} uses a cross-family coding agent to validate claims by verifying statements such as executing code or searching the reference source.

Our experiments show that \methodshort{} produces multimodal articles that readers find compelling and are independently verifiable, with built-in evidence traceability at the claim level.
Human raters judge them favorably across multiple quality dimensions;
however, human journalists retain a clear edge in editorial angle, creative design, and informative presentation.
\methodshort{}'s greatest advantage instead lies in \textit{auditability}: it makes the evidentiary basis of each claim explicit and measurable — something even carefully crafted human articles rarely provide natively. 

We therefore position \methodshort{} as a collaborator rather than a replacement: humans set the perspective and editorial judgment, while (i) agents handle labor-intensive computation and graphics design and (ii) open the door to specialised, data-rich stories that newsrooms do not have the bandwidth to cover.
\section{Related Work}
\label{sec:related}

\begin{table*}[!t]
\centering
\small
\caption{\textbf{Comparison with related works.}
\textit{Ext. Search}: the system actively browses the web.
\textit{Narr. Angle}: the output is organized around a story angle rather than merely presenting data.
\textit{Multimodal} (\textit{Image}, \textit{Video}, \textit{Audio}, \textit{Interact.}): whether the system generates the corresponding modality or produces reader-interactive output.
\textit{Evidence} (\textit{Source}, \textit{Code}, \textit{Grounded}): whether the output cites sources, ships runnable code, and makes each claim independently verifiable.
\yes~present, \yespart~partially present or not provided by default, \no~absent.}
\label{tab:related_compare}
\footnotesize
\setlength{\tabcolsep}{1pt}
\renewcommand{\arraystretch}{1}
\begin{tabularx}{\textwidth}{@{}
  >{\raggedright\arraybackslash}p{3cm}
  >{\raggedright\arraybackslash}p{2.2cm}
  >{\raggedright\arraybackslash}p{1.2cm}
  *{9}{>{\centering\arraybackslash}X} @{}}
\toprule
\multirow{2}{*}{\textbf{System}}
 & \multirow{2}{*}{\textbf{Inputs}}
 & \multirow{2}{*}{\textbf{Outputs}}
 & \multirow{2}{*}{\shortstack{\textbf{Ext.}\\\textbf{Search}}}
 & \multirow{2}{*}{\shortstack{\textbf{Narr.}\\\textbf{Angle}}}
 & \multicolumn{4}{c}{\textbf{Multimodal Generative?}}
 & \multicolumn{3}{c}{\textbf{Evidence}} \\
\cmidrule(lr){6-9}\cmidrule(lr){10-12}
 & & & &
 & \textbf{Image}
 & \textbf{Video}
 & \textbf{Audio}
 & \textbf{Interact.}
 & \textbf{Source}
 & \textbf{Code}
 & \textbf{Grounded} \\
\midrule
\rowcolor{grouprow}\multicolumn{12}{@{}l}{\textit{Search Agents}} \\
MindSearch~\citep{mindsearch}                        & Query                    & Report                    & \yes & \no & \no  & \no  & \no  & \no  & \yes & \no  & \yespart \\
MMSearch~\citep{mmsearch}                           & Query+Image            & Text                      & \yes & \no & \no  & \no  & \no  & \no  & \yes  & \no  & \yespart  \\
DR Tulu~\citep{drtulu}                           & Query            & Text                      & \yes & \no & \no  & \no  & \no  & \no  & \yes  & \no  & \yespart  \\
\midrule
\rowcolor{grouprow}\multicolumn{12}{@{}l}{\textit{Data Visualization Agents}} \\
MatplotAgent~\citep{matplotagent}                      & Query+Data     & Infographic       & \no  & \no & \yes & \no  & \no  & \no  & \no  & \yes  & \yes  \\
LIDA~\citep{lida}                                    & Query+Data     & Infographic       & \no  & \no & \yes & \no  & \no  & \no  & \no  & \yes  & \yes  \\
CoDA~\citep{coda} & Query+Data     & Infographic              & \no  & \no & \yes & \no  & \no  & \yespart  & \no  & \yes & \yes  \\
\midrule
\rowcolor{grouprow}\multicolumn{12}{@{}l}{\textit{Data Science Agents}} \\
DSGym~\citep{dsgym}                                  & Query+Data     & Score              & \no  & \no & \no  & \no  & \no  & \no  & \no  & \yes & \yespart  \\
Data Interpreter~\citep{datainterpreter}             & Query+Data & Report         & \yes & \no & \yes & \no  & \no  & \yespart & \yes & \yes & \yespart  \\
AI Scientist~\citep{aiscientist,aiscientistv2}                & Query                    & Report                    & \yes & \yespart & \yes & \no  & \no  & \no  & \yes & \yes & \yespart \\
\midrule
\rowcolor{grouprow}\multicolumn{12}{@{}l}{\textit{Data Journalist Agents}} \\
LLM writer~\citep{journalistplan} & Press release            & Angle           & \no  & \yes & \no  & \no  & \no  & \no  & \yes  & \no  & \yespart  \\
\textcolor{gray}{Human writer}~\cite{handbook}              & Data            & Article                   & \yes & \yes & \yespart & \yespart & \yespart & \yespart & \yespart & \no & \yespart \\
\rowcolor{ourrow}\textbf{\methodshort (Ours)}        & Data  & \textbf{Article}   & \yes & \yes & \yes & \yes & \yes & \yes & \yes & \yes & \yes \\
\bottomrule
\label{tab:comparison}
\end{tabularx}
\end{table*}
In this section, we compare \methodshort{} against representative works in relevant fields. The comparison is illustrated in Tab.\ref{tab:comparison}.

\textbf{Computational Journalism} 
\label{sec:related:journalist}
is not a new field and auditability has been one of its central concerns from the start. For over a decade, researchers have studied how computation can support the discovery, production, and verification of public-interest news~\cite{cohen2011computational, hamilton2009accountability}. Coddington~\cite{coddington2015clarifying} separates computer-assisted reporting, data journalism, and computational journalism, placing the automation of editorial labour within a longer professional history. Diakopoulos~\cite{diakopoulos2019automating} shows that newsrooms already delegate parts of the workflow to algorithms, and argues that automation should augment rather than replace the journalist. A parallel strand treats transparency of data, method, and provenance as a precondition for trust in such systems~\cite{diakopoulos2017algorithmic}. Data journalism itself targets general-audience communication, either producing a publishable artefact for non-expert readers or studying the workflow empirically~\cite{handbook}. Recent work surveys practitioners on where generative tools help and where they threaten accuracy and credibility~\cite{diakopoulos2024generative}, and explores language models in journalistic roles such as planning articles, recommending angles, and identifying sources~\cite{brigham2024developing,journalistplan,cheng2025journalism,spangher2025novel,alshomary2026llms}. Closest to our setting, DataDirector~\cite{datadirector} fuses Vega-Lite charts, narration, and animation into a passive video.

Across this literature, one risk recurs: fluent, well-produced stories that do not expose how each claim was derived. Even the human data-journalist baseline, a multimedia article with inline citations and the gold-standard reader-facing form, rarely carries code-line provenance. \methodshort{} targets that gap. It does not study adoption or automate a single task; it routes structured multi-source data through seven specialized roles into a multimedia article whose Inspector binds every rendered sentence and chart to specific code lines or source URLs. We therefore frame \methodshort{} as a collaborator that strengthens editorial verification, not a replacement for the reporter.

\textbf{Deep Search Agents}
\label{sec:related:search}
take a natural-language query and autonomously browse the web to produce a retrieval-augmented text deliverable~\cite{rag}.
OpenAI’s Deep Research~\citep{openaidr} is the representative commercial demonstration, which browses the web~\citep{browsecomp} to collect knowledge, then augments the answer.
MindSearch~\citep{mindsearch} decomposes the query into a graph of atomic sub-questions, each answered by a search-and-summarize role, while DeepResearcher~\citep{deepresearcher} trains the browsing policy end-to-end with reinforcement learning. MMSearch~\citep{mmsearch} casts the requery, rerank, and summarize loop as a benchmark over short-answer outputs, and OpenResearcher~\citep{openresearcher} and DR Tulu~\citep{drtulu} extend the open-source side of this line with retrieval-augmented scientific question answering and long-form report generation. 
These systems optimize the retrieval and synthesis of sources in response to a given query, but their deliverable remains a source-centric text document: they surface and summarize evidence rather than construct a narrative angle, and the query, not an editorial judgment about what is worth telling, drives the output.

\textbf{Data Visualization Agents}
\label{sec:related:design}
convert a fixed input into a visual or narrative artifact. 
LIDA~\citep{lida} compiles a tabular dataset into executable visualization code, optionally restyled into an infographic. 
DataNarrative~\citep{datanarrative} pairs a generator and an evaluator to turn tables and a story intent into a narrative interleaved with chart specifications.
MatplotAgent~\citep{matplotagent} generates plotting code through a collaborative agent system, but fails in metadata analysis.
CoDA~\citep{coda} further coordinates specialized agents to carry a dataset through analysis and into a composed visual report. 
On the other hand, 
these systems operate on the data they are given: they assume the input dataset as fixed and do not actively search for external evidence, and their output is for the most part a static visual artifact rather than an interactive one.

\textbf{Data Science Agents}
\label{sec:related:dsagent}
take a task description with data files and use executed code to produce their deliverable. DSGym~\citep{dsgym} scores answer strings or CSV submissions in a sandbox with external tools disabled. DeepAnalyze~\citep{deepanalyze} trains an agentic model end-to-end to interleave analysis, code, and execution into a research report. Data Interpreter~\citep{datainterpreter} plans a task as a hierarchical subtask graph and emits whatever artifact the task requires, from a numeric answer to a playable mini-game. PublicAgent~\citep{publicagent} routes an ambiguous question through four agents that discover an open-data table and run validated experiments into a traceable report. 
DataStorm~\citep{datastorm} extends deep research from unstructured web corpora to large-scale structured databases, reframing the task as thesis-driven exploratory data analysis and data storytelling.
AI Scientist~\citep{aiscientist,aiscientistv2} chains literature retrieval, experimentation, and writing into a workshop paper that cleared peer review. Across these systems the deliverable is a structured text artifact, and the form stays text-and-charts even when the target reader is a non-expert. In contrast, \methodshort\ packages the analysis as a multimedia article rather than a static PDF, the form a data-journalism reader actually consumes.

\section{Data Journalist Agent}
\label{sec:method}

Given any raw data $\mathcal{D}$, the goal of \method\ is to produce an article $\mathcal{U}$ that is narratively compelling, visually appealing, and verifiable in its content. 
Notably, we do not provide a predefined angle or preference. Instead, \methodshort{} itself determines which aspects are worth reporting.
\begin{figure*}[!t]
    \centering
    \includegraphics[width=\textwidth]{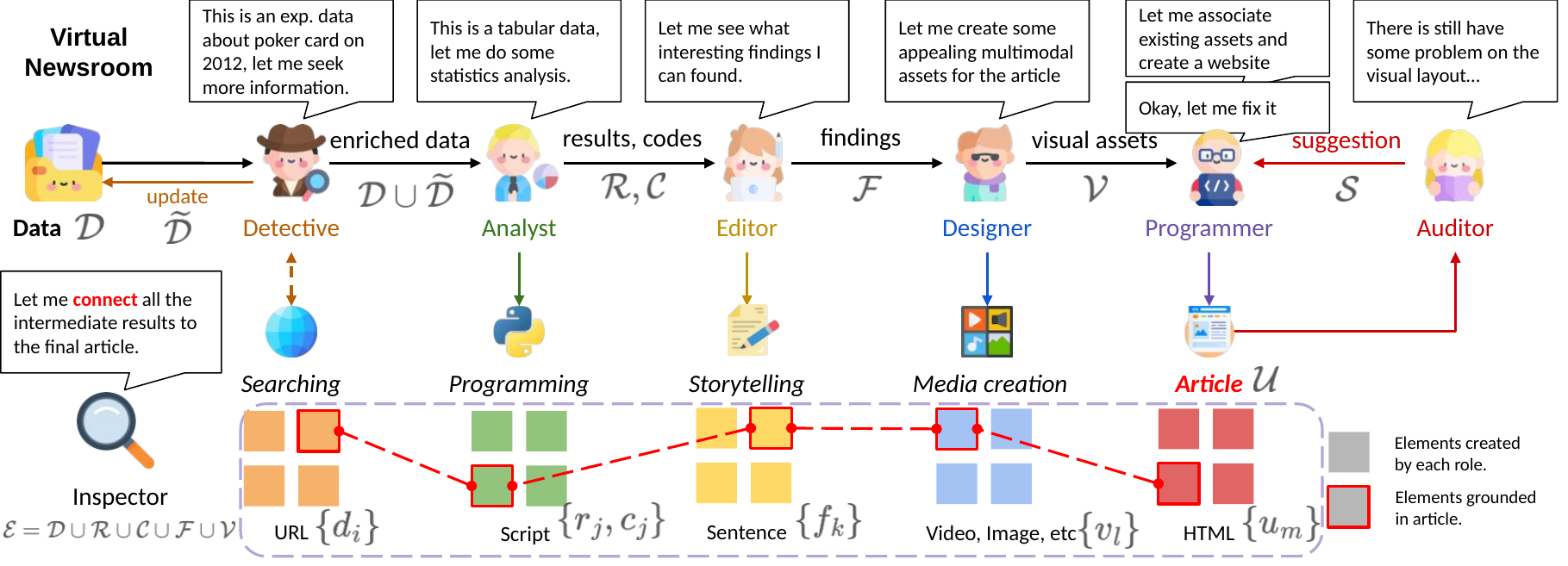}
    \caption{\textbf{The Virtual Newsroom for \methodshort{}.} A raw dataset $\mathcal{D}$ flows through a sequence of specialist roles: the \textit{Detective} gathers external context $\widetilde{\mathcal{D}}$ from the web, the \textit{Analyst} writes Python code $\mathcal{C}$ and emits results $\mathcal{R}$ with code-line provenance $\mathcal{R} \xleftarrow{\mathcal{C}} \mathcal{D} \cup \widetilde{\mathcal{D}}$, the \textit{Editor} drafts several findings $\mathcal{F}$ from different angles, the \textit{Designer} produces multimedia assets $\mathcal{V}$ via tool calls, and the \textit{Programmer} renders the final HTML $\mathcal{U}$. The page is then audited by the \textit{Auditor}, which provides suggestions $\mathcal{S}$ for visual and structural defects, and the \textit{Inspector}, which binds every published claim back to its supporting evidence $\mathcal{E} = {\mathcal{D}} \cup \mathcal{R} \cup \mathcal{C} \cup \mathcal{F} \cup \mathcal{V}$. Each role produces a set of intermediate elements (grey); those that ground the final article are highlighted (red outline), and the Inspector links them into a traceable evidence chain.}
    \label{fig:pipeline}
\vspace{-1em}
\end{figure*}

\subsection{The Virtual Newsroom}
\label{sec:method:roles}

As illustrated in Figure \ref{fig:pipeline}, we define our multi-agent solution as a virtual newsroom composed of specialised agent roles.

\paragraph{Detective.}
A raw data source is rarely enough on its own: an article almost always depends on context the dataset does not contain. For example, historical events often need to be associated with the time the data were released. The Detective gathers this context before any number is computed, so that downstream roles can frame the data rather than invent claims about it. Concretely, it augments the raw dataset $\mathcal{D}$ via web search into an enriched corpus $\mathcal{D} \cup \widetilde{\mathcal{D}}$, where $\widetilde{\mathcal{D}} \xleftarrow{\text{Web search}} \mathcal{D}$ contains additional context items tagged with category and source URLs, together with a small library of reference media (photographs, maps, short clips) that other agents can later reuse.

\paragraph{Analyst.}
A news article typically cites dozens of statistics to arrive at its insights. However, given a dataset, it is rarely clear in advance which statistical findings it admits, or which of them will prove most meaningful. The Analyst therefore prioritises completeness: it enumerates every analysis the dataset can support, profiles every column, and runs actual code rather than asking the model to estimate. From the augmented dataset, it derives a set of results $\mathcal{R}=\{r_i\}$ and supporting code $\mathcal{C}=\{c_i\}$ with $r_i \xleftarrow{c_i} \mathcal{D} \cup \widetilde{\mathcal{D}}$, where each finding $r_i$ carries a pointer to the script $c_i$ that generated it,  ensuring that every outcome is traceable.

\paragraph{Editor.}
An interesting analysis is not yet a story. Given a set of findings, the Editor decides what the article actually argues: which findings should lead, which should support, which add colour, and which should be cut. Reasoning over the Analyst's findings, it produces an editorial plan $\mathcal{F} \xleftarrow{\text{LLM}} \mathcal{R}$ that ranks each item by priority, selects the items worth keeping, and drafts a paragraph-level prose outline. Each finding $f_i$ in $\mathcal{F}$ is annotated with the upstream items it draws on, $f_i \sim (r_i, c_i)$.

\paragraph{Designer.}
An article is not just plain text: multimedia elements can substantially improve readability, such as maps for geography, audio for music, video for events, and interactive widgets for complex findings. For each finding $f_i$ of the editorial plan, the Designer reasons about what a reader would most want to see, then selects the medium that best fits the data, drawing on a suite of external generative tools such as text-to-image and text-to-video. 
The resulting per-section visual assets $\mathcal{V} \xleftarrow{\text{Tool}} \mathcal{F}$ include the corresponding asset calls needed to realise each medium, where we store every prompt or parameter.
Notably, although these generated media assets may not introduce new knowledge, they primarily serve an aesthetic purpose by making the article more engaging and easier to understand.

\paragraph{Programmer.}
Static formats such as PDF cannot natively coordinate multimedia elements; an HTML webpage, by contrast, is the ideal medium for what a reader actually sees. We therefore introduce a Programmer that renders the final page in HTML from the upstream artifacts. The Programmer generates no new facts or numbers; it operates in two modes. (i) In assembly mode, it quotes the upstream artifacts $\{\mathcal{F}, \mathcal{V}\}$ and composes them into a complete interactive article $\mathcal{U} \xleftarrow{} \{\mathcal{F}, \mathcal{V}\}$. (ii) In revision mode, it additionally takes the Auditor's revision suggestions $\mathcal{S}$ and revises the page accordingly, forwarding the audited article $\mathcal{U} \xleftarrow{} \{\mathcal{U}, \mathcal{S}\}$ to the Inspector.

\paragraph{Auditor.}
The rendered HTML may still harbour visual or structural defects: overlapping elements, broken charts, missing assets, or unresponsive interactions. Such defects can quietly undermine an otherwise well-grounded story. The Auditor therefore reviews the rendered page, $\mathcal{S} \xleftarrow{} \mathcal{U}$, and flags these issues; it returns the page to the Programmer for repair.

\subsection{How to ensure claims are verifiable?}
\label{sec:method:trace}

\begin{figure}[!h]
\centering
\includegraphics[width=\linewidth]{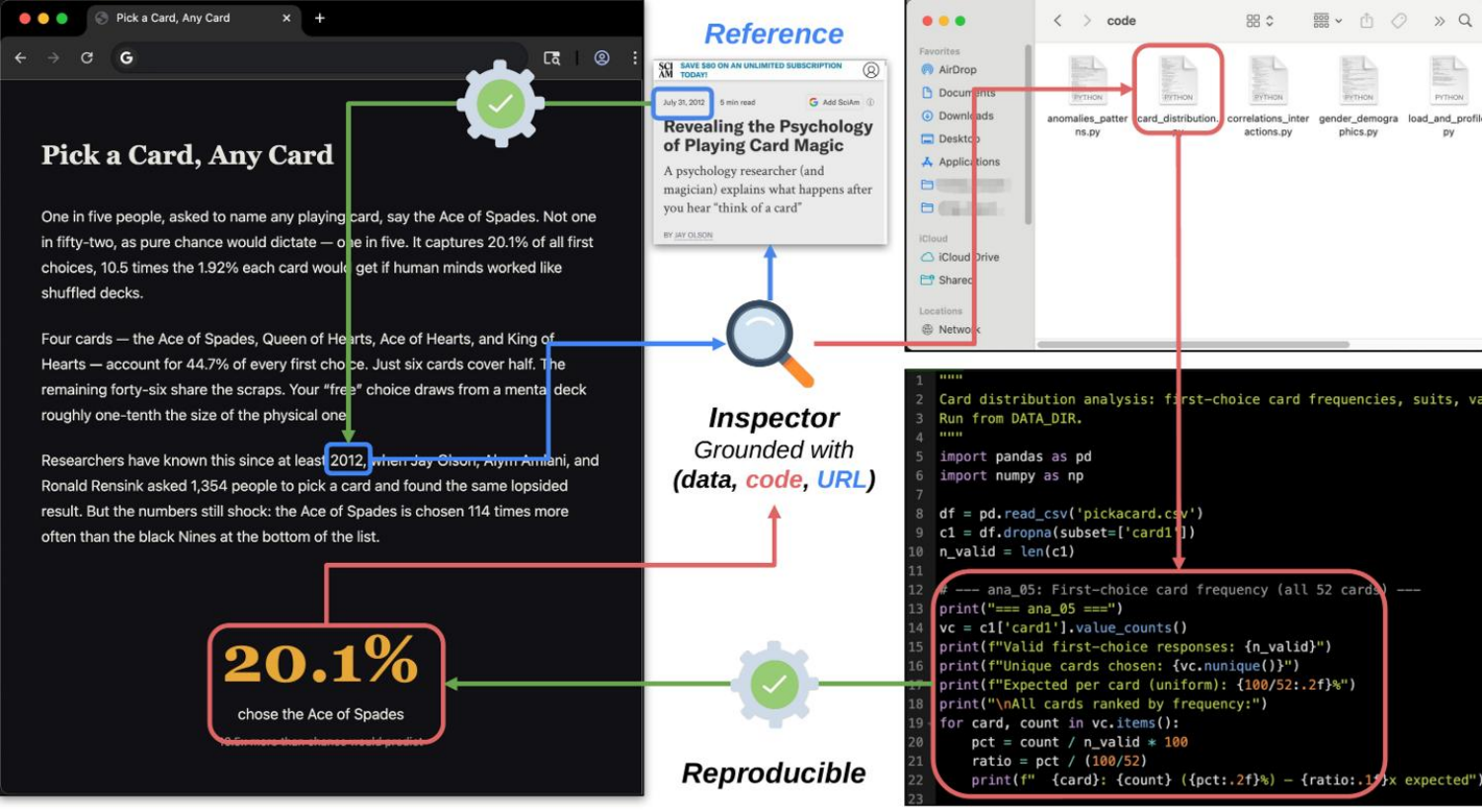}
\caption{\textbf{Illustration of the Inspector.} The Inspector binds every output finding back to its supporting evidence, which falls into two types: (i) \textit{code evidence}, the source file and specific line that produced a reported number, and (ii) \textit{reference evidence}, the external article or URL that grounds a contextual claim. The binding establishes auditability / traceability rather than factual correctness.
}
\vspace{-1em}
\label{fig:viewer}
\end{figure}

\paragraph{Inspector \faSearch.}
A central challenge for any multi-agent system that produces an article is that the reader has no reason to trust the page unless every visible element, from the lede sentence to the final tooltip, resolves to something concrete upstream (such as code or reference). We therefore introduce the Inspector, which closes this loop at the level of individual items.

We let all upstream agents each contribute \textit{atomic units of evidence}. 
The Detective contributes a context ${\mathcal{D}} = \{d_i\}$, where each $d_i$ is a context item with a source URL. The Analyst contributes findings $\mathcal{R} = \{r_j\}$ paired one-to-one with code $\mathcal{C} = \{c_j\}$, so that every $r_j$ is supported by the script $c_j$ that produced it. The Editor contributes a finding $\mathcal{F} = \{f_k\}$, where each $f_k$ is a paragraph with upstream pointers, and the Designer contributes specifications $\mathcal{V} = \{v_\ell\}$, where each $v_\ell$ is a per-section specification, and we record the tool call and parameters (such as prompts).
Together these form the pool of upstream evidence $\mathcal{E} = {\mathcal{D}} \cup \mathcal{R} \cup \mathcal{C} \cup \mathcal{F} \cup \mathcal{V}$.

The Inspector decomposes the audited page into a set of partial findings $\mathcal{U} = \{u_m\}$, where each $u_m$ is a self-contained HTML fragment realising a sentence, chart, or interactive element. It then binds every fragment $u_m$ to the entries of the evidence base $\mathcal{E}$ that ground it, \ie~$u_m \sim (d_i, r_j, c_j, f_k, v_l)$, so that each fragment carries an explicit link back to the evidence from which it was derived. 

The Inspector recognises two types of evidence link, as illustrated in Figure~\ref{fig:viewer}: \textit{code evidence}, where a claim traces back to the specific script and line that produced it, and \textit{reference evidence}, where a contextual claim is grounded in an external URL. 
The result is a page where truthfulness is \textbf{evidence-traceable}: every claim can be followed back through the Programmer, the Designer, and the Analyst to the original data file or source reference.
\begin{figure*}[!t]
\centering
\setlength{\tabcolsep}{6pt}
\renewcommand{\arraystretch}{1.15}
\begin{tabular}{@{}c|c|c@{}}
\toprule
\shortstack{\textbf{Sport \& Climate}\\[1pt] \footnotesize FIFA 2026 schedule \href{https://data2story.github.io/new/fifa26_schedule/blog_opus47_0525_1345/viewer.html}{[link]}} &
\shortstack{\textbf{Science}\\[1pt] \footnotesize ArXiv submissions 1991--2026
\href{https://data2story.github.io/new/arxiv/blog_opus47_0525_1802/viewer.html}{[link]}
} &
\shortstack{\textbf{Society}\\[1pt] \footnotesize Time-use diaries (MTUS)
\href{https://data2story.github.io/new/mtus/blog_opus47_0525_1248/viewer.html}{[link]}
}
\\[3pt]

\begin{subfigure}[b]{0.30\textwidth}
    \centering
    \includegraphics[width=\linewidth]{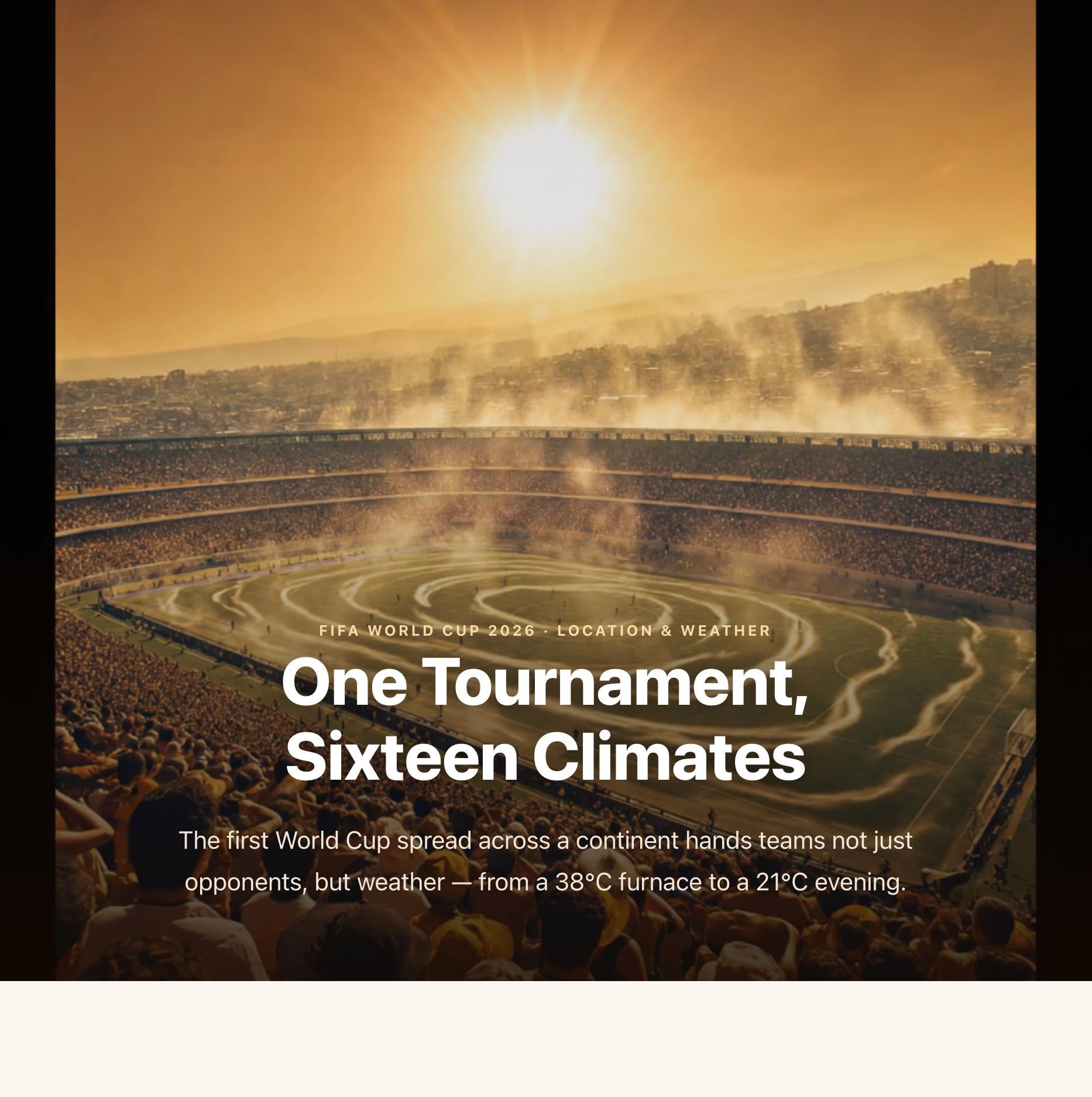}
    \caption{\textit{Sixteen Climates}}
    \label{fig:disc-a1}
\end{subfigure} &
\begin{subfigure}[b]{0.30\textwidth}
    \centering
    \includegraphics[width=\linewidth]{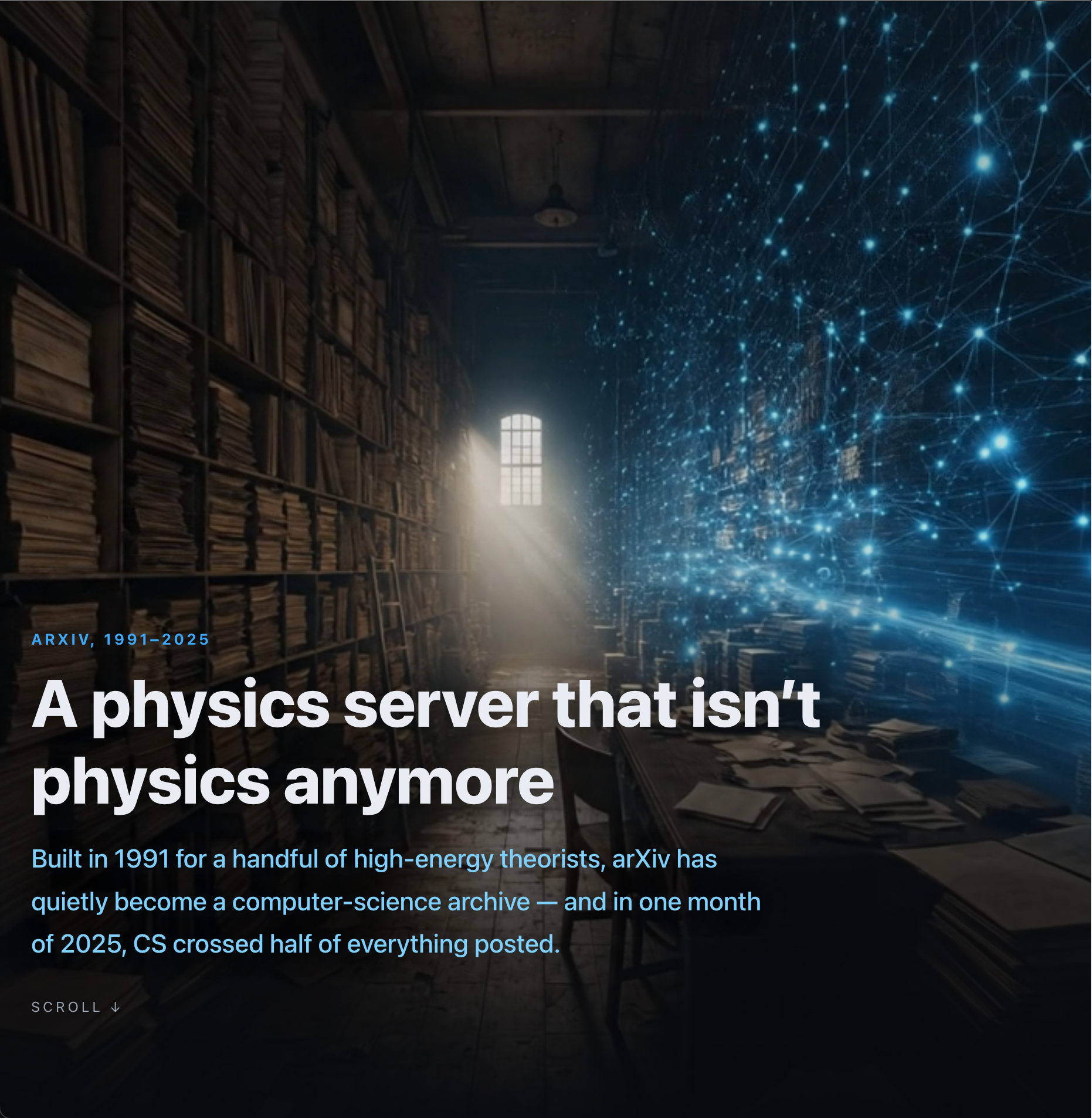}
    \caption{\textit{Not Physics Anymore}}
    \label{fig:disc-b1}
\end{subfigure} &
\begin{subfigure}[b]{0.30\textwidth}
    \centering
    \includegraphics[width=\linewidth]{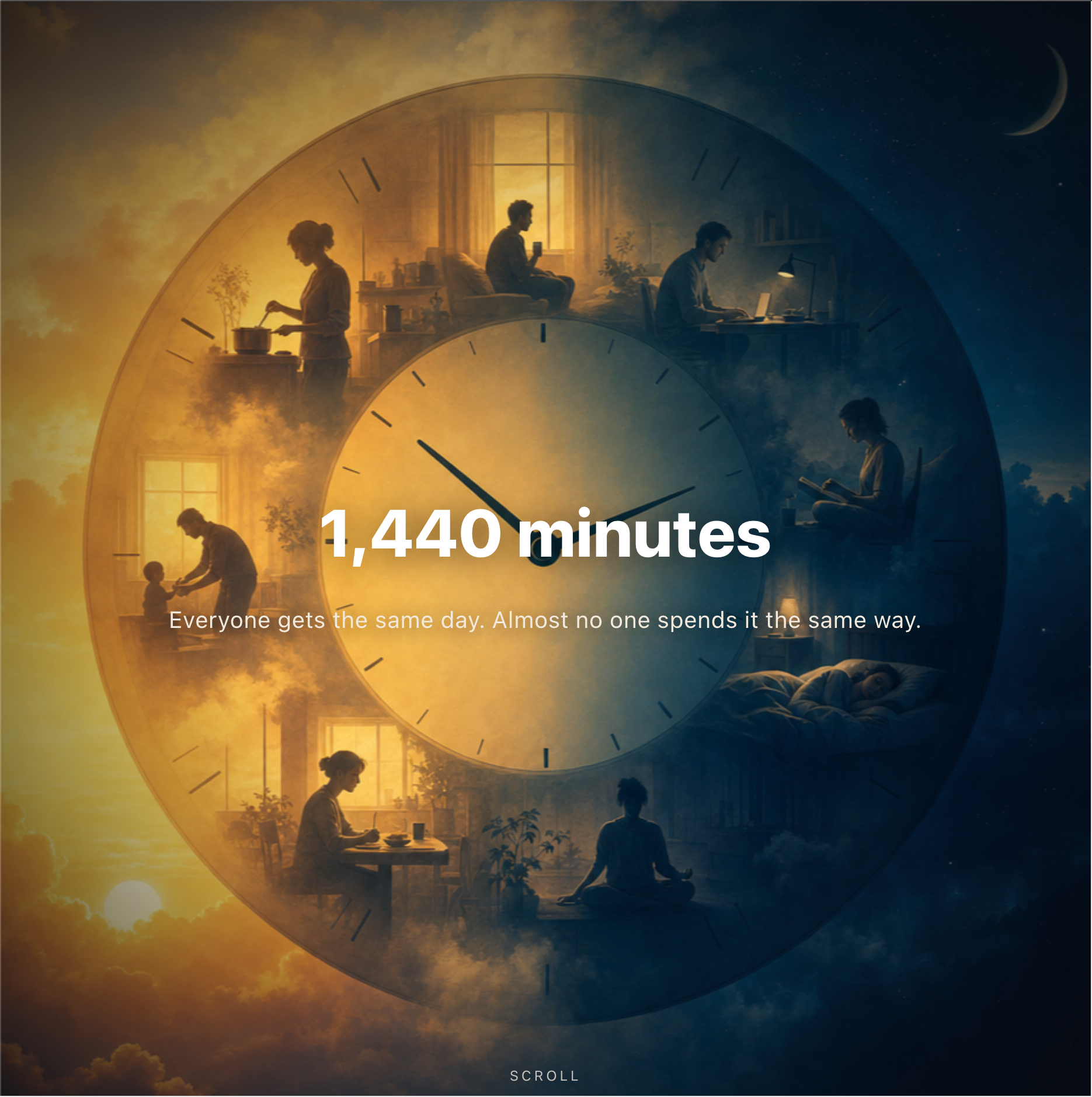}
    \caption{\textit{1,440 Minutes}}
    \label{fig:disc-c1}
\end{subfigure} \\[4pt]

\begin{subfigure}[b]{0.30\textwidth}
    \centering
    \includegraphics[height=4.4cm]{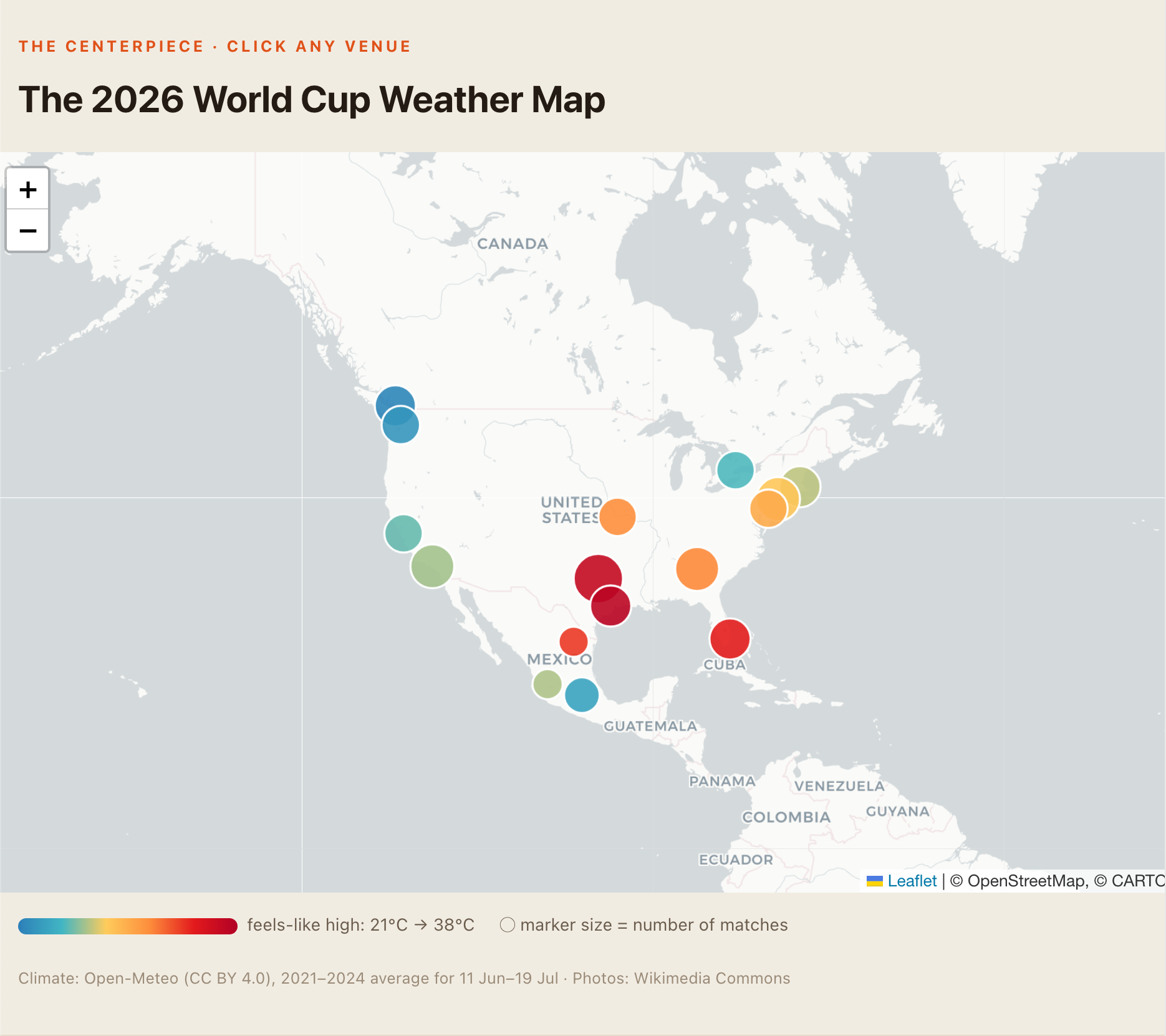}
    \caption{Interactive venue weather map}
    \label{fig:disc-a2}
\end{subfigure} &
\begin{subfigure}[b]{0.30\textwidth}
    \centering
    \includegraphics[height=4.4cm]{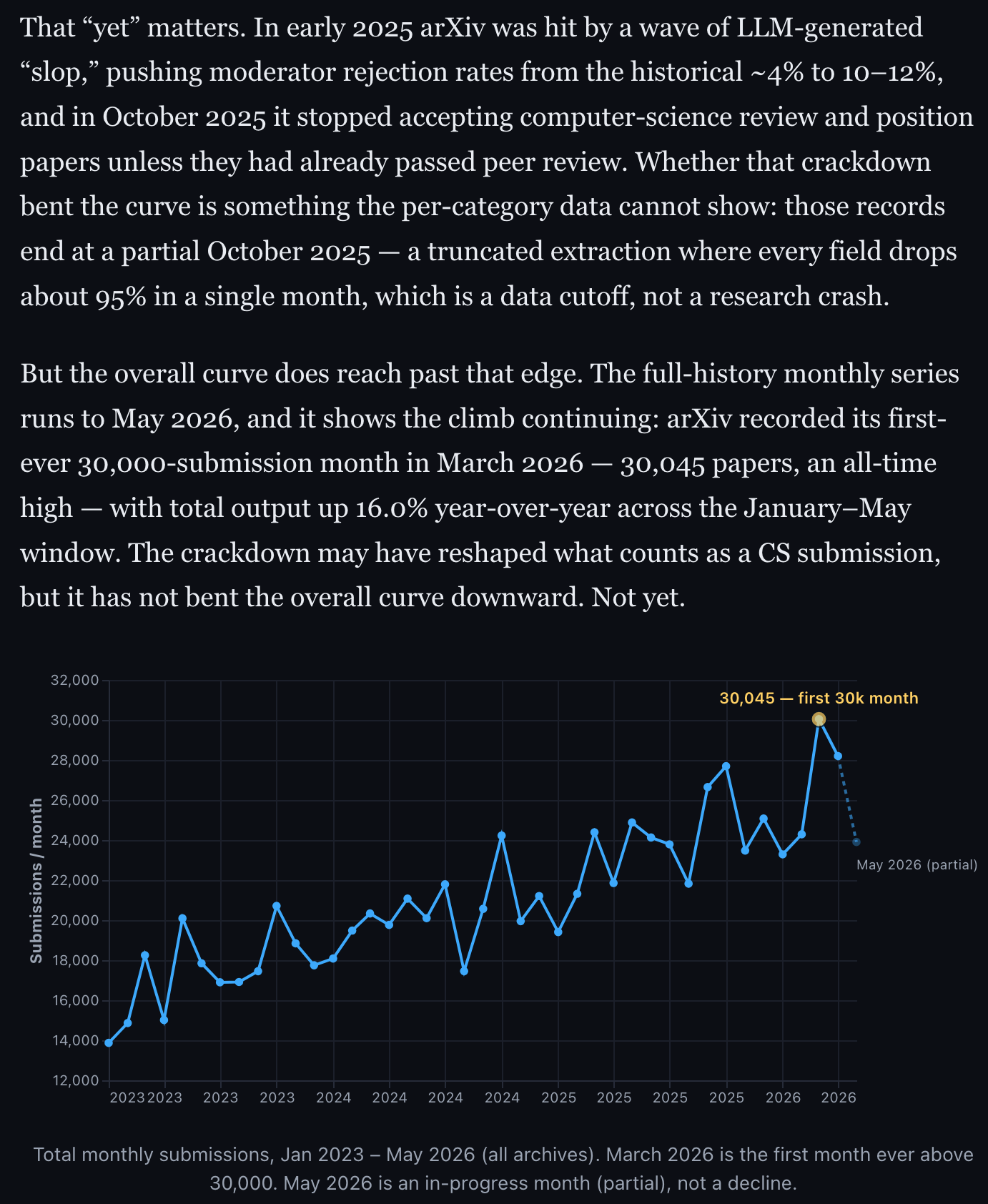}
    \caption{The climb past 30{,}000 a month}
    \label{fig:disc-b2}
\end{subfigure} &
\begin{subfigure}[b]{0.30\textwidth}
    \centering
    \includegraphics[height=4.4cm]{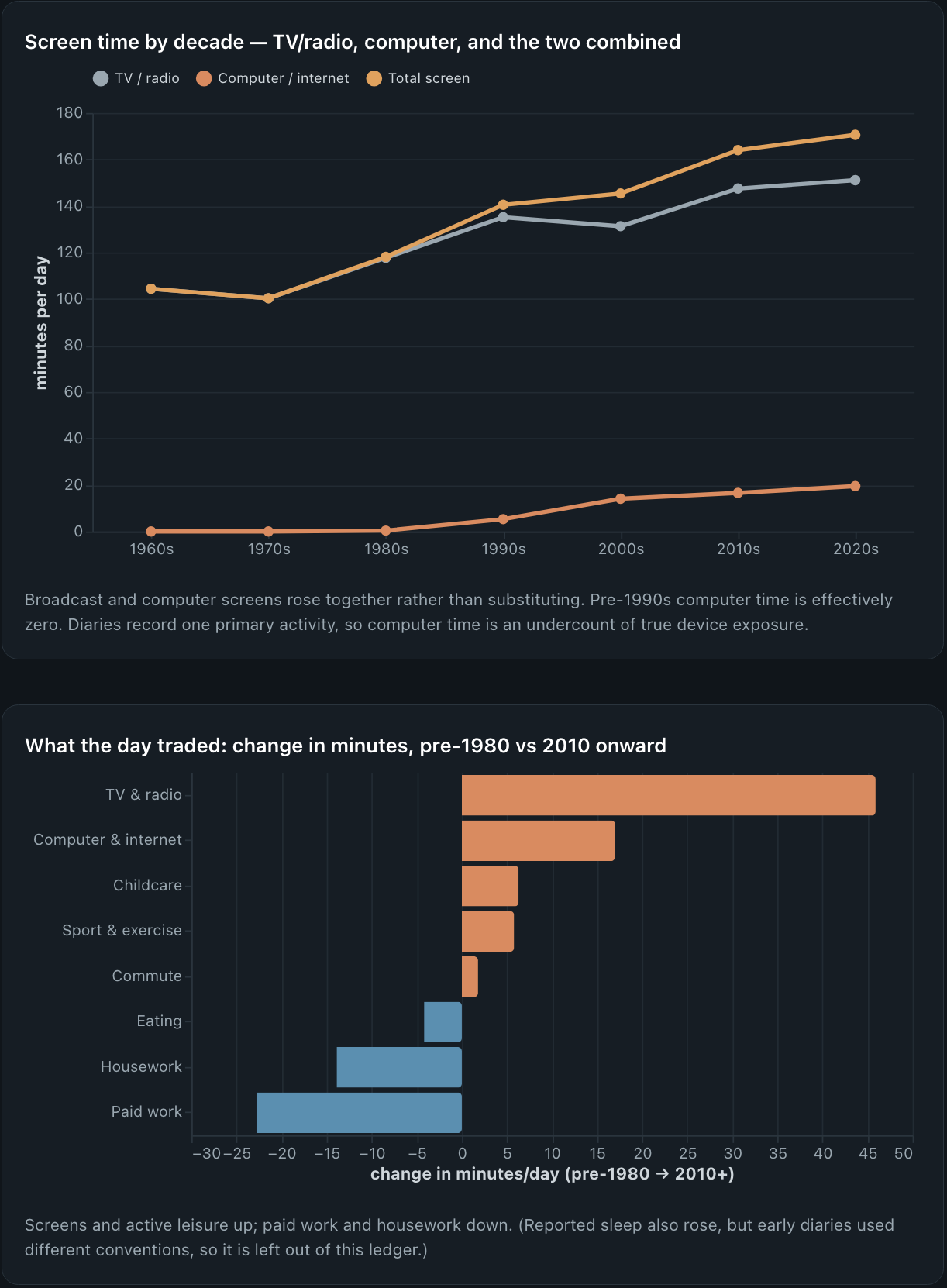}
    \caption{Screens vs.\ the day's trade-off}
    \label{fig:disc-c2}
\end{subfigure} \\
\bottomrule
\end{tabular}
\caption{\textbf{\methodshort{} discovering findings on new data with no human reference.} Three datasets from 2026 that have no canonical human-written piece, covering sport (\subref{fig:disc-a1}), science (\subref{fig:disc-b1}), and society (\subref{fig:disc-c1}). 
The \textbf{top row} is the opening cover of each piece; the \textbf{bottom row} is its signature data view.
}
\vspace{-1em}
\label{fig:discovery}
\end{figure*}

\subsection{\methodshort{} discovers findings on underexplored data}
\label{sec:discovery}
To illustrate \methodshort{}, we apply it to \emph{new} datasets rarely written by journalists, to show that it can autonomously \emph{discover} an original angle and back it with its own analysis. We chose three datasets that are publicly available in 2026 (Figure~\ref{fig:discovery}), spanning society, sport, and the AI industry. For each, we describe \methodshort{}'s writing angle and the core findings the article surfaces.

\textbf{(a) FIFA 2026 schedule\footnote{\url{https://www.fifa.com/}}: geography fused with climate.} 
The $2026$ World Cup is the first one to spread across a whole continent, so \methodshort{} fuses each venue's geographic location with its typical climate (Open-Meteo) and FIFPRO's heat-risk flags, interpreting the fixture list as a climate document rather than a sports calendar. The cover (Figure~\ref{fig:disc-a1}) embodies that tension with a blazing sun over a packed stadium under the title \textit{``One Tournament, Sixteen Climates,''} dramatising a feels-like gulf between a furnace-like Houston and a cool Vancouver, baked into the bracket before a ball is kicked. The core finding is striking: roughly four in ten matches are booked at the venues FIFPRO flags as ``extremely high risk,'' and humidity, not air temperature, drives the worst penalties. The interactive weather map (Figure~\ref{fig:disc-a2}), the article's centerpiece, lets the reader see this venue by venue. Throughout, the piece keeps the caveat visible: these are typical-climate odds, not a $2026$ forecast.

\textbf{(b) ArXiv submissions\footnote{\url{https://arxiv.org/stats/main}}: a physics preprint server that has become a computer science platform.} Reading three decades of submissions to arXiv, the preprint server that physicist Paul Ginsparg launched in $1991$, \methodshort{} writes from a contrarian angle: the ``physics archive'' everyone still pictures has quietly become a computer-science one. 
The cover (Figure~\ref{fig:disc-b1}), \textit{``A Physics Server That Isn't Physics Anymore,''} embodies this as a sunlit corridor of dusty paper stacks dissolving into a glowing data-network on the right, the founding discipline giving way to the field that overran it. 
The core finding is stark: computer science is now $42.5\%$ of everything posted, and in May $2025$ it crossed half of all submissions in a single month for the first time. 
The chart (Figure~\ref{fig:disc-b2}), with data running through $2026$, traces the total monthly output still bending upward, reaching arXiv's first-ever $30,000$-submission month in March $2026$. The piece ties this surge to a sharper policy turn: facing a wave of LLM-generated ``slop'' that sent rejection rates climbing, \textit{arXiv stopped treating an institutional email as enough to endorse a first-time submitter in January $2026$}, so for the first time the archive is actively deciding who gets to post.

\textbf{(c) Time-use diaries\footnote{\url{https://rdr.ucl.ac.uk/articles/dataset/Multinational_time_use_study_release_version_11/28682660}}: the day as a fairness ledger.} From the Multinational Time Use Study, harmonised from large-scale national diary surveys across dozens of countries and six decades, \methodshort{} writes from a single angle: a day is the one resource everyone owns in equal measure, exactly $1{,}440$ minutes, yet who spends them on unpaid work splits sharply by sex, country, and decade. The cover (Figure~\ref{fig:disc-c1}) embodies that angle with a luminous $24$-hour clock face filled with silhouettes of cooking, childcare, and sleep, captioned \textit{``everyone gets the same day, almost no one spends it the same way,''} so an abstract statistic becomes the reader's own morning. The core finding follows from the diaries: women do more than twice men's unpaid work, and once paid and unpaid hours are summed, they work longer days overall.
Read by decade (Figure~\ref{fig:disc-c2}), “\textbf{screen time}” (TV, radio, computer, internet) rose while paid work and housework fell, and the gender gap narrowed not because women were freed but because men slowly did more at home. The total work society performs barely moves over the decades; only its division by sex and kind shifts, the ``work-time invariance'' pattern.

\section{Evaluation}
\label{sec:eval}
In this section, we investigate three research questions:
\textbf{(i)} How can we fairly evaluate data-journalism articles produced by either humans or agents -- what metrics and protocols faithfully capture the quality of such outputs?
\textbf{(ii)} How do \methodshort-generated articles compare against human-written counterparts, and along which dimensions?
\textbf{(iii)} To what extent do human and agent judges agree, and how consistent are they across samples?

\subsection{Setting}
\label{sec:setting}

\begin{table}[!h]
\centering
\caption{\textbf{Evaluation set.} Each row pairs a dataset with a published human-written piece. 
Human articles rarely ship complete code, so \yespart in \textbf{Code} marks partial code (\eg~data-cleaning only).
}
\label{tab:datasets}
\footnotesize
\setlength{\tabcolsep}{4pt}
\renewcommand{\arraystretch}{1.15}
\begin{tabularx}{\linewidth}{@{}c c l l c >{\raggedright\arraybackslash}X@{}}
\toprule
\textbf{\#} & \textbf{Year} & \textbf{Domain} & \textbf{Modality} & \textbf{Code} & \textbf{Title of Human Article} \\
\midrule
\rowcolor{grouprow}\multicolumn{6}{@{}l}{\normalfont\rmfamily\textit{Source: The Economist (professional economics and statistical charts)}} \\
1 & 2018 & Science  & timeseries & & The space race is dominated by new contenders \href{https://www.economist.com/graphic-detail/2018/10/18/the-space-race-is-dominated-by-new-contenders}{[link]} \\
2 & 2018 & Media    & timeseries & & TV's golden age is real \href{https://www.economist.com/graphic-detail/2018/11/24/tvs-golden-age-is-real}{[link]} \\
3 & 2019 & Sports   & panel      & & Managers in football matter much less than most fans think \href{https://www.economist.com/graphic-detail/2019/01/19/managers-in-football-matter-much-less-than-most-fans-think}{[link]} \\
4 & 2019 & Politics & geospatial & & Israel's growing settlements force stark choices about its future \href{https://www.economist.com/graphic-detail/2019/02/02/israels-growing-settlements-force-stark-choices-about-its-future}{[link]} \\
5 & 2019 & Media    & tabular    & & The Oscars' influence has waned \href{https://www.economist.com/graphic-detail/2019/03/02/the-oscars-influence-has-waned}{[link]} \\
6 & 2020 & Health   & panel      & & Tourism flows and death rates suggest covid-19 is being under-reported \href{https://www.economist.com/graphic-detail/2020/03/07/tourism-flows-and-death-rates-suggest-covid-19-is-being-under-reported}{[link]} \\
\midrule
\rowcolor{grouprow}\multicolumn{6}{@{}l}{\normalfont\rmfamily\textit{Source: The Pudding (engaging visual storytelling and art-led design)}} \\
7  & 2018 & Culture  & text       & & The Structure of Stand-Up Comedy \href{https://pudding.cool/2018/02/stand-up}{[link]} \\
8  & 2019 & Music    & tabular    & & Vocal Register in Pop Music \href{https://pudding.cool/2019/08/register/}{[link]} \\
9  & 2023 & Music    & tabular    & & They Won't Play a Lady-O on Country Radio \href{https://pudding.cool/2023/05/country-radio/}{[link]} \\
10 & 2018 & Music    & tabular    & & The Eras of Boy Bands \href{https://pudding.cool/2018/11/boy-bands/}{[link]} \\
11 & 2017 & Health   & geospatial & & How Far Is Too Far for an Abortion Clinic \href{https://pudding.cool/2017/09/clinics/}{[link]} \\
12 & 2018 & Food     & text       & & Baking the Most Average Chocolate Chip Cookie \href{https://pudding.cool/2018/05/cookies}{[link]} \\
\midrule
\rowcolor{grouprow}\multicolumn{6}{@{}l}{\normalfont\rmfamily\textit{Source: TidyTuesday (casual and diverse topics)}} \\
13 & 2019 & Tech     & timeseries & \yespart & Technological Progress (Moore's Law) \href{https://ourworldindata.org/moores-law}{[link]} \\
14 & 2026 & Science  & text       & \yespart & How Many Decimals of Pi Do We Really Need? \href{https://www.jpl.nasa.gov/edu/news/how-many-decimals-of-pi-do-we-really-need/}{[link]} \\
15 & 2020 & Music    & text       & \yespart & Taylor Swift and Beyonc\'e Lyric Analysis \href{https://rpubs.com/RosieB/taylorswiftlyricanalysis}{[link]} \\
16 & 2026 & Climate  & tabular    & \yespart & Repair Caf\'es and Consumer Waste \href{https://insideclimatenews.org/news/11112025/todays-climate-repair-cafe-consumer-waste/}{[link]} \\
17 & 2026 & Sports   & tabular    & \yespart & Milano Cortina 2026 Olympic Schedule \href{https://www.olympics.com/en/milano-cortina-2026/schedule}{[link]} \\
18 & 2025 & Climate  & timeseries & \yespart & Sechsel\"auten Snowman (B\"o\"ogg) Forecast \href{https://www.meteoswiss.admin.ch/weather/weather-and-climate-from-a-to-z/boeoegg-prediction.html}{[link]} \\
\bottomrule
\end{tabularx}
\end{table}

We evaluate \methodshort\ on various  examples drawn from three stylistically distinct sources, deliberately chosen to span the spectrum of contemporary data storytelling. In curating the evaluation set, we sought diversity along the following axes: domain (science, media, sports, politics, health, and others), temporal coverage (spanning 2018–2026), and data modality (time series and tabular data, among others). 

As different publishers may have distinct editorial styles and audiences. We thus select three distinct editorial styles:
\href{https://www.economist.com/graphic-detail}{(i) The Economist}, featuring concise, analytical economics-style reporting; 
\href{https://pudding.cool/}{(ii) The Pudding}, known for artistically rich, long-form interactive essays; 
and \href{https://github.com/rfordatascience/tidytuesday}{(iii) TidyTuesday}, a community initiative providing more diverse datasets together with data-processing code and their original source articles. 
For every example, we pair the underlying data with the human-written reference piece, enabling head-to-head comparison against the \methodshort\ outcome. 
Table~\ref{tab:datasets} lists all 18 original articles. 

\textbf{Potential training-data contamination.} 
We acknowledge that well-known articles from `The Economist' and `The Pudding' may be included in the models’ pretraining corpora. Moreover, because the scope of the models’ web searches is difficult to control, we cannot completely rule out potential exposure to these articles. There is why we position agents as partner that augment human journalists rather than replace them.
Notably, under our evaluation framework, \textit{merely recalling or retrieving text from human-written articles receives no credit}:
(i) coverage is bidirectional, rewarding not only matching the human angle but also surfacing claims the human article omits, which memorising that article cannot supply;
and (ii) human articles ship no code, so even a memorised angle cannot help pass verifiability, which a cross-family verifier checks by re-running code against the data.

\methodshort\ articles are produced using Claude Code with \texttt{claude-opus-4.7}. We provide full details information in Appendix \ref{sec:app:rubric}.

\subsection{Evaluation Metrics}
\label{sec:eval:metrics}

We evaluate \methodshort\ along three orthogonal axes.

\begin{figure}[t]
\centering
\includegraphics[width=\linewidth]{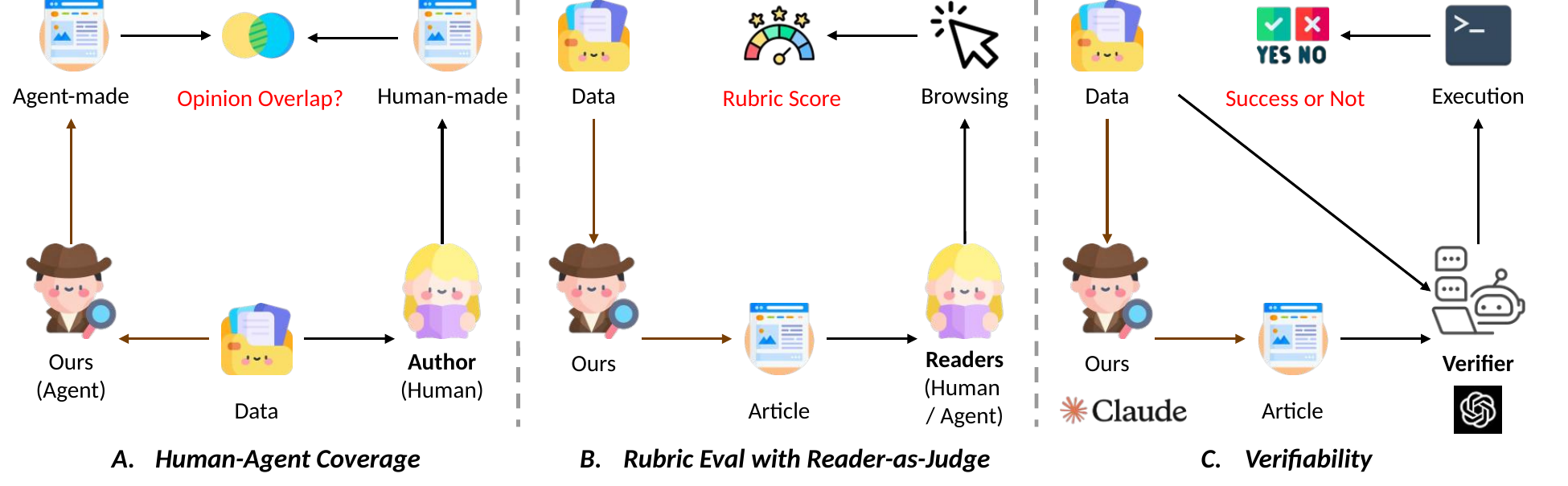}
\caption{\textbf{Three complementary evaluation protocols for \methodshort.} 
\textbf{(A) Human-agent angle coverage:} the agent and a human author independently produce articles from the same dataset; we measure overlap in the claims and insights surfaced by each. \textbf{(B) Rubric evaluation with reader as judge:} a human (or a computer-use agent) reader scores the agent's article against the human-written reference along five rubric dimensions, yielding graded quality assessments. \textbf{(C) Verifiability:} a verifier agent attempts to reproduce the agent's output from the same inputs, yielding a binary judgment of whether the artefact is faithfully verifiable.}
\vspace{-1em}
\label{fig:metric}
\end{figure}

\textbf{Human-agent angle coverage.}
For every paired human--agent article, we measure how much overlap exists between the human-written reference and the \methodshort\ output (as shown in Figure \ref{fig:metric}A). We parse the article into various sentences, then apply \texttt{gpt-4o-mini} to filter the article content (such as advertisements), resulting in a set of factual claims from the human article $\mathrm{Human}$ and from the agent article $\mathrm{Agent}$ respectively.
We then match claims across the two sides: OpenAI's \texttt{text-embedding-3-small} retrieves the top-3 nearest candidates by cosine similarity, and \texttt{gpt-4o-mini} decides under a relaxed prompt whether the candidate pair covers the same topic. A claim is covered if at least one of its candidates passes the LLM check. This gives us two directional coverage scores:
\begin{itemize}
    \item \textbf{Human-in-Agent} $\mathrm{P}(\mathrm{Agent} \mid \mathrm{Human})$: the fraction of human claims that the agent article also surfaces. \textit{\ie~did the agent catch what a journalist would catch?}
    \item \textbf{Agent-in-Human} $\mathrm{P}(\mathrm{Human} \mid \mathrm{Agent})$: the fraction of agent claims that also appear in the human article, indicating how closely the agent's claims track the human-curated angle.
\end{itemize}
Formally,
\[
\mathrm{P}(\mathrm{Agent} \mid \mathrm{Human}) \;=\; \frac{|\mathrm{Human} \cap \mathrm{Agent}|}{|\mathrm{Human}|}, \qquad
\mathrm{P}(\mathrm{Human} \mid \mathrm{Agent}) \;=\; \frac{|\mathrm{Human} \cap \mathrm{Agent}|}{|\mathrm{Agent}|},
\]
where $\mathrm{Human} \cap \mathrm{Agent}$ denotes the set of claims matched across both sides. A high $\mathrm{P}(\mathrm{Agent} \mid \mathrm{Human})$ indicates that the agent covers more of the human's angle, while a high $\mathrm{P}(\mathrm{Human} \mid \mathrm{Agent})$ indicates that the human covers more of the agent's angle;
a high value on either side indicates strong coverage of that side's angle, while a gap between the two reflects claims unique to one side, whether from divergence or broader coverage.

\textbf{Human as judge \& Rubric evaluation.}
We clarify the scope of our evaluation. The evaluation measures reader reception rather than newsroom adoption: our participants judge the final articles as readers would, not whether a practising journalist would adopt \methodshort{}. Robust evaluation with professional journalists requires expert recruitment and realistic editorial settings, which is hard to perform at scale, so we treat it as future work (Section~\ref{sec:con}).

An article is ultimately meant to be read, so the \textit{primary} evaluation is to place it in front of readers (illustrated in Figure~\ref{fig:metric}B). 
Because a data-driven article is not a single output but a composite artefact spanning prose, visuals, and analysis, a one-dimensional score cannot capture its quality~\cite{handbook,datajournalism}; we thus assess it along a rubric. We recruit $53$ reviewers via the Prolific platform\footnote{https://www.prolific.com/}; each is assigned one \methodshort{}--human pair (presentation order randomised, blind) and scores both versions along the five rubric dimensions below on a $1$--$7$ scale:

\begin{enumerate}
\item \textbf{Visual Design~\cite{tufte2001visual,ware2004information}.} Whether palette, typography, layout, and chart-type choice are polished and well matched to the claim each chart supports.
\item \textbf{Narrative \& Pacing~\cite{segel2010narrative,knaflic2025storytelling}.} Whether the hook, ordering, rhythm, and ending make the artefact read as a guided tour rather than a list of facts.
\item \textbf{Data \& Method Transparency~\cite{cohen2011computational,diakopoulos2015algorithmic}.} Whether sources are cited specifically, methodology is described, data is accessible, and limitations are acknowledged with concrete numbers or exclusions.
\item \textbf{Claim--Data Alignment~\cite{gelman2013garden,cairo2016truthful}.} Whether quantitative claims are bounded by what the data can support, confounders are named, and chart encodings are unambiguous.
\item \textbf{Insight Value~\cite{grice1975logic,north2006toward}.} Whether the reader gains a non-trivial cognitive update; capped at 3 if the takeaway restates common knowledge, capped at 5 if the update is meaningful only to a lay reader.
\end{enumerate}

After viewing both, each reviewer also expresses a binary preference indicating which version they prefer overall.

\textbf{Computer-use agent as Judge.}
Beyond human evaluation (which requires costly manual efforts), we also consider a cost-saving protocol that uses a model as judge. This follows the same setup as Figure~\ref{fig:metric}B, with the human reader replaced by an agent. We use an across-family agent, OpenAI’s browser-use \texttt{gpt-5.5-xhigh}. 
An article, however, is an interactive website: standard LLM~\cite{zheng2023judging,zhuge2024agent} or VLM~\cite{chen2024mllm} judges perceive only static screenshots and cannot scroll, hover, or trigger animations, missing precisely the dynamic elements that distinguish a polished interactive piece from a static one. We therefore employ a computer-use agent~\cite{zhou2024webarena} as judge, which navigates the rendered page like a human reader and scores it along the same rubric dimensions used in our human studies. 

\textbf{Verifiability.}
To verify that the published narrative is faithfully grounded in the underlying data (Figure \ref{fig:metric} C), we replay every article with an across-family verifier (OpenAI’s Coder \texttt{codex-GPT-5.4}). 
From each article, we extract the set of factual statements $\mathrm{S}$, which fall into two categories:
(a) \emph{computational claims}, \ie~numbers or findings derived from the data, which the checker verifies by re-executing the supporting \texttt{Python} or \texttt{R} scripts against the raw dataset;
and (b) \emph{reference-supported claims}, \ie~statements backed by an external reference, which the checker verifies by re-fetching the cited source URL and confirming the claim against its content.
For each claim, the checker returns a boolean result. We report the average pass rate as the article verifiability rate.

Notably, in verifiability experiments, the verifier has access to the original dataset when evaluating human-written articles, rather than the article text alone. For agent-written articles, the verifier additionally receives the full reasoning trajectory (by our Inspector) — a form of provenance made possible by evidence-grounded design.
\subsection{Experiment Results}
\label{sec:res}

\subsubsection{Distribution of article composition: where do humans and agents differ?}
\begin{figure}[!h]
    \centering
    \begin{subfigure}[t]{0.32\linewidth}
        \centering
        \includegraphics[width=\linewidth]{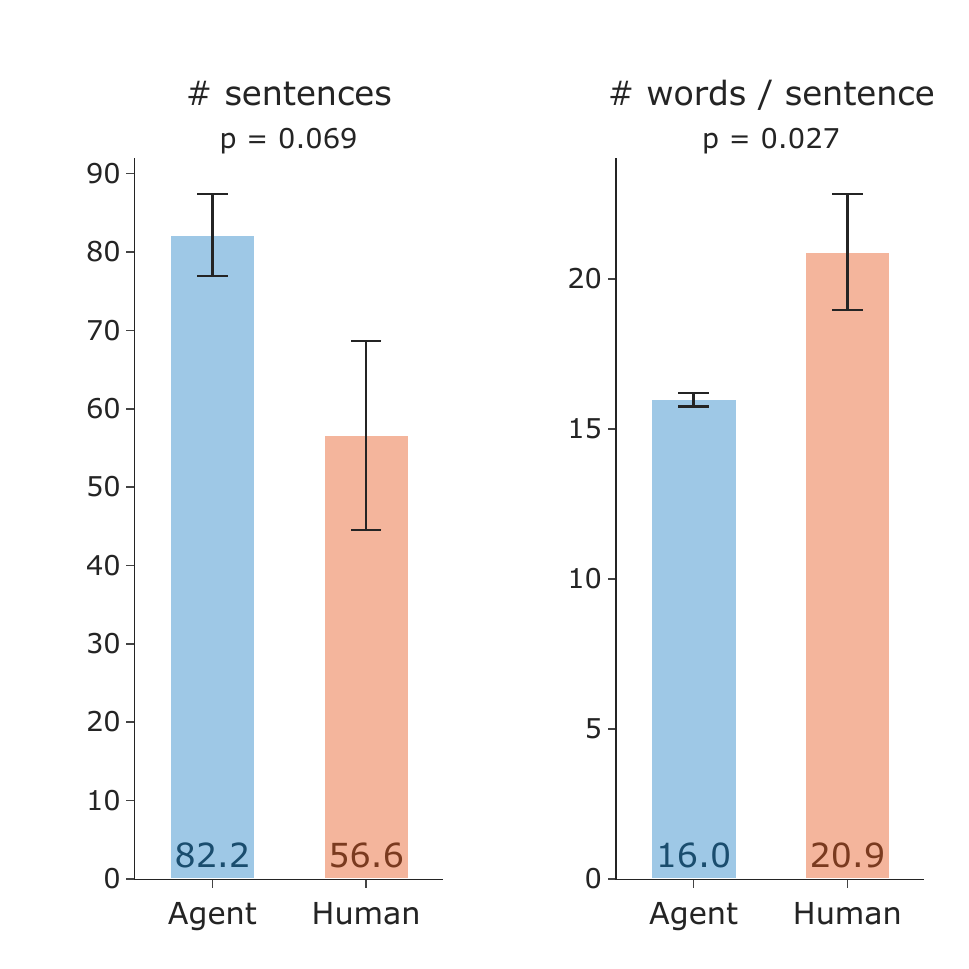}
        \caption{Num. of sentences per article and Avg. words per sentence.}
        \label{fig:dist-periods}
    \end{subfigure}
    \begin{subfigure}[t]{0.32\linewidth}
        \centering
        \includegraphics[width=\linewidth]{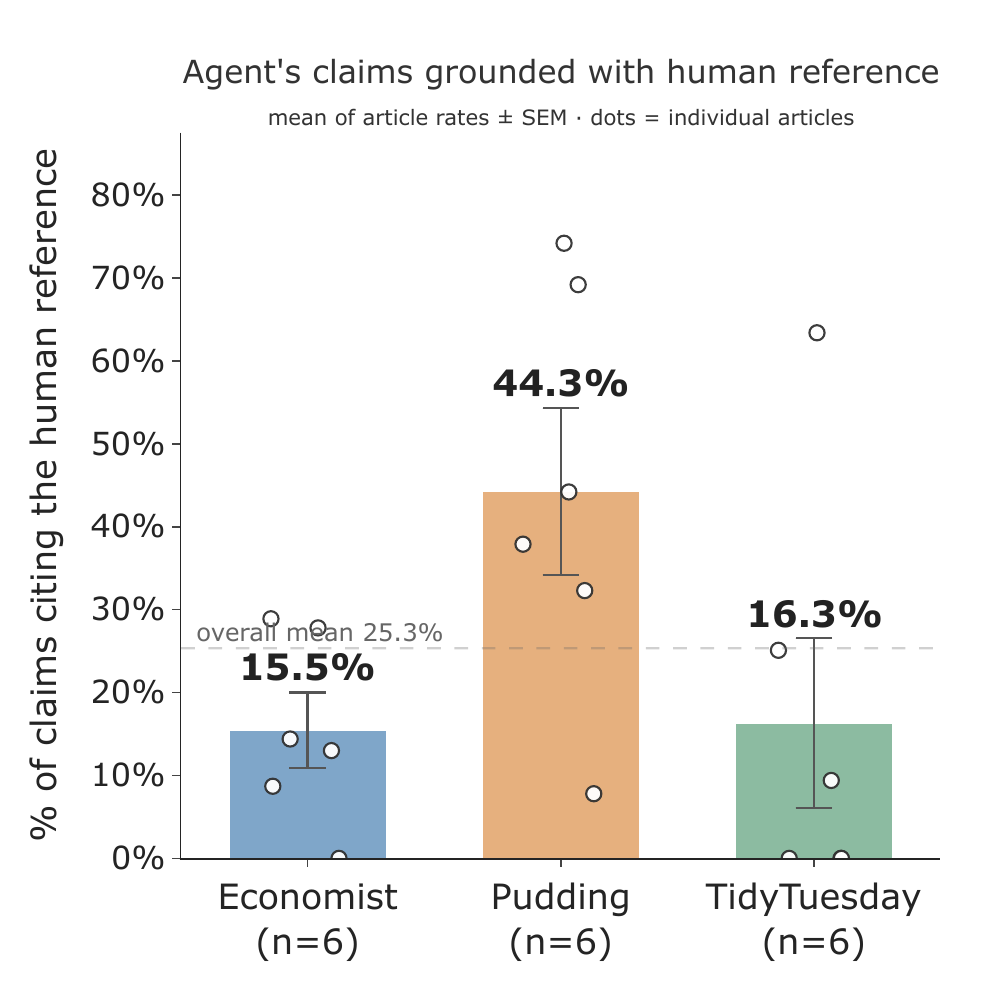}
        \caption{Percentage of agent’s claims cited with human reference.}
        \label{fig:dist-ground}
    \end{subfigure}
    \begin{subfigure}[t]{0.32\linewidth}
        \centering
        \includegraphics[width=\linewidth]{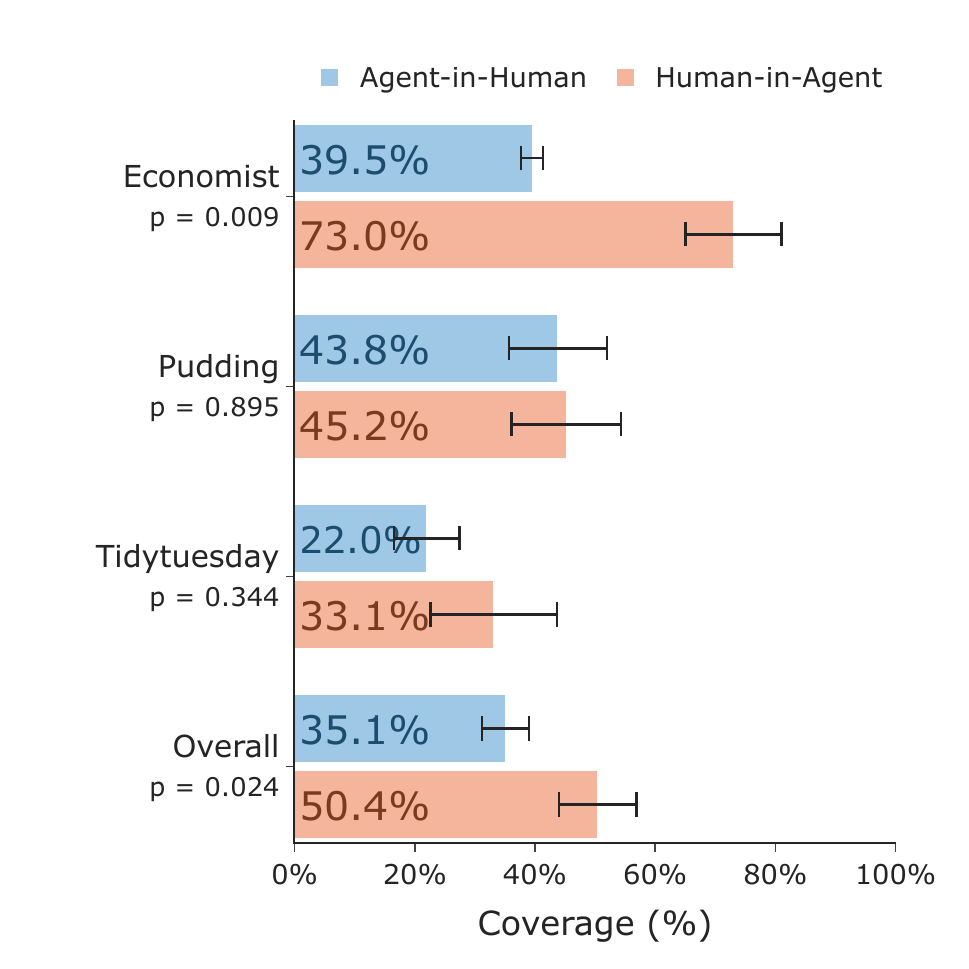}
        \caption{Claim coverage between human-written and agent-generated articles.}
        \label{fig:dist-coverage}
    \end{subfigure}
    \caption{\textbf{Textual distribution (a), Claims overlap with humans (b) and Content coverage (c)} across 18 samples, reported by “mean $\pm$SEM” with p value.}
    \label{fig:dist}
\end{figure}

Before examining the article content, a natural first check is whether \methodshort{} writes at human scale. Across the 18 paired articles (Figure~\ref{fig:dist-periods}), the total writing volume comes out roughly comparable (1305 for \methodshort{} and 1557 for humans), while the agent uses $1.45{\times}$ as many sentences but each is shorter ($0.77{\times}$); the articles made by \methodshort{} are broken into shorter, more granular statements.

Matching textual statistics is one thing; the angle behind the text is what matters.

First, \textit{how much does the agent lean on the human article itself?}
Figure~\ref{fig:dist-ground} measures the share of the agent's claims whose sentence-level provenance cites the human reference: $14/18$ articles cite it at all, but the density is strongly source-shaped --- highest on `Pudding' ($44.3\%$), where the human article \emph{is} the dataset (data-as-article), and far lower on `Economist' ($15.5\%$) and `TidyTuesday' ($16.3\%$).
Second, \textit{how much does the agent reproduce the human's angle?}
As shown in Figure~\ref{fig:dist-coverage}, claim-level coverage points clearly one way: about half of the human article angle (50.4\%) lands in the agent's article, while only a third (35.1\%) of the agent’s angle maps back.

Crucially, the per-source coverage pattern is the \emph{inverse} of grounding: coverage is widest exactly where grounding is lowest. It peaks on `Economist' short briefings ($\Delta=73.0\%-\ 39.5\%=33.5\%$), whose narrow single-topic scope (typically a standard statistic or chart) makes them easy for the agent to predict and cover from primary data \emph{without} leaning on the original (hence the low $15.5\%$ grounding);
it stays uniformly lower on `Pudding' and `TidyTuesday', whose source articles either carry a single editorial thesis the agent does not fully reproduce (creative long-form storytelling --- \emph{even though} it grounds most heavily in those pieces) or span diverse topics and external sources (`TidyTuesday').
Read together, the two views separate two distinct skills --- \emph{grounding in a source} versus \emph{reproducing its angle}: \methodshort{} reliably absorbs and rewrites the easy, predictable angles, and leans on the human text most where that text is itself the data, but \textit{reproducing a human author's narrative arc remains the harder problem.}


\begin{figure}[!h]
    \centering
    \begin{subfigure}[t]{0.4\linewidth}
        \centering
        \includegraphics[width=\linewidth]{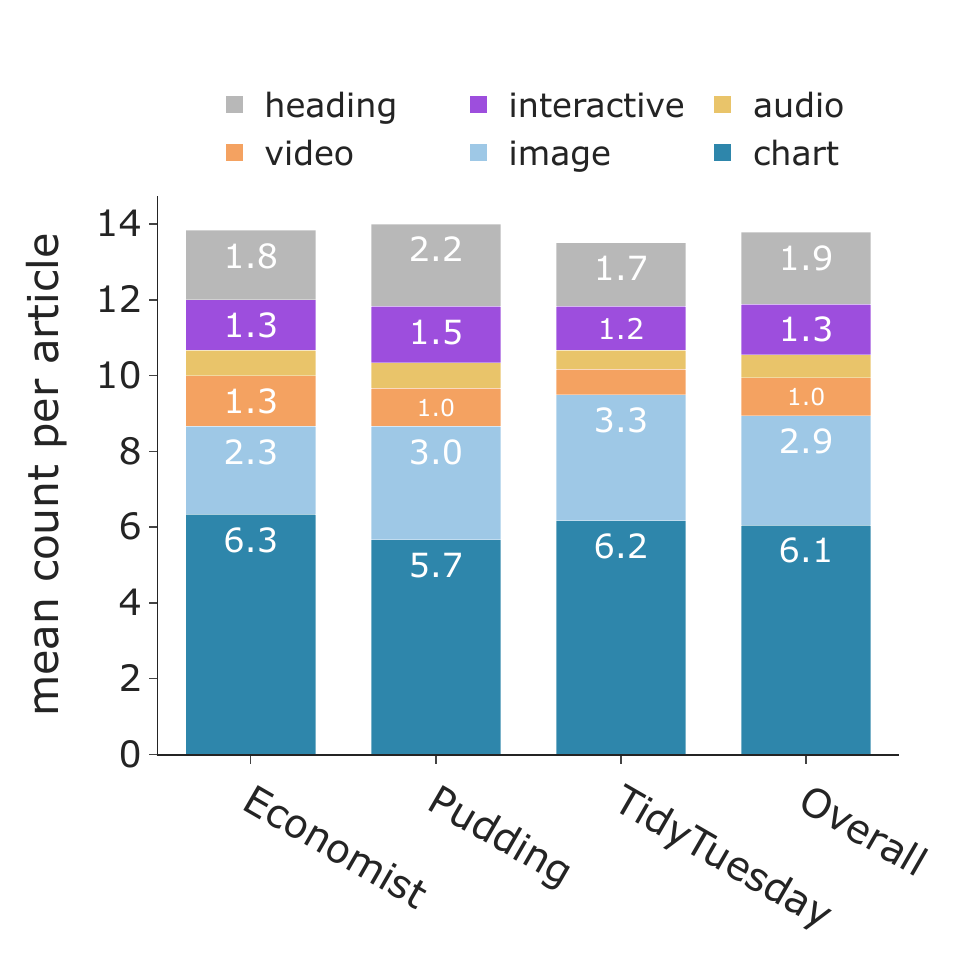}
        \caption{Articles made by \methodshort{}.}
        \label{fig:multimodal-agent-bygroup}
    \end{subfigure}
    \begin{subfigure}[t]{0.4\linewidth}
        \centering
        \includegraphics[width=\linewidth]{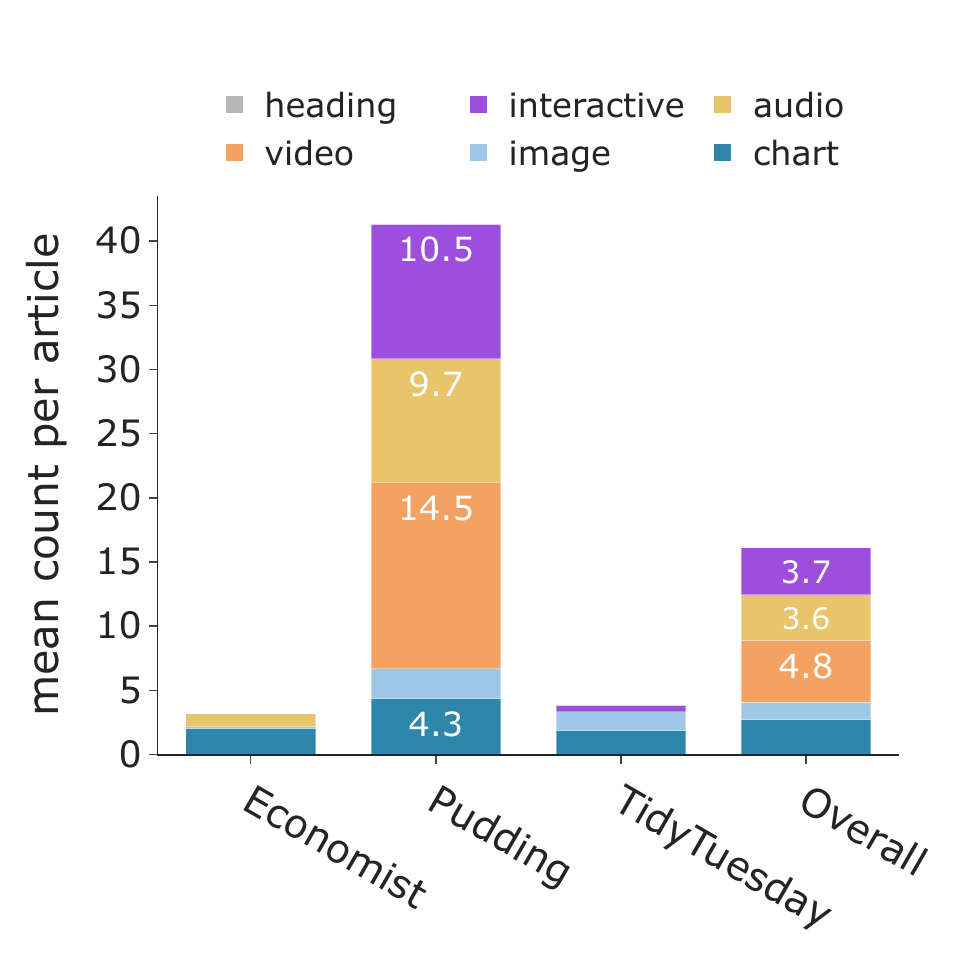}
        \caption{Articles made by human.}
        \label{fig:multimodal-human-bygroup}
    \end{subfigure}
    \caption{\textbf{Multimodal media asset distributions} (\eg~video, image, audio, interactive, etc) between \methodshort{} (left) and human (right).}
    \label{fig:media}
\end{figure}

Beyond text, every article may carry various multimedia assets, leading the article style to diverge sharply. 
We classify multimedia assets by six categories: heading (big short title), interactive, audio, video, image, and chart. 
As illustrated in Figure~\ref{fig:multimodal-agent-bygroup}, \methodshort{}'s media distribution is uniform across all three sources: it averages 13--14 assets per article and covers every modality in similar proportions. 
By contrast, Figure~\ref{fig:multimodal-human-bygroup} shows that human authors tune the kit to the publication: `Pudding' carries about 41 assets per article, rich in video, audio, and interactives, while `The Economist' and `TidyTuesday' stay near 3--4, almost all charts and images. 
\textit{\methodshort{} robustly produces every modality across topics, whereas human designers vary their distribution substantially with editorial style.}

\subsubsection{Human studies as primary testbed}
\begin{figure*}[!h]
    \centering
    \begin{subfigure}[t]{0.32\textwidth}
        \centering
        \includegraphics[width=\linewidth]{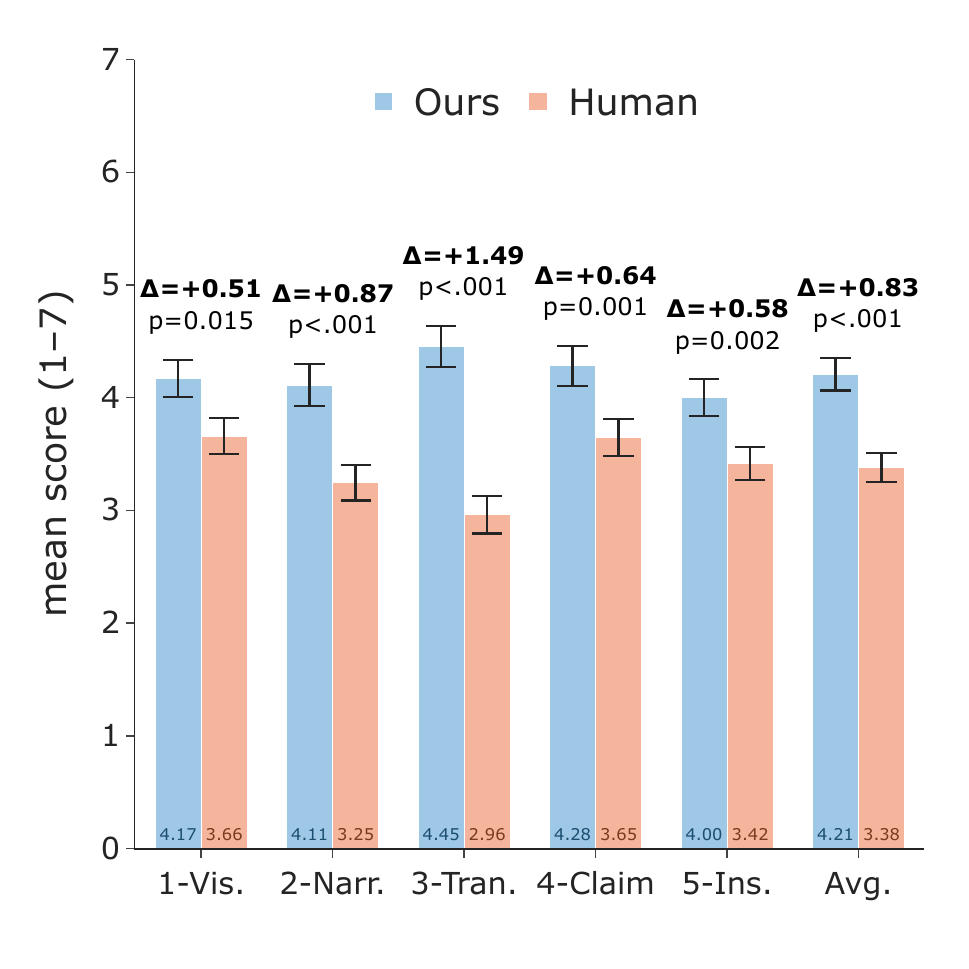}
        \caption{By rubric dimension.}
        \label{fig:judgehuman-dims}
    \end{subfigure}
    \hfill
    \begin{subfigure}[t]{0.32\textwidth}
        \centering
        \includegraphics[width=\linewidth]{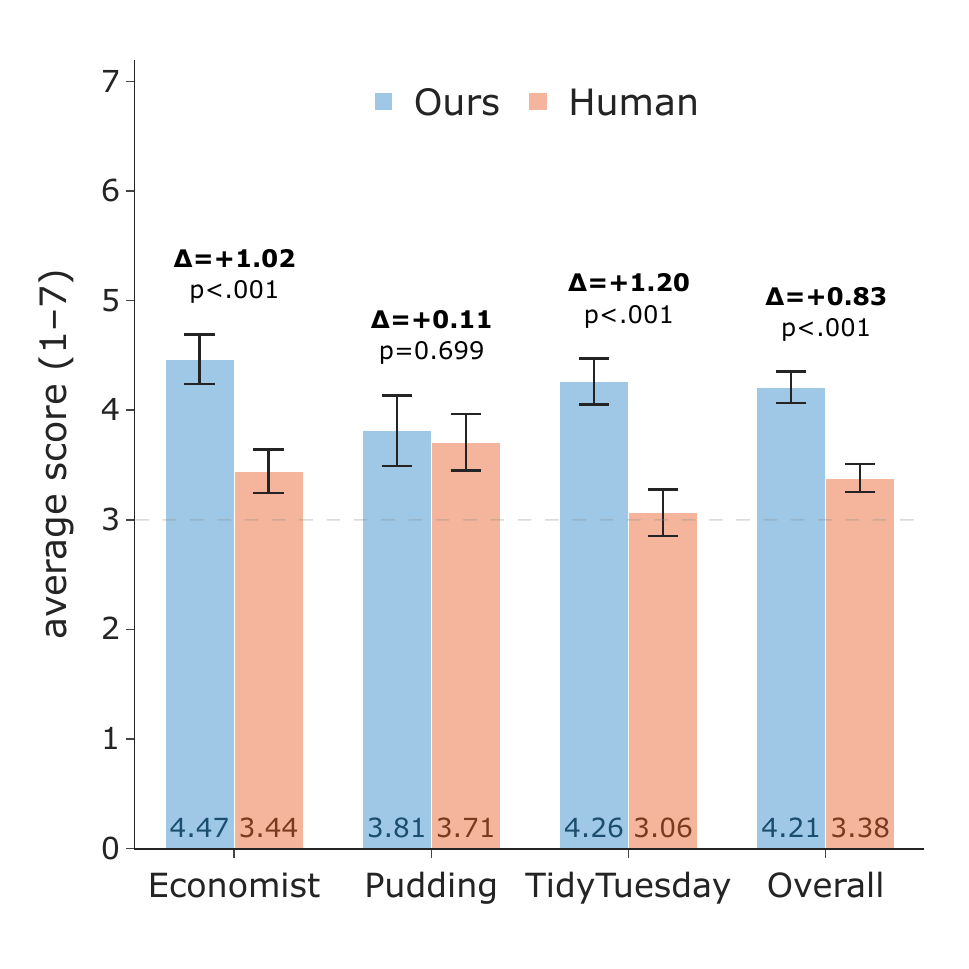}
        \caption{By source category.}
        \label{fig:judgehuman-bycat}
    \end{subfigure}
    \hfill
    \begin{subfigure}[t]{0.32\textwidth}
        \centering
        \includegraphics[width=\linewidth]{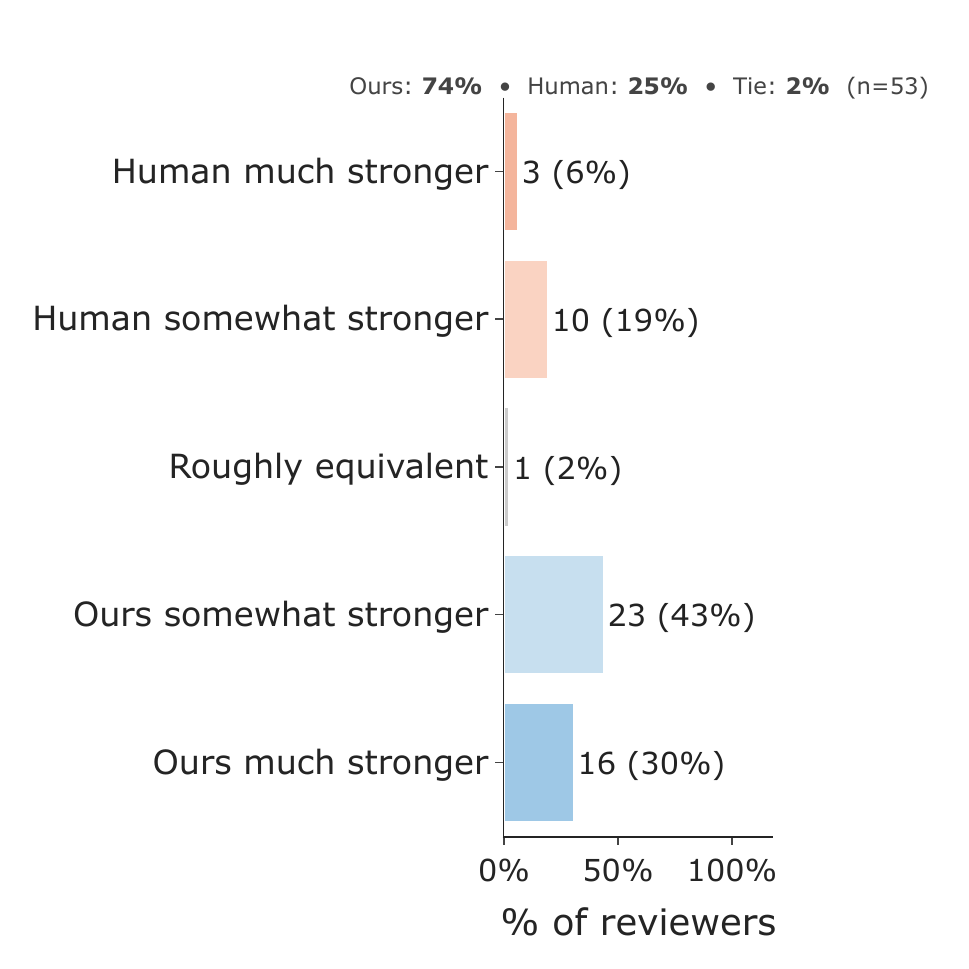}
        \caption{Overall pairwise preference.}
        \label{fig:judgehuman-pref}
    \end{subfigure}
    \caption{\textbf{Human evaluation ($n{=}53$ reviewers).} Scores are grouped by rubric dimension (a) and source category (b). Finally, reviewers were asked to choose the better article through pairwise comparisons (c).}
\label{fig:judgehuman}
\end{figure*}

\paragraph{\methodshort{} articles are appreciated by humans across various rubrics.}
Figure~\ref{fig:judgehuman-dims} reports per-dimension means across the 53 participants, with the agent ahead on all five axes (with an overall mean of $4.21$ for \methodshort{} and $3.38$ for humans). 
The largest gap is on “Transparency” ($+1.49$), a margin we attribute to the Inspector's per-sentence provenance and we provide an ablation in §\ref{sec:Inspector}; the smallest is on “Visual” ($+0.51$), which we further ablate next.

\paragraph{Analytical genres amplify the agent's advantage, while editorial scrollytelling narrows it.}
Figure~\ref{fig:judgehuman-bycat} shows the breakdown by source: \textit{Economist} ($\Delta{=}{+}1.02$, $p{<}.001$) and \textit{TidyTuesday} ($\Delta{=}{+}1.20$, $p{<}.001$) both clearly favour the agent, while \textit{Pudding} is a statistical tie. Pudding's long-form scrollytelling pieces are produced by art designer teams who spend weeks per article on bespoke design and a single committed thesis, an authorial investment the agent does not yet match. 
\textit{The agent performs best in genres where analytical framing matters more than authorial voice; in the most designer-curated genre, it merely matches human performance.}

\paragraph{The holistic preference is consistent with the rubric.}
Beyond the per-dimension scores, each reviewer also gave a single overall preference after seeing both versions. Figure~\ref{fig:judgehuman-pref} shows the result: of the $53$ reviewers, $39$ preferred \methodshort{}, $13$ preferred the human version, and $1$ ($2\%$) called it a tie. 
The holistic preference moves in the same direction as the per-dimension rubric, which suggests that the dimensions the rubric isolates (transparency, claim-data alignment, and so on) are also the ones reviewers weigh when forming an overall judgment. 

\subsubsection{Computer-use agent preserve rankings cost-efficiently}
\begin{figure*}[t]
    \centering
    \begin{subfigure}[t]{0.32\textwidth}
        \centering
        \includegraphics[width=\linewidth]{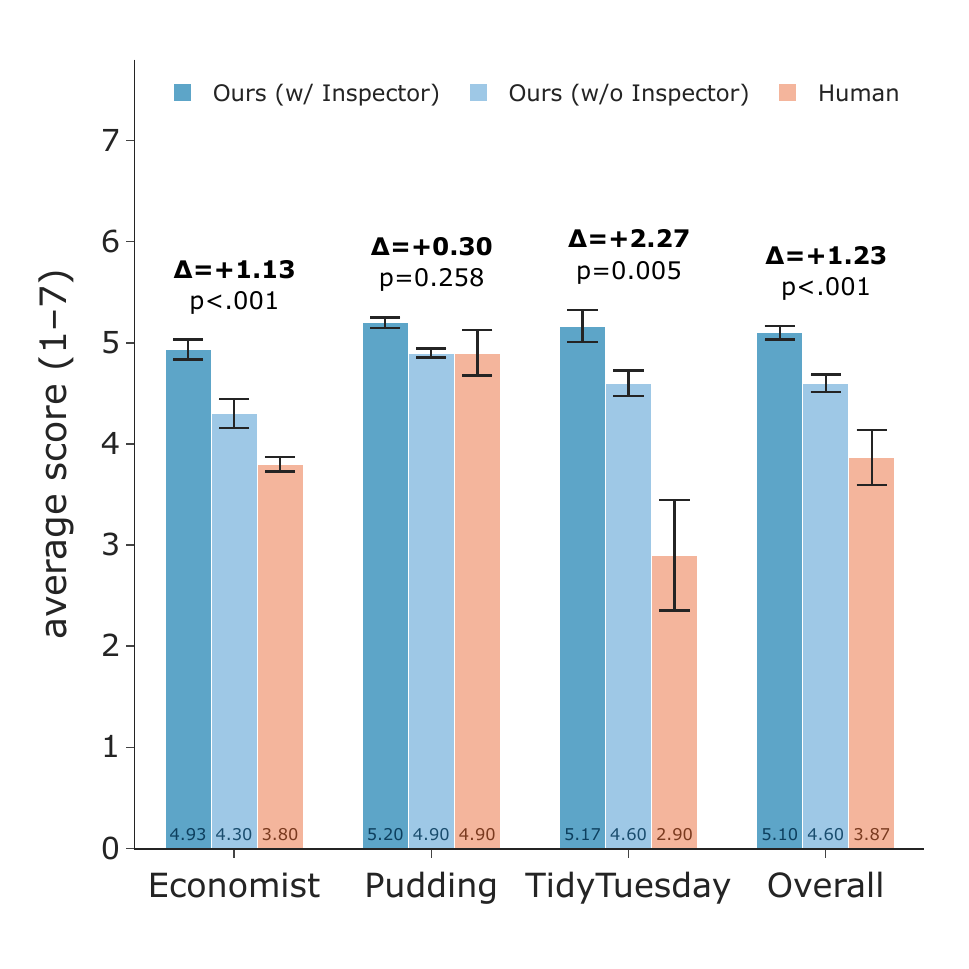}
        \caption{By source category.}
        \label{fig:judgevlmcodex-avg}
    \end{subfigure}
    \hfill
    \begin{subfigure}[t]{0.32\textwidth}
        \centering
        \includegraphics[width=\linewidth]{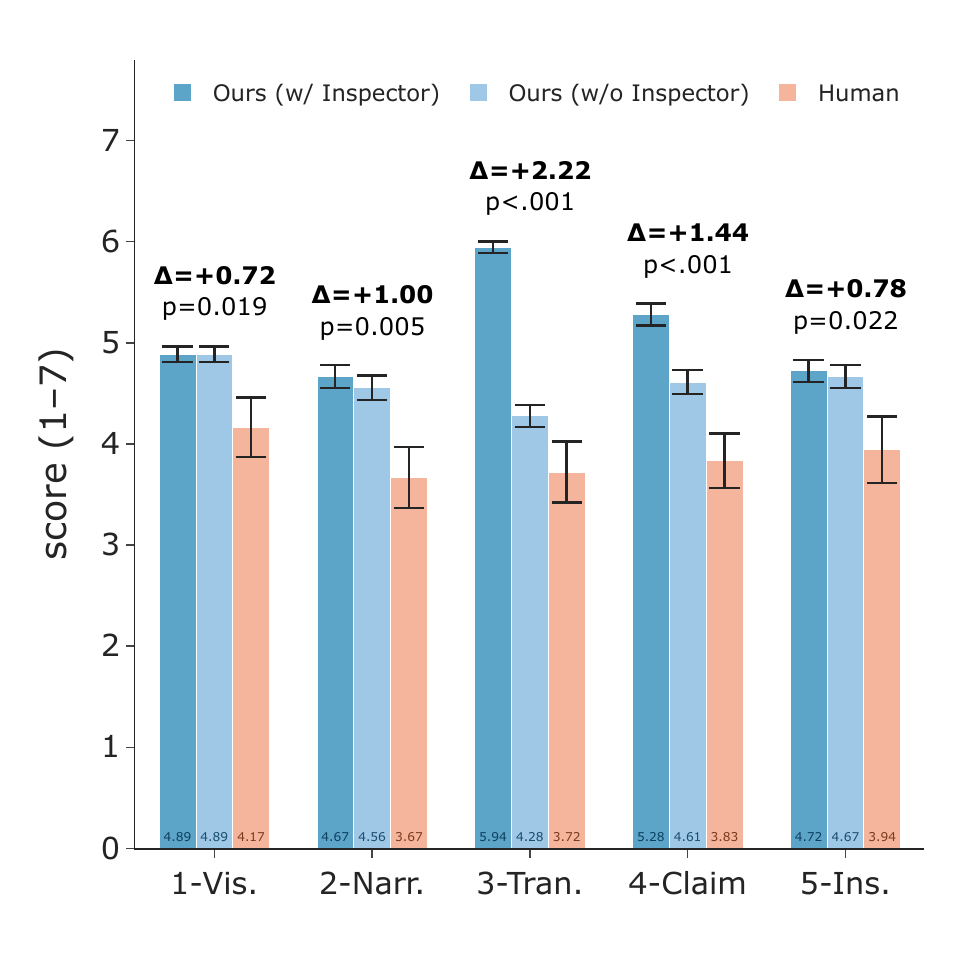}
        \caption{By rubric dimension.}
        \label{fig:judgevlmcodex-dims}
    \end{subfigure}
    \hfill
    \begin{subfigure}[t]{0.32\textwidth}
        \centering
        \includegraphics[width=\linewidth]{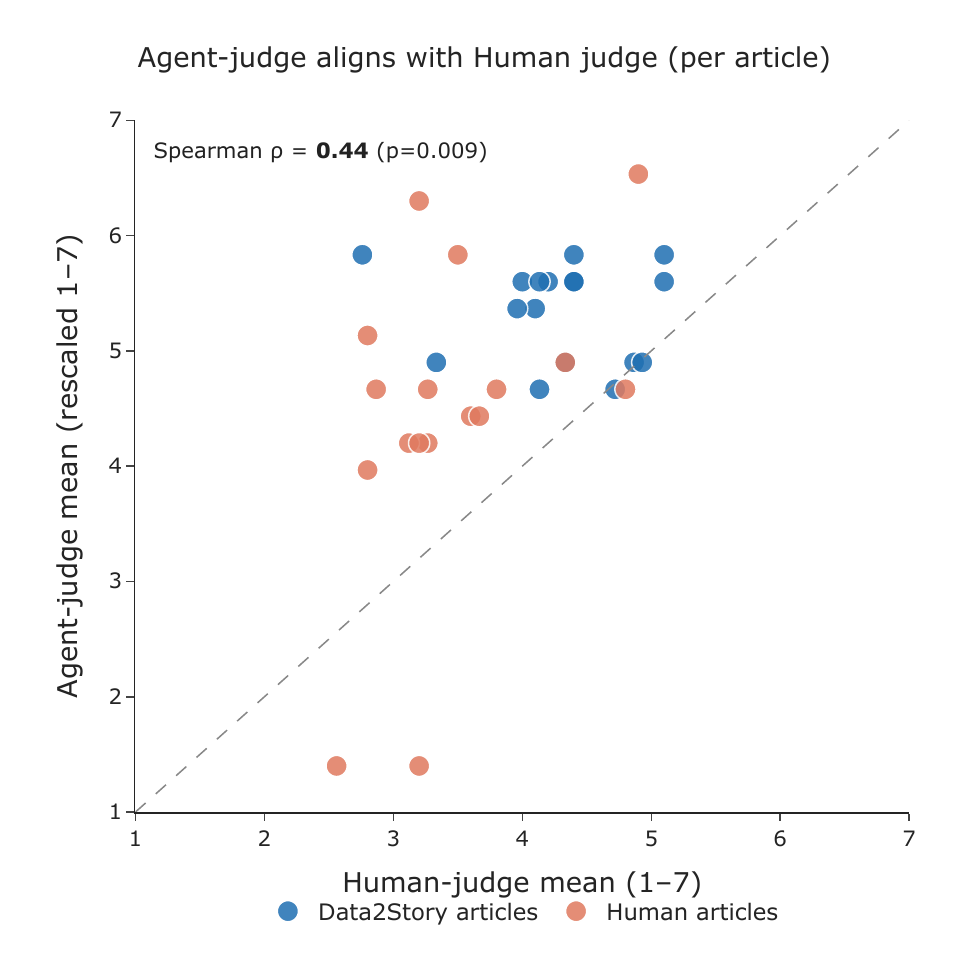}
        \caption{Agent judge aligns with human judge.}
        \label{fig:judgevlmcodex-agree}
    \end{subfigure}
    \caption{\textbf{Agent-as-judge evaluation.} Scores are compared across \methodshort{} articles with the Inspector, \methodshort{} articles without the Inspector, and human-written articles. Results are grouped by source category (a) and rubric dimension (b), with score distributions from agent-judge and human-judge compared in (c).}
    \label{fig:judgevlmcodex}
\end{figure*}

Figure~\ref{fig:judgevlmcodex-avg} reports the Agent judge's average score with ablation of the Inspector across the three sources. 
We treat the agent judge as a cost-efficient solution for ranking articles rather than a major quality signal; quality claims rest on the human study alone.
\paragraph{Transparency's Inspector lift is roughly $2.5{\times}$ the next-largest dimension and dwarfs the rest.}
With the Inspector off, the agent's overall mean is $4.60$ (human reference: $3.87$), and on \textit{Pudding} the two are identical ($4.90$ each), consistent with the human study. Opening the Inspector raises the overall mean to $5.10$, a further $\sim{}0.50$.
Figure~\ref{fig:judgevlmcodex-dims} breaks the same three conditions down by rubric dimension: the effect concentrates almost entirely on \emph{3-Transparency} ($4.28{\to}5.94$, $\Delta{=}{+}1.67$), with \emph{4-Claim} a distant second ($\Delta{=}{+}0.67$) and the remaining three dimensions barely shifting ($\Delta{\le}0.11$).
The Inspector thus buys a single dominant transparency channel plus a modest claim--data assist, with little spillover onto visual, narrative, or insight.

\paragraph{The agent judge preserves the human ranking at a fraction of the cost.}
A practical question is whether the cheaper agent judge can stand in for the 53-reviewer study on the same articles. The two judges rank articles together ($\rho{=}0.44$, $p{<}.01$;
Figure~\ref{fig:judgevlmcodex-agree}), and almost every point ($29{/}34$) sits above the $y{=}x$ line, so the agent keeps the human ordering while scoring both \methodshort{} and human articles higher in absolute terms. 
This show a moderate agreement. The possible reasons might be this limited by the computer-use agent interaction behaviors, such as human might focus on a particular area. 
While considering the low cost of agent automation, we remain see it usable stan-in for the ranking the human study produces.

\subsubsection{Verifiability analysis: auditability rather than factuality}
\begin{figure}[!h]
    \centering
    \begin{subfigure}[t]{0.32\textwidth}
        \centering
        \includegraphics[width=\linewidth]{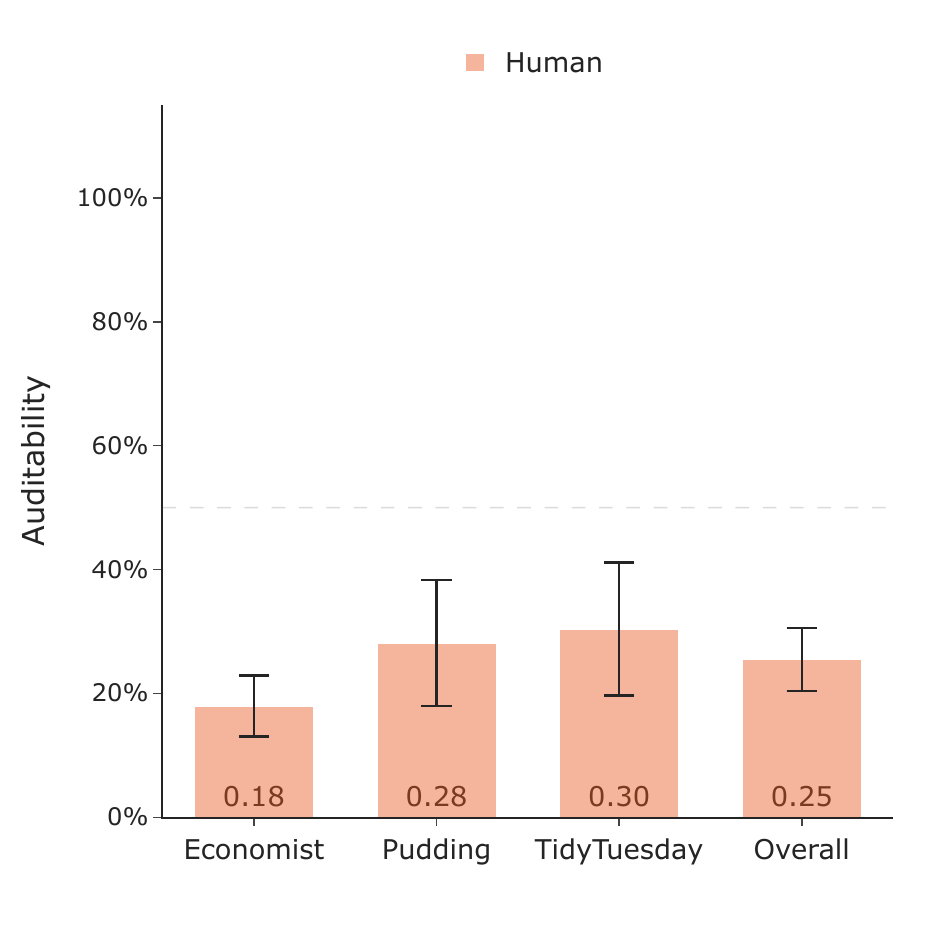}
        \caption{Human, per source.}
        \label{fig:repro-bycat-human}
    \end{subfigure}
    \begin{subfigure}[t]{0.32\textwidth}
        \centering
        \includegraphics[width=\linewidth]{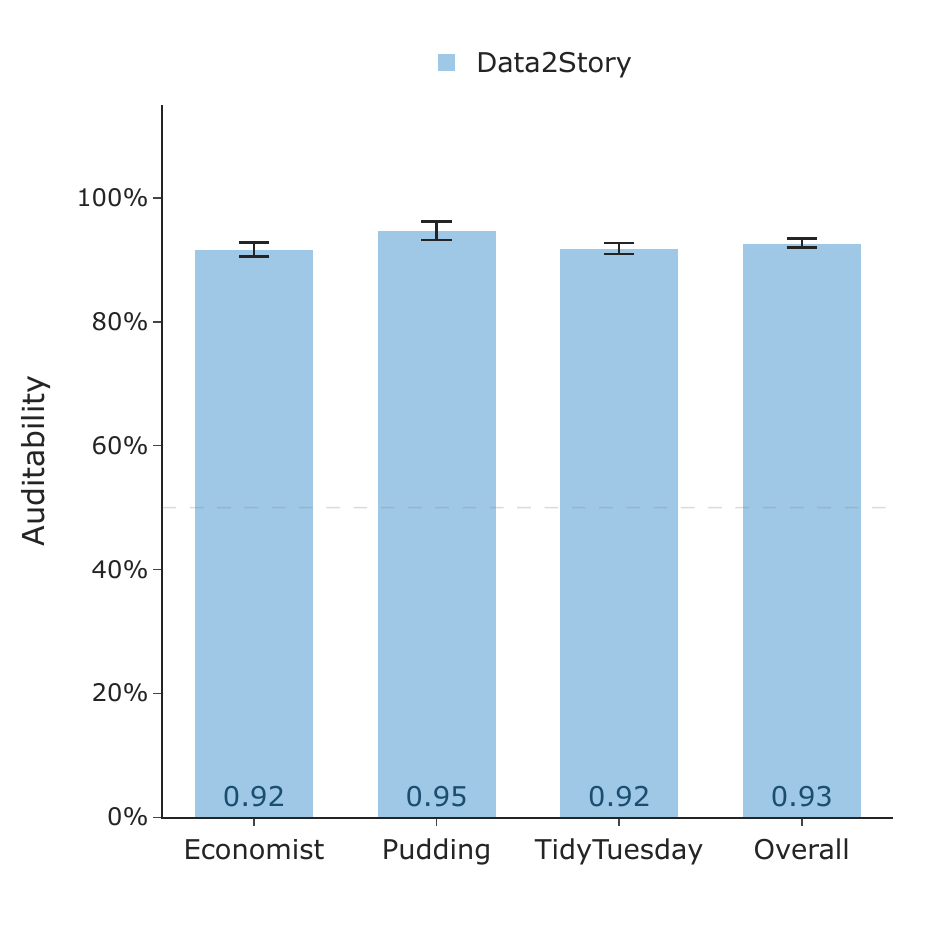}
        \caption{\methodshort, per source.}
        \label{fig:repro-bycat-ours}
    \end{subfigure}
    \begin{subfigure}[t]{0.32\textwidth}
        \centering
        \includegraphics[width=\linewidth]{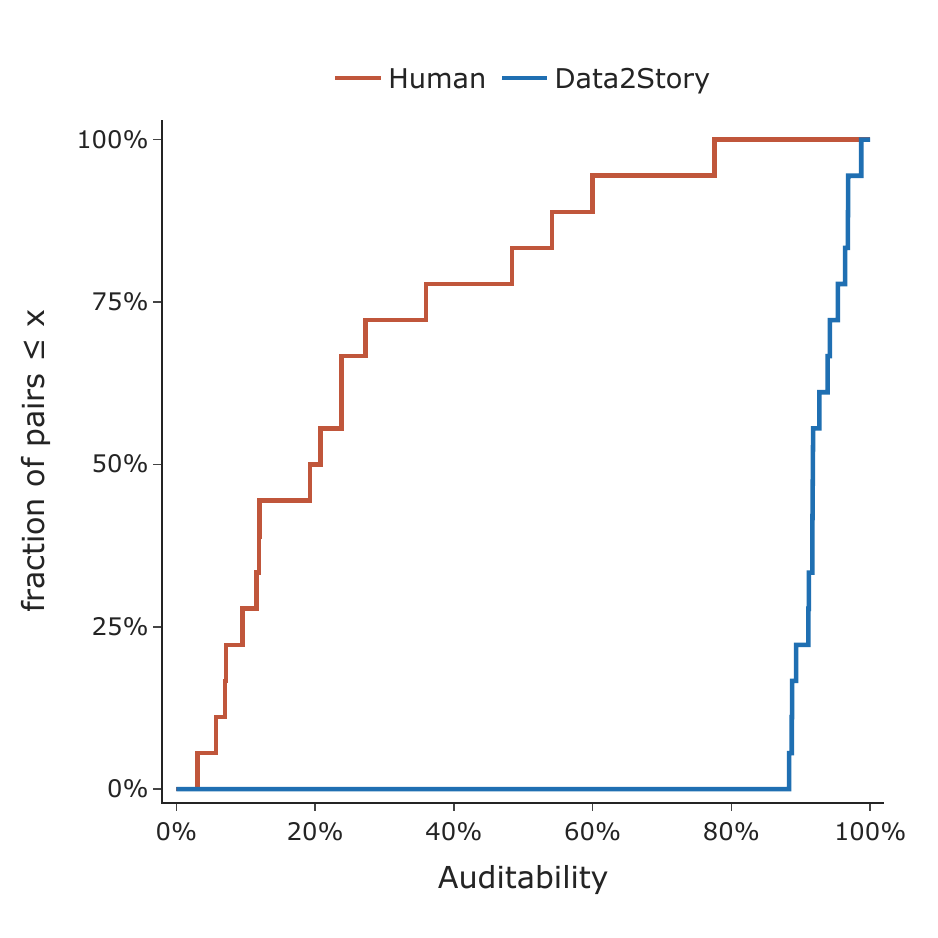}
        \caption{Empirical CDF over all 18 articles.}
        \label{fig:repro-ecdf}
    \end{subfigure}
    \caption{\textbf{Auditability between \methodshort-generated and human-written articles.} Per-source means with SEM error bars for human (a) and \methodshort{} (b); empirical CDF over all 18 articles (c).}
    \label{fig:repro}
\end{figure}

Figure~\ref{fig:repro} (a,b) reports machine-checkable provenance coverage across publication sources.
For \methodshort{} articles, 93\% of visible claims resolve to a traceable binding between the rendered text and its upstream evidence.
Human reference articles ship no accompanying code, so by construction most claims cannot be checked this way; the Codex verifier has to guess at a plausible reproduction on its own from the raw data and the published text alone.
Thus, text-only audit recovers such a binding for 25\% of claims. This makes sense as human-written statements are written for general readers and rarely attach a line of code or a traceable source to each claim, whereas our Inspector question bank probes for exactly that.
It is worth noting that \textbf{they measure whether a claim carries a verifiable provenance trail, not whether it is factually correct.}
The gap therefore reflects the availability of machine-checkable provenance, not the quality of human journalism.


All three sources show a wide and significant gap. The gap is narrowest for {Economist}, whose briefing-style articles foreground more explicit and standard numerical analysis. This makes the likely human findings easier to anticipate, because many insights can be inferred from visible statistics, comparisons, and trends. In contrast, the gap is widest for {Pudding}, whose scrollytelling pieces often center on creative editorial ideas and qualitative framing rather than enumerated sub-population statistics. These ideas are less formulaic and therefore harder to guess from the pre-registered questions alone. 

Figure~\ref{fig:repro-ecdf} shows the empirical distribution of article auditability. \methodshort{} articles concentrate in a tight band near the top of the auditability axis, while the human distribution is spread more broadly. This suggests that the auditability that \methodshort{} offers is largely the pipeline itself rather than of any particular reference article; claims are bound to upstream evidence by construction, whereas in human’s articles the same kind of binding only appears when the author chose to expose it.

\begin{figure}[!h]
    \centering
    \begin{subfigure}[t]{0.28\linewidth}
        \centering
        \includegraphics[width=\linewidth]{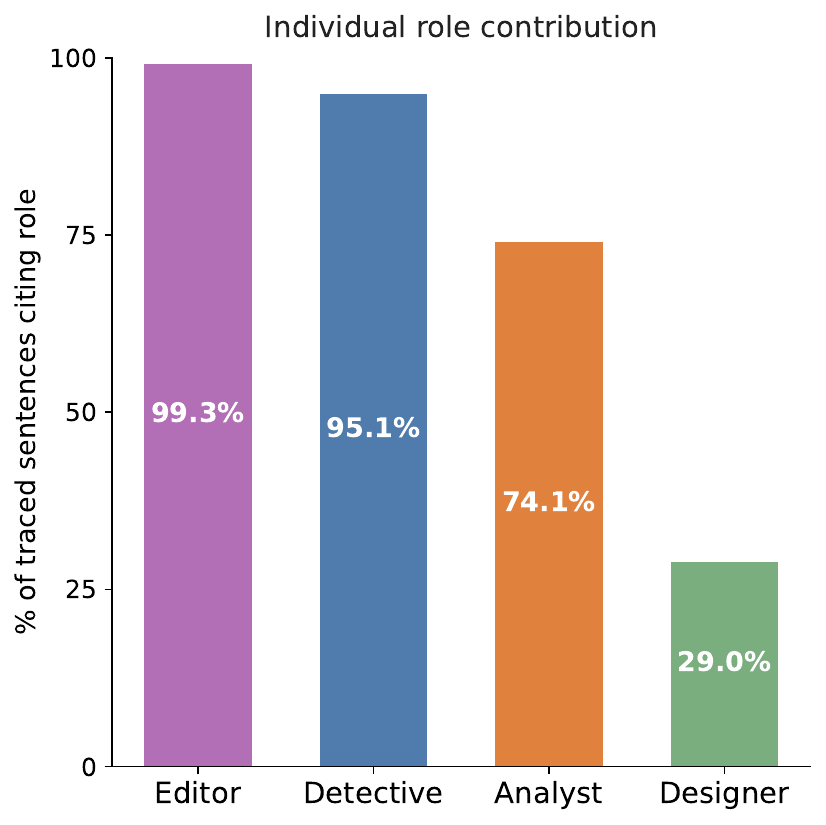}
        \caption{Individual role contribution.}
    \label{fig:Inspector-diff}
    \end{subfigure}
    \begin{subfigure}[t]{0.32\linewidth}
        \centering
        \includegraphics[width=\linewidth]{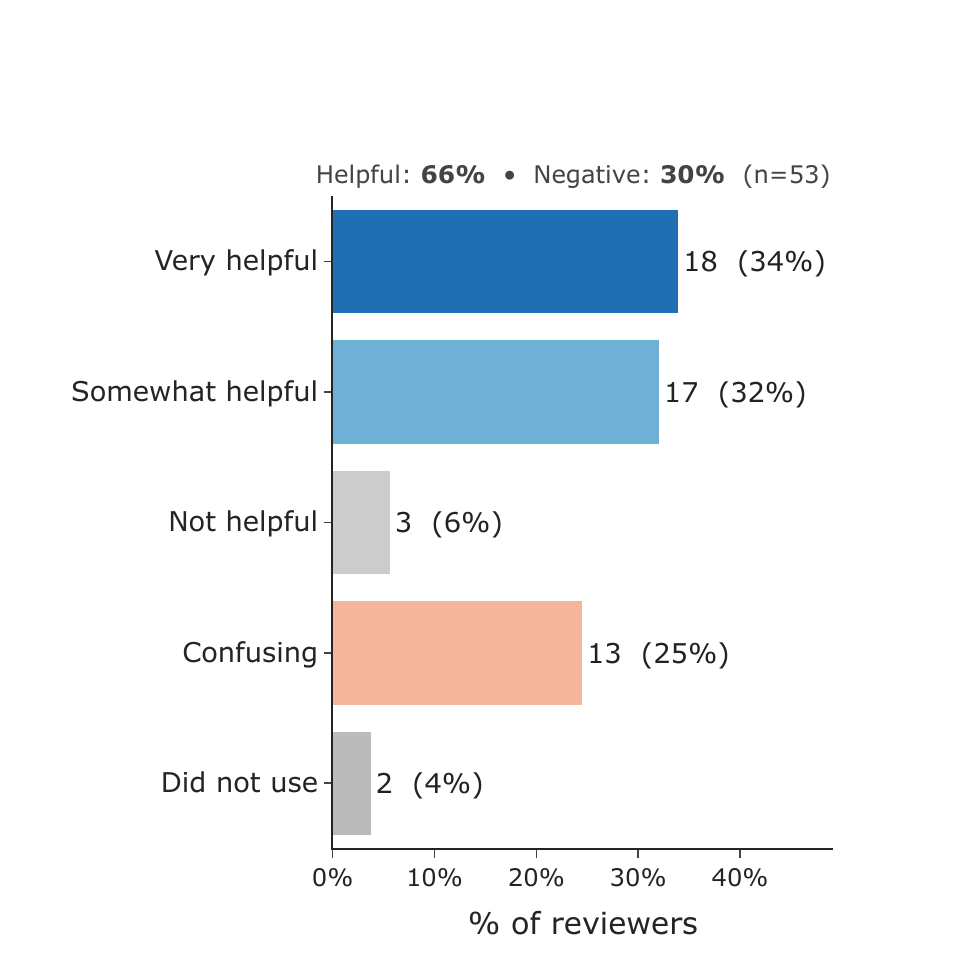}
    \caption{Human participants' votes on whether the Inspector was useful.}
    \label{fig:Inspector-selfreport}
    \end{subfigure}
    \begin{subfigure}[t]{0.32\linewidth}
        \centering
        \includegraphics[width=\linewidth]{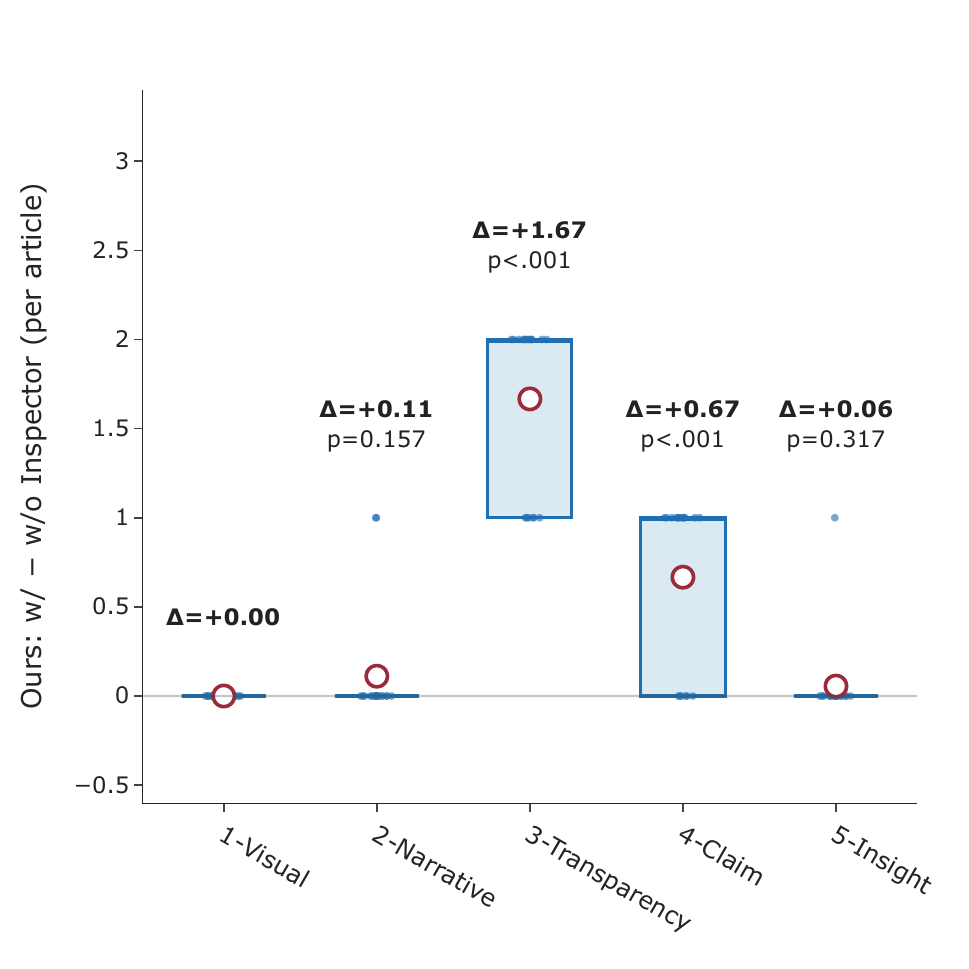}
        \caption{Within-article Inspector effect on different rubrics.}
        \label{fig:Inspector-paneleffect}
    \end{subfigure}
    \caption{\textbf{Analysis of Inspector effect.} Human participants' usefulness ratings of the Inspector (a), and Agent judges inspector-related gains across rubric dimensions (b).}
    \label{fig:Inspector}
\end{figure}

\subsubsection{Analysis of different roles}
\label{sec:Inspector}
The inspector subagent exposes the per-sentence provenance produced by four different roles: {Detective} (sourcing), {Analyst} (computation), {Designer} (chart authoring), and {Editor} (storytelling). Figure~\ref{fig:Inspector-diff} reports per-role coverage across all articles: \textit{Editor} $99.3\%$, \textit{Detective} $95.1\%$, \textit{Analyst} $74.1\%$ and \textit{Designer} $29.0\%$. These shares reflect each role's working character more than the data itself: Editor and Detective participate in nearly every traced sentence --- every claim is storyboarded, and Detective's search-heavy sourcing names at least one external reference; Analyst adds computation to roughly three quarters of sentences (the quantitative subset); Designer is selective, anchoring visual assets in about a third.

\textbf{Effect of Inspector.}
Figure~\ref{fig:Inspector-selfreport} reports how the $n{=}53$ reviewers experienced the Inspector, which attaches per-claim provenance information including analyst notes, code-line references, and source datasets---to the rendered article. 
Overall, $66\%$ of participants found the Inspector helpful for forming their evaluations, only $3 (6\%)$ rated it not helpful, whereas $25\%$ found it distracting. The main concern was that the provenance traces were sometimes dense and complex, linking each claim to multiple scripts, quoted sources, and data references.

Figure~\ref{fig:Inspector-paneleffect} isolates the Inspector behavioral effect: same article, same computer-use agent judge, only difference is whether the Inspector is open. The within-pair lift concentrates on \emph{3-Transparency} ($\Delta{=}{+}1.67$, paired, $p{<}.001$). This further highlights that \textit{opening the Inspector lifts mainly the transparency rubric}.

\subsection{Qualitative assessment: where human did better?}
We examine individual examples from each publication source, comparing the articles produced by \method~with those written by human journalists. 
This qualitative view surfaces values that our numerical experiments miss. 
Across the paired set, the human edge shows up in three recurring forms: the editorial angle, the creative design, and the informative presentation.

\textbf{(i) Editorial Angle.}
The human advantage we could not close is the angle that comes from outside the data. The Repair Cafés reporter (Table~\ref{tab:caseT4-opening}) frames repair as a matter of accountability, attributing failure to manufacturers that build ``phones, cars and tractors'' so that mechanics cannot access ``diagnostic tools or broken parts'' without them. Such a claim is reported rather than computed: it rests on expert testimony and on outside knowledge the dataset never holds. Working only from the table, \methodshort{} can rank what breaks (knives are saved far more often than printers), but it leaves the cause to the reader. This is the qualitative face of our coverage finding (\S\ref{sec:res}). Across the paired set the agent recovers only about half of the human's editorial angle, because the other half lives in reporting it cannot reach.

\textbf{(ii) Creative Design.}
On the \textit{Pudding} pieces, human teams invest weeks of bespoke interaction the agent does not attempt. The Stand-Up Comedy article (Table~\ref{tab:caseP1-opening}) turns the transcript into the interface: ``every line'' of Ali Wong's special is on the page, and each laugh is marked beside its line as a circle scaled to its length. For the same material, \methodshort{} links out to a static YouTube thumbnail and summarises the set in standard charts. A similar gap appears in the Internet Boy Band Database (Table~\ref{tab:caseP4-opening}), which plays as an audio-visual jukebox of all $55$ acts, its hand-animated members morphing from band to band as each act's song plays; the agent re-tells the same history in static charts behind click-to-play embeds. The numbers survive, but not the crafted experience built around them.

\textbf{(iii) Informative Presentation.}
Even in a single static figure, human designers carry more meaning per frame. The space-race chart (Table~\ref{tab:caseE1-opening}) sets state against commercial launch providers on one timeline and folds in a second variable for free: each band is shaded lighter where launches failed, with an annotation explaining why the Soviet count runs so high. Its satellites ``lasted only a year and a half on average, compared with nine years for their American counterparts.'' \methodshort{} distributes the same material across many single-variable charts, so no one figure carries the story. The football-managers chart (Table~\ref{tab:caseE3-opening}) overlays managers and star players on one axis, placing Messi and Ronaldo at the high end where the chart's own annotation reads: ``Star players' impact can reach ten points per season. Managers rarely add more than two.'' Our agent plots only the managers, and the comparison the headline promises never appears.

These cases show that the human contribution should not be understated. \methodshort{} leads on coverage, analysis, and auditable transparency, yet the reported angle, together with the hand-built craft behind a design or a chart, remains a human strength.
\section{Discussion}
\label{sec:con}

We introduced \method, a multi-agent framework that orchestrates specialised roles into a single virtual newsroom for data journalism assistant. \methodshort{} contributes two properties absent from prior approaches: an evidence-traceable Inspector that binds each number, quote, and asset to a specific code line or reference, and multimodal generative storytelling in which the agent reasons about audience needs before deploying sub-agents and tools that fit both the data and the reader. 
Across 18 samples paired with expert references, \methodshort{} receives favourable ratings from 53 human participants and from computer-use agent judges on both rubric dimensions and side-by-side preference, with the Inspector specifically improving data and method transparency. 

We position \methodshort{} as a collaborator for human journalists: 
(i) agent-generated articles can augment the newsroom workflow by contributing creative multimodal assets and an auditability dimension that is rarely formalised. 
(ii) Beyond augmenting existing coverage, \methodshort{} opens a complementary path: surfacing specialised or niche datasets that human journalists rarely have the bandwidth to investigate in depth, turning overlooked data into accessible, verifiable stories.
We hope this work moves us toward a trustworthy agentic data system.
\textbf{Limitations.}
\methodshort{} so far runs fully automatically. A more reliable design would let it take human feedback and adjust in the loop -- exploring whether an agent can interpret reader feedback and revise as professionally as a journalist.
Meanwhile, our multimodal storytelling offers a new perspective on presenting data, yet the depth a human writer brings to the written angle should not be underestimated.
Article generation may also raise concerns around fairness and bias. We leave these points to future work.

\textbf{The Uniqueness of Journalism.}
(i) Agents might produce hallucinated claims, and it remains hard to ensure zero hallucination. Human journalists may not provide explicit auditability, but their work is grounded in verified reporting.
(ii) Beyond article production, human journalists often speak to sources, contact organisations, and seek responses from affected parties. This cannot be replicated.

{
    \small
    \bibliographystyle{unsrt} 
    \bibliography{main}

@String(ICLR = {Int. Conf. Learn. Represent.})

@String(ICLR  = {ICLR})

@article{aiscientist,
  title={The AI scientist: Towards fully automated open-ended scientific discovery},
  author={Lu, Chris and Lu, Cong and Lange, Robert Tjarko and Foerster, Jakob and Clune, Jeff and Ha, David},
  journal={arXiv preprint arXiv:2408.06292},
  year={2024}
}

@article{aiscientistv2,
  title={The AI scientist-v2: Workshop-level automated scientific discovery via agentic tree search},
  author={Yamada, Yutaro and Lange, Robert Tjarko and Lu, Cong and Hu, Shengran and Lu, Chris and Foerster, Jakob and Clune, Jeff and Ha, David},
  journal={arXiv preprint arXiv:2504.08066},
  year={2025}
}

@article{dsbench,
  title={DSBench: How Far Are Data Science Agents from Becoming Data Science Experts?},
  author={Jing, Liqiang and Huang, Zhehui and Wang, Xiaoyang and Yao, Wenlin and Yu, Wenhao and Ma, Kaixin and Zhang, Hongming and Du, Xinya and Yu, Dong},
  journal={arXiv preprint arXiv:2409.07703},
  year={2024}
}

@article{dsgym,
  title={{DSGym}: A Holistic Framework for Evaluating and Training Data Science Agents},
  author={Nie, Fan and Wang, Junlin and Hua, Harper and Bianchi, Federico and Kwon, Yongchan and Qi, Zhenting and Queen, Owen and Zhu, Shang and Zou, James},
  journal={arXiv preprint arXiv:2601.16344},
  year={2026}
}

@article{scienceagentbench,
  title={Scienceagentbench: Toward rigorous assessment of language agents for data-driven scientific discovery},
  author={Chen, Ziru and Chen, Shijie and Ning, Yuting and Zhang, Qianheng and Wang, Boshi and Yu, Botao and Li, Yifei and Liao, Zeyi and Wei, Chen and Lu, Zitong and others},
  journal={arXiv preprint arXiv:2410.05080},
  year={2024}
}

@article{mlebench,
  title={Mle-bench: Evaluating machine learning agents on machine learning engineering},
  author={Chan, Jun Shern and Chowdhury, Neil and Jaffe, Oliver and Aung, James and Sherburn, Dane and Mays, Evan and Starace, Giulio and Liu, Kevin and Maksin, Leon and Patwardhan, Tejal and others},
  journal={arXiv preprint arXiv:2410.07095},
  year={2024}
}

@article{mlagentbench,
  title={Mlagentbench: Evaluating language agents on machine learning experimentation},
  author={Huang, Qian and Vora, Jian and Liang, Percy and Leskovec, Jure},
  journal={arXiv preprint arXiv:2310.03302},
  year={2023}
}

@book{handbook,
  title={The data journalism handbook: How journalists can use data to improve the news},
  author={Gray, Jonathan and Chambers, Lucy and Bounegru, Liliana},
  year={2012},
  publisher={" O'Reilly Media, Inc."}
}

@book{datajournalism,
  title={The Data Journalism Handbook 2: Towards a Critical Data Practice},
  author={Bounegru, Liliana and Gray, Jonathan},
  year={2021},
  publisher={Amsterdam University Press},
  url          = {https://s3.eu-central-1.amazonaws.com/datajournalismcom/handbooks/The-Data-Journalism-Handbook-2.pdf}  
}

@article{deepanalyze,
  title={{DeepAnalyze}: Agentic Large Language Models for Autonomous Data Science},
  author={Zhang, Shaolei and Fan, Ju and Fan, Meihao and Li, Guoliang and Du, Xiaoyong},
  journal={arXiv preprint arXiv:2510.16872},
  year={2025}
}

@inproceedings{datainterpreter,
  title={{Data Interpreter}: An {LLM} Agent for Data Science},
  author={Hong, Sirui and Lin, Yizhang and Liu, Bang and Liu, Bangbang and Wu, Binhao and Zhang, Ceyao and Wei, Chenxing and Li, Danyang and Chen, Jiaqi and Zhang, Jiayi and Wang, Jinlin and Zhang, Li and Zhang, Lingyao and Yang, Min and Zhuge, Mingchen and Guo, Taicheng and Zhou, Tuo and Tao, Wei and Tang, Robert and Lu, Xiangtao and Zheng, Xiawu and Liang, Xinbing and Fei, Yaying and Cheng, Yuheng and Ni, Yongxin and Gou, Zhibin and Xu, Zongze and Luo, Yuyu and Wu, Chenglin},
  booktitle={Findings of the Association for Computational Linguistics: ACL 2025},
  pages={19796--19821},
  year={2025}
}

@misc{openaidr,
  title={Introducing deep research},
  author={{OpenAI}},
  howpublished={\url{https://openai.com/index/introducing-deep-research/}},
  year={2025}
}

@inproceedings{lida,
  title={{LIDA}: A Tool for Automatic Generation of Grammar-Agnostic Visualizations and Infographics using Large Language Models},
  author={Dibia, Victor},
  booktitle={Proceedings of the 61st Annual Meeting of the Association for Computational Linguistics (Volume 3: System Demonstrations)},
  pages={113--126},
  year={2023}
}

@article{narrative,
  title={Narrative visualization: Telling stories with data},
  author={Segel, Edward and Heer, Jeffrey},
  journal={IEEE transactions on visualization and computer graphics},
  volume={16},
  number={6},
  pages={1139--1148},
  year={2010},
  publisher={IEEE}
}

@inproceedings{design2code,
  title={Design2code: Benchmarking multimodal code generation for automated front-end engineering},
  author={Si, Chenglei and Zhang, Yanzhe and Li, Ryan and Yang, Zhengyuan and Liu, Ruibo and Yang, Diyi},
  booktitle={Proceedings of the 2025 Conference of the Nations of the Americas Chapter of the Association for Computational Linguistics: Human Language Technologies (Volume 1: Long Papers)},
  pages={3956--3974},
  year={2025}
}

@inproceedings{mmsearch,
  title={{MMSearch}: Benchmarking the Potential of Large Models as Multi-modal Search Engines},
  author={Jiang, Dongzhi and Zhang, Renrui and Guo, Ziyu and Wu, Yanmin and Lei, Jiayi and Qiu, Pengshuo and Lu, Pan and Chen, Zehui and Song, Guanglu and Gao, Peng and Liu, Yu and Li, Chunyuan and Li, Hongsheng},
  booktitle={International Conference on Learning Representations (ICLR)},
  year={2025},
  note={arXiv:2409.12959}
}

@inproceedings{datanarrative,
  title={{DataNarrative}: Automated Data-Driven Storytelling with Visualizations and Texts},
  author={Islam, Mohammed Saidul and Laskar, Md Tahmid Rahman and Parvez, Md Rizwan and Hoque, Enamul and Joty, Shafiq},
  booktitle={Proceedings of the 2024 Conference on Empirical Methods in Natural Language Processing (EMNLP)},
  pages={19253--19286},
  year={2024},
  note={arXiv:2408.05346}
}

@inproceedings{mindsearch,
  title={{MindSearch}: Mimicking Human Minds Elicits Deep {AI} Searcher},
  author={Chen, Zehui and Liu, Kuikun and Wang, Qiuchen and Liu, Jiangning and Zhang, Wenwei and Chen, Kai and Zhao, Feng},
  booktitle={International Conference on Learning Representations (ICLR)},
  year={2025},
  note={arXiv:2407.20183}
}

@article{deepresearcher,
  title={{DeepResearcher}: Scaling Deep Research via Reinforcement Learning in Real-world Environments},
  author={Zheng, Yuxiang and Fu, Dayuan and Hu, Xiangkun and Cai, Xiaojie and Ye, Lyumanshan and Lu, Pengrui and Liu, Pengfei},
  journal={arXiv preprint arXiv:2504.03160},
  year={2025}
}

@article{publicagent,
  title={{PublicAgent}: Multi-Agent Design Principles From an {LLM}-Based Open Data Analysis Framework},
  author={Montazeri, Sina and Feng, Yunhe and Sha, Kewei},
  journal={arXiv preprint arXiv:2511.03023},
  year={2025}
}

@inproceedings{journalistplan,
  title={Do {LLMs} Plan Like Human Writers? Comparing Journalist Coverage of Press Releases with {LLMs}},
  author={Spangher, Alexander and Peng, Nanyun and Gehrmann, Sebastian and Dredze, Mark},
  booktitle={Proceedings of the 2024 Conference on Empirical Methods in Natural Language Processing (EMNLP)},
  year={2024}
}

@inproceedings{datadirector,
  title={From Data to Story: Towards Automatic Animated Data Video Creation with {LLM}-based Multi-Agent Systems},
  author={Shen, Leixian and Li, Haotian and Wang, Yun and Qu, Huamin},
  booktitle={IEEE VIS Workshop on Generative AI for Data Storytelling (Gen4DS)},
  year={2024},
  note={arXiv:2408.03876}
}

@book{ware2004information,
  title     = {Information Visualization: Perception for Design},
  author    = {Ware, Colin},
  edition   = {2nd},
  year      = {2004},
  publisher = {Morgan Kaufmann},
  address   = {San Francisco, CA}
}

@book{tufte2001visual,
  title     = {The Visual Display of Quantitative Information},
  author    = {Tufte, Edward R.},
  edition   = {2nd},
  year      = {2001},
  publisher = {Graphics Press},
  address   = {Cheshire, CT}
}

@article{segel2010narrative,
  title   = {Narrative Visualization: Telling Stories with Data},
  author  = {Segel, Edward and Heer, Jeffrey},
  journal = {IEEE Transactions on Visualization and Computer Graphics},
  volume  = {16},
  number  = {6},
  pages   = {1139--1148},
  year    = {2010},
  publisher = {IEEE}
}

@book{cairo2016truthful,
  title={The truthful art: Data, charts, and maps for communication},
  author={Cairo, Alberto},
  year={2016},
  publisher={New Riders}
}

@book{knaflic2025storytelling,
  title={Storytelling with data: A data visualization guide for business professionals},
  author={Knaflic, Cole Nussbaumer},
  year={2025},
  publisher={John Wiley \& Sons}
}

@article{north2006toward,
  title   = {Toward Measuring Visualization Insight},
  author  = {North, Chris},
  journal = {IEEE Computer Graphics and Applications},
  volume  = {26},
  number  = {3},
  pages   = {6--9},
  year    = {2006},
  publisher = {IEEE}
}

@article{diakopoulos2015algorithmic,
  title   = {Algorithmic Accountability: Journalistic Investigation of Computational Power Structures},
  author  = {Diakopoulos, Nicholas},
  journal = {Digital Journalism},
  volume  = {3},
  number  = {3},
  pages   = {398--415},
  year    = {2015},
  publisher = {Taylor \& Francis}
}

@article{gelman2013garden,
  title   = {The Garden of Forking Paths: Why Multiple Comparisons Can Be a Problem, Even When There Is No ``Fishing Expedition'' or ``p-Hacking'' and the Research Hypothesis Was Posited Ahead of Time},
  author  = {Gelman, Andrew and Loken, Eric},
  journal = {Department of Statistics, Columbia University},
  year    = {2013},
  note    = {Unpublished manuscript}
}

@incollection{grice1975logic,
  title     = {Logic and Conversation},
  author    = {Grice, H. Paul},
  booktitle = {Syntax and Semantics, Vol. 3: Speech Acts},
  editor    = {Cole, Peter and Morgan, Jerry L.},
  pages     = {41--58},
  year      = {1975},
  publisher = {Academic Press},
  address   = {New York}
}

@article{drtulu,
  title={Dr tulu: Reinforcement learning with evolving rubrics for deep research},
  author={Shao, Rulin and Asai, Akari and Shen, Shannon Zejiang and Ivison, Hamish and Kishore, Varsha and Zhuo, Jingming and Zhao, Xinran and Park, Molly and Finlayson, Samuel G and Sontag, David and others},
  journal={arXiv preprint arXiv:2511.19399},
  year={2025}
}

@article{openresearcher,
  title={Openresearcher: A fully open pipeline for long-horizon deep research trajectory synthesis},
  author={Li, Zhuofeng and Jiang, Dongfu and Ma, Xueguang and Zhang, Haoxiang and Nie, Ping and Zhang, Yuyu and Zou, Kai and Xie, Jianwen and Zhang, Yu and Chen, Wenhu},
  journal={arXiv preprint arXiv:2603.20278},
  year={2026}
}

@article{coda,
  title={Coda: Agentic systems for collaborative data visualization},
  author={Chen, Zichen and Chen, Jiefeng and Arik, Sercan {\"O} and Sra, Misha and Pfister, Tomas and Yoon, Jinsung},
  journal={arXiv preprint arXiv:2510.03194},
  year={2025}
}

@article{browsecomp,
  title={Browsecomp: A simple yet challenging benchmark for browsing agents},
  author={Wei, Jason and Sun, Zhiqing and Papay, Spencer and McKinney, Scott and Han, Jeffrey and Fulford, Isa and Chung, Hyung Won and Passos, Alex Tachard and Fedus, William and Glaese, Amelia},
  journal={arXiv preprint arXiv:2504.12516},
  year={2025}
}

@article{rag,
  title={Retrieval-augmented generation for knowledge-intensive nlp tasks},
  author={Lewis, Patrick and Perez, Ethan and Piktus, Aleksandra and Petroni, Fabio and Karpukhin, Vladimir and Goyal, Naman and K{\"u}ttler, Heinrich and Lewis, Mike and Yih, Wen-tau and Rockt{\"a}schel, Tim and others},
  journal={Advances in neural information processing systems},
  volume={33},
  pages={9459--9474},
  year={2020}
}

@inproceedings{matplotagent,
  title={Matplotagent: Method and evaluation for llm-based agentic scientific data visualization},
  author={Yang, Zhiyu and Zhou, Zihan and Wang, Shuo and Cong, Xin and Han, Xu and Yan, Yukun and Liu, Zhenghao and Tan, Zhixing and Liu, Pengyuan and Yu, Dong and others},
  booktitle={Findings of the Association for Computational Linguistics: ACL 2024},
  pages={11789--11804},
  year={2024}
}

@article{cheng2025journalism,
  title={When Journalism meets AI: Risk or opportunity?},
  author={Cheng, Sophia},
  journal={Digital Government: Research and Practice},
  volume={6},
  number={1},
  pages={1--12},
  year={2025},
  publisher={ACM New York, NY}
}

@article{brigham2024developing,
  title={Developing Story: Case Studies of Generative AI's Use in Journalism},
  author={Brigham, Natalie Grace and Gao, Chongjiu and Kohno, Tadayoshi and Roesner, Franziska and Mireshghallah, Niloofar},
  journal={arXiv preprint arXiv:2406.13706},
  year={2024}
}

@inproceedings{spangher2025novel,
  title={A Novel Multi-Document Retrieval Benchmark: Journalist Source-Selection in Newswriting},
  author={Spangher, Alexander and Huang, Tenghao and Huang, Yiqin and Spangher, Lucas and Min, Sewon and Dredze, Mark},
  booktitle={Proceedings of the 4th International Workshop on Knowledge-Augmented Methods for Natural Language Processing},
  pages={180--204},
  year={2025}
}

@article{alshomary2026llms,
  title={LLMs as Science Journalists: Supporting Early-stage Researchers in Communicating Their Science to the Public},
  author={Alshomary, Milad and Li, Grace and Jangra, Anubhav and Hou, Yufang and McKeown, Kathleen and Muresan, Smaranda},
  journal={arXiv preprint arXiv:2601.05821},
  year={2026}
}

@article{ji2023survey,
  title={Survey of hallucination in natural language generation},
  author={Ji, Ziwei and Lee, Nayeon and Frieske, Rita and Yu, Tiezheng and Su, Dan and Xu, Yan and Ishii, Etsuko and Bang, Ye Jin and Madotto, Andrea and Fung, Pascale},
  journal={ACM computing surveys},
  volume={55},
  number={12},
  pages={1--38},
  year={2023},
  publisher={ACM New York, NY}
}

@article{zheng2023judging,
  title={Judging llm-as-a-judge with mt-bench and chatbot arena},
  author={Zheng, Lianmin and Chiang, Wei-Lin and Sheng, Ying and Zhuang, Siyuan and Wu, Zhanghao and Zhuang, Yonghao and Lin, Zi and Li, Zhuohan and Li, Dacheng and Xing, Eric and others},
  journal={Advances in neural information processing systems},
  volume={36},
  pages={46595--46623},
  year={2023}
}

@article{zhuge2024agent,
  title={Agent-as-a-judge: Evaluate agents with agents},
  author={Zhuge, Mingchen and Zhao, Changsheng and Ashley, Dylan and Wang, Wenyi and Khizbullin, Dmitrii and Xiong, Yunyang and Liu, Zechun and Chang, Ernie and Krishnamoorthi, Raghuraman and Tian, Yuandong and others},
  journal={arXiv preprint arXiv:2410.10934},
  year={2024}
}

@inproceedings{chen2024mllm,
  title={Mllm-as-a-judge: Assessing multimodal llm-as-a-judge with vision-language benchmark},
  author={Chen, Dongping and Chen, Ruoxi and Zhang, Shilin and Wang, Yaochen and Liu, Yinuo and Zhou, Huichi and Zhang, Qihui and Wan, Yao and Zhou, Pan and Sun, Lichao},
  booktitle={Forty-first International Conference on Machine Learning},
  year={2024}
}

@inproceedings{zhou2024webarena,
  title={Webarena: A realistic web environment for building autonomous agents},
  author={Zhou, Shuyan and Xu, Frank F and Zhu, Hao and Zhou, Xuhui and Lo, Robert and Sridhar, Abishek and Cheng, Xianyi and Ou, Tianyue and Bisk, Yonatan and Fried, Daniel and others},
  booktitle={International Conference on Learning Representations},
  volume={2024},
  pages={15585--15606},
  year={2024}
}

@misc{rusch2025aicitycouncil,
  author       = {Holly Rusch},
  title        = {Should {AI} Cover Your City Council Meeting? {P}revalence of {AI}-Generated Articles Summarizing Public Meetings Grows in {San Mateo County}},
  howpublished = {San Mateo Daily Journal},
  year         = {2025},
  url          = {https://www.smdailyjournal.com/news/local/should-ai-cover-your-city-council-meeting-prevalence-of-ai-generated-articles-summarizing-public-meetings/article_cb1c8474-a5b7-44d1-b57c-f5d22034f089.html},
  note         = {Accessed: 2026-06-08}
}

@article{datastorm,
  title={DataSTORM: Deep Research on Large-Scale Databases using Exploratory Data Analysis and Data Storytelling},
  author={Liu, Shicheng and Jiang, Yucheng and Farook, Sajid and Sanchez, Camila Nicollier and Pena, David Fernando Castro and Lam, Monica S},
  journal={arXiv preprint arXiv:2604.06474},
  year={2026}
}

@article{cohen2011computational,
  title   = {Computational Journalism},
  author  = {Cohen, Sarah and Hamilton, James T. and Turner, Fred},
  journal = {Communications of the ACM},
  volume  = {54},
  number  = {10},
  pages   = {66--71},
  year    = {2011},
  doi     = {10.1145/2001269.2001288}
}

@techreport{hamilton2009accountability,
  title       = {Accountability Through Algorithm: Developing the Field of
                 Computational Journalism},
  author      = {Hamilton, James T. and Turner, Fred},
  institution = {Center for Advanced Study in the Behavioral Sciences,
                 Stanford University},
  type        = {White Paper},
  year        = {2009}
}

@article{coddington2015clarifying,
  title     = {Clarifying Journalism's Quantitative Turn: A Typology for
               Evaluating Data Journalism, Computational Journalism, and
               Computer-Assisted Reporting},
  author    = {Coddington, Mark},
  journal   = {Digital Journalism},
  volume    = {3},
  number    = {3},
  pages     = {331--348},
  year      = {2015},
  doi       = {10.1080/21670811.2014.976400}
}

@book{diakopoulos2019automating,
  title     = {Automating the News: How Algorithms Are Rewriting the Media},
  author    = {Diakopoulos, Nicholas},
  publisher = {Harvard University Press},
  address   = {Cambridge, MA},
  year      = {2019}
}

@article{diakopoulos2017algorithmic,
  title     = {Algorithmic Transparency in the News Media},
  author    = {Diakopoulos, Nicholas and Koliska, Michael},
  journal   = {Digital Journalism},
  volume    = {5},
  number    = {7},
  pages     = {809--828},
  year      = {2017}
}

@techreport{diakopoulos2024generative,
  title       = {Generative {AI} in Journalism: The Evolution of Newswork and
                 Ethics in a Generative Information Ecosystem},
  author      = {Diakopoulos, Nicholas and Cools, Hannes and Li, Charlotte and
                 Helberger, Natali and Kung, Ernest and Rinehart, Aimee},
  institution = {Associated Press},
  month       = apr,
  year        = {2024},
  doi         = {10.13140/RG.2.2.31540.05765}
}
}

\clearpage

\beginappendix{
    \section{Model Settings}
\label{sec:app:setup}

\method~is based on Claude-code \texttt{opus-4.7}.
We detail the tools employed in the Designer role. We use OpenRouter as the unified provider for all generative models, as summarized in Table~\ref{tab:abilities}.

\begin{table}[!h]
    \centering
    \caption{Generative capabilities and the OpenRouter API model backing each tool.}
    \label{tab:abilities}
    \begin{tabular}{@{}lll@{}}
      \toprule
      \textbf{Tool} & \textbf{Modality} & \textbf{API Model (version)} \\
      \midrule
      \texttt{openrouter-text2image}   & Text $\rightarrow$ Image  & \texttt{openai/gpt-5.4-image-2} \\
      \texttt{openrouter-text2video}   & Text $\rightarrow$ Video  & \texttt{bytedance/seedance-2.0} \\
      \texttt{openrouter-image2video}  & Image $\rightarrow$ Video & \texttt{google/veo-3.1-fast} \\
      \texttt{openrouter-text2music}   & Text $\rightarrow$ Audio  & \texttt{google/lyria-3-pro-preview} \\
      \texttt{openrouter-embeddings}   & Text $\rightarrow$ Vector & \texttt{qwen/qwen3-embedding-8b} \\
      \bottomrule
    \end{tabular}
\end{table}

In Human-agent angle coverage, we use OpenAI’s \texttt{text-embedding-3-small} for retrieval similarity calculation, then use \texttt{gpt-4o-mini} to decide matching.

In Computer-use agent as judge experiments, we use the OpenAI’s browser-use \texttt{gpt-5.5-xhigh}.
\section{Additional Analysis}

\begin{figure}[!h]
    \centering
    \includegraphics[width=0.55\linewidth]{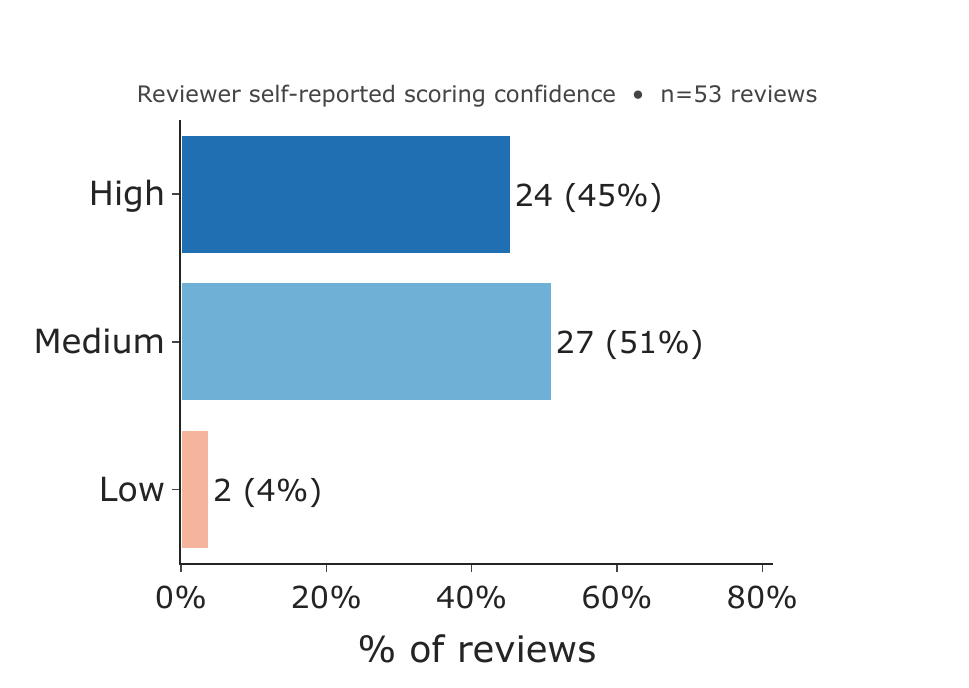}
    \caption{\textbf{Reviewer self-reported scoring confidence ($n{=}53$ reviews).}
    The single available self-assessment is each reviewer's confidence in their own dimension scores. Confidence is \textbf{High} in $45\%$ ($24$) and \textbf{Medium} in $51\%$ ($27$) of reviews, with only $4\%$ ($2$) \textbf{Low}. \textit{$96\%$ of reviews were submitted at Medium-or-High self-confidence, indicating reviewers generally felt able to apply the rubric}; this is a task-confidence proxy and \textbf{not} a measure of domain expertise.}
    \label{fig:judgehuman-level}
\end{figure}

\textbf{Reviewers confidence}
In Figure~\ref{fig:judgehuman-level}, we report the reviewers' self-reported scoring confidence to characterise the evaluation panel. Because the recruitment instrument did not elicit reviewer expertise in journalism, statistics, or the subject areas; We instead use each reviewer's confidence in their own dimension scores as the closest available self-assessment. $96\%$ of reviews ($51$ of $53$) were submitted at Medium-or-High confidence (High $45\%$, Medium $51\%$), with only $4\%$ Low, indicating that \textit{reviewers generally felt able to apply the rubric} even though the panel's domain expertise was not measured.

\begin{figure*}[!h]
    \centering
    \includegraphics[width=\linewidth]{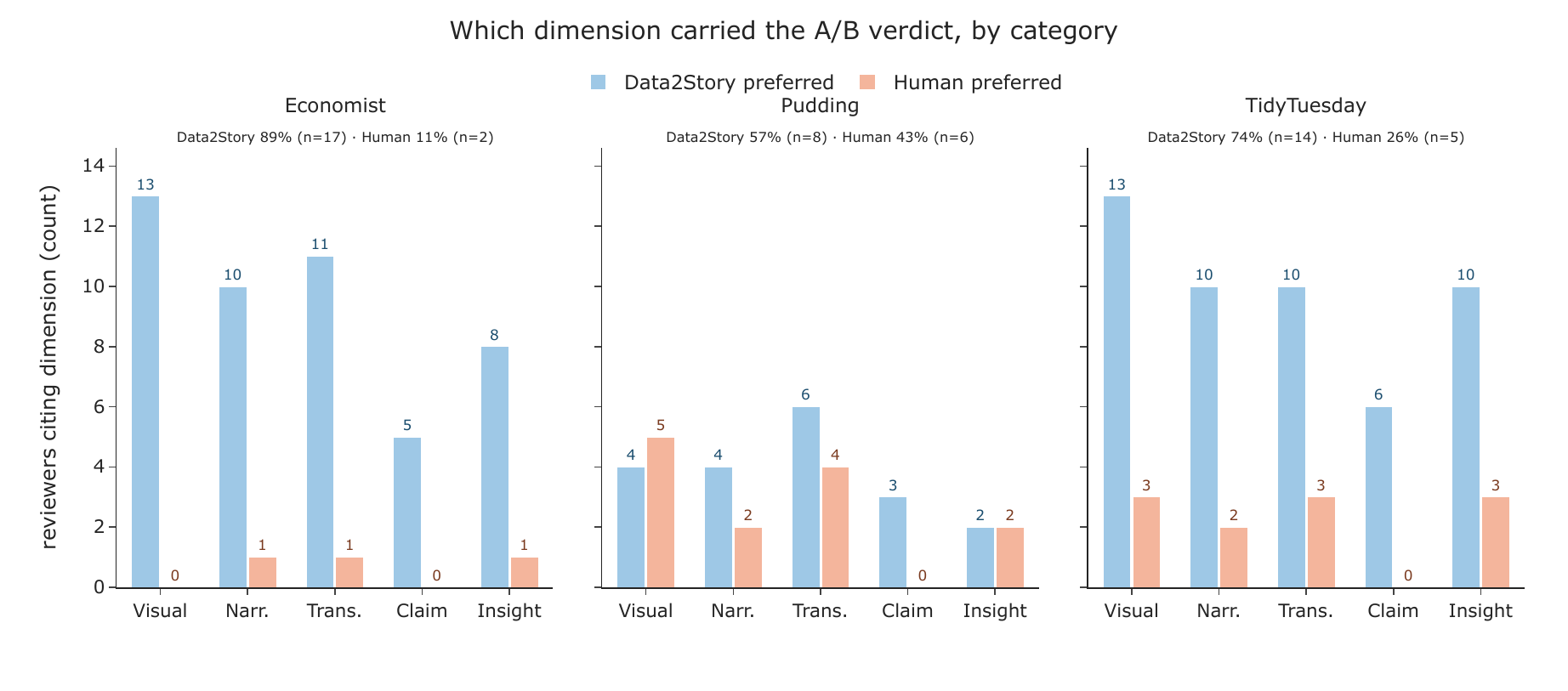}
    \caption{\textbf{Which rubric dimension carried the A/B verdict, by source category.} For each category, reviewers are split by their overall preference (ties dropped) into Data2Story-preferred (blue) and Human-preferred (orange). 
    }
    \label{fig:judgehuman-whycat}
\end{figure*}

\subsection{Which dimension carries the A/B verdict}
In Figure~\ref{fig:judgehuman-whycat} we pair each reviewer's overall A/B verdict with the dimensions they flag as separating the two versions (reviewers may select more than one), split by winner within each source category. Across all three categories reviewers prefer the agent, yet the deciding axis is visual craft and exposed provenance, almost never claim grounding. The Data2Story version is the majority choice everywhere: Economist $89\%$ ($n{=}17$ vs.\ $2$), TidyTuesday $74\%$ ($n{=}14$ vs.\ $5$), and Pudding $57\%$ ($n{=}8$ vs.\ $6$), the genre where the human stays closest. Among reviewers who pick the agent, the two dimensions cited most as the source of the gap are `Visual' (Economist $76\%$, TidyTuesday $93\%$) and `Transparency', the single most consistent agent-edge axis ($65\%/75\%/71\%$ across the three categories), which we read as the inspector panel's per-sentence provenance. `Claim--Data Alignment' is essentially never the deciding axis: it is the least-cited dimension for the agent ($29\%/38\%/43\%$) and is cited by $0\%$ of every human-prefer group.

Pudding is the lone exception, and it is exceptional on exactly the visual axis: it is the only category whose human-prefer reviewers cite `Visual' most ($83\%$, the highest single bar for either human group), matching the per-dimension Pudding result where the humans' one distinguishable edge is bespoke visual craft. Even there the agent keeps a transparency-cited majority ($75\%$), so the picture is consistent: the agent wins on exposed procedural credibility, the human's only foothold is visual design, and substantive `Claim' alignment never separates the two.

\section{Rubric Evaluation Scoring Standard}
In this section, we present the detailed scoring standard used in our rubric evaluation, which applies to both the human study and the agent judges. For each of the five dimensions, we provide detailed instructions for scores ranging from 1 to 7, where a score of 3 serves as our typical default.

\label{sec:app:rubric}

\definecolor{promptframe}{HTML}{5B9BD5}   
\definecolor{promptback}{HTML}{EAF3FC}    
\definecolor{prompttitle}{HTML}{1F4E79}   
\tcbset{
  promptbox/.style={
    breakable, enhanced,
    colback=promptback, colframe=promptframe,
    boxrule=1pt, arc=2.5mm,
    coltitle=white, colbacktitle=promptframe,
    fonttitle=\bfseries, fontupper=\normalsize,
    left=3.5mm, right=3.5mm, top=3mm, bottom=3mm,
  }
}
\providecommand{\promptsec}[1]{%
  \par\medskip{\color{prompttitle}\bfseries\large #1}\par\smallskip}
\providecommand{\promptsub}[1]{%
  \par\smallskip{\color{prompttitle}\bfseries #1}\par\smallskip}
\providecommand{\promptnote}[1]{%
  \par\smallskip\begin{quote}\itshape #1\end{quote}\par}

\begin{tcolorbox}[promptbox, title={Rubric Evaluation scoring guideline}]

\promptsec{Dimension 1 --- Visual Design}
\textbf{What it evaluates}: the artifact as a visual product --- palette,
typography, layout, whitespace, fit between chart type and the claim it
supports, overall polish.
\begin{center}
\small
\begin{tabularx}{\linewidth}{cX}
\toprule
Score & Description \\
\midrule
\textbf{7} & Indistinguishable from the best work published in the past year by top editorial graphics desks. Every chart is the optimal encoding for its claim. At least one design choice is so well-targeted that you can describe what trade-off it earns. \\
\textbf{6} & Coherent design system \textbf{plus} at least one design move you've rarely seen executed this well --- name it specifically. No wrong chart types, no missing labels or legends. \\
\textbf{5} & Cohesive design with intentional palette, typographic hierarchy, and chart selection. At most one minor named weakness. \\
\textbf{4} & Polished but with two or three specific weaknesses you can point to (busy chart, label collision, palette drift, missing in-figure legend). No chart is the wrong tool for its claim. \\
\textbf{3} & Looks like a competent person used a charting library with sensible defaults. Nothing offensive, nothing memorable. \textbf{Typical score.} \\
\textbf{2} & At least one of: mismatched chart type, jarring palette, truncated/missing labels, cluttered layout, visible default-styling residue. \\
\textbf{1} & Reads as raw library output or a database UI. Aesthetically mishandled to the point of impeding reading. \\
\bottomrule
\end{tabularx}
\end{center}

\promptsec{Dimension 2 --- Narrative \& Pacing}
\textbf{What it evaluates}: the author as a tour guide --- hook, ordering,
rhythm, ending.
\begin{center}
\small
\begin{tabularx}{\linewidth}{cX}
\toprule
Score & Description \\
\midrule
\textbf{7} & A piece you remember the shape of years later. Hook, layered progression, \textbf{and} a structural move (reframe, inversion, reveal) that recasts everything before it. The closing line bends back to the opening. The argument has a one-sentence summary that is itself non-obvious. \\
\textbf{6} & Strong arc with a clear structural move and a closing reframe that wasn't already spoiled by the standfirst. \\
\textbf{5} & Clear arc, real ending, at most one minor lull. \\
\textbf{4} & Logical ordering, real ending, two or three minor lulls or one stretched section. \\
\textbf{3} & Three-act competence: setup $\rightarrow$ data $\rightarrow$ conclusion. Readable but unsurprising in shape; no structural move. \textbf{Typical score.} \\
\textbf{2} & Sections organized by topic, not by argument. Reader has to stitch meaning together. \\
\textbf{1} & No through-line. A list of facts or charts; the artifact does not act as a guide. \\
\bottomrule
\end{tabularx}
\end{center}

\promptnote{\textbf{For non-narrative artifacts (datasets, tools)}: score on
whatever framing exists (intro, README, landing copy). Bare dataset with title
only $\rightarrow$ 1. README that motivates use without an arc $\rightarrow$ 2.
README with a clear analytic arc $\rightarrow$ 3. Higher requires explicit
narrative scaffolding.}

\promptsec{Dimension 3 --- Data \& Method Transparency}
\textbf{What it evaluates}: procedural credibility. Four things to look for:
\begin{enumerate}
\item \textbf{Sources cited specifically} --- named dataset or URL, not ``public data''
\item \textbf{Methodology described} --- definitions, transformations, exclusions
\item \textbf{Data accessible} --- link, download, repo, or in-page table
\item \textbf{Limitations acknowledged concretely} --- specific percentages, specific exclusion rules
\end{enumerate}
\begin{center}
\small
\begin{tabularx}{\linewidth}{cX}
\toprule
Score & Description \\
\midrule
\textbf{7} & Fully replicable \textbf{and} audit-grade. All four components present; multiple are unusually deep (line-level code references, public repo, processed-data table, dedicated methods note). A motivated outsider could replicate without contacting the author. \\
\textbf{6} & Replicable in principle. All four components present \textbf{and} at least one is unusually deep. \\
\textbf{5} & All four components present; methodology and caveats are concrete with specific numbers/exclusions; data access at least gestured at (e.g., named CSVs). \\
\textbf{4} & Three of four components, with sources and methodology mandatory among them. Caveats present but light. \\
\textbf{3} & Sources cited, methodology gestured at briefly, no data access, caveats minimal. \textbf{Industry-typical.} \\
\textbf{2} & Sources mentioned without specifics (``government data''); no methodology; no caveats. \\
\textbf{1} & No sources, no methods, claims of unclear origin. \\
\bottomrule
\end{tabularx}
\end{center}

\promptsub{Inspector panel rule for this dimension}
\begin{itemize}
\item \textbf{Panel \texttt{not present}}: rate based on the normal reading surface only.
\item \textbf{Panel \texttt{disabled}}: only count sources, methods, and caveats visible in the \textbf{normal reading surface} (prose, footnotes, in-page methodology sections, etc.). Do not credit anything that requires the \faSearch{} button.
\item \textbf{Panel \texttt{enabled}}: you may credit panel contents, \textbf{but only if you actually completed the exploration in \S1.4}. Reference specific card numbers and tags you saw, not just the card's surface text.
\end{itemize}

\promptsec{Dimension 4 --- Claim--Data Alignment}
\textbf{What it evaluates}: substantive credibility --- are claims actually
supported by the data?
\begin{center}
\small
\begin{tabularx}{\linewidth}{cX}
\toprule
Score & Description \\
\midrule
\textbf{7} & Surgical. Every quantitative claim is bounded by what the data can support; every place a reader might overinterpret is explicitly defused; uncertainty is quantified where relevant; chart encodings are unambiguous; correlation is never dressed as causation. \\
\textbf{6} & Precisely scoped, with at least one explicit sensitivity check or robustness gesture (alternative cut, what-if, named confound addressed quantitatively). \\
\textbf{5} & Precisely scoped. Quantitative claims bounded; major confounders named in prose; encodings unambiguous; no causal overreach. \\
\textbf{4} & Solid, with one or two minor stretches where an adjective or annotation slightly exceeds the data. Nothing structurally misleading. \\
\textbf{3} & Most claims supported; some loose framing; no major distortions. \textbf{Typical score.} \\
\textbf{2} & Several unsupported claims \textbf{or} at least one misleading visual element (truncated axis without flag, dual axis implying correlation, cherry-picked window, missing baseline). \\
\textbf{1} & Charts illustrate unproven claims; clear cherry-picking; encoding is materially deceptive. \\
\bottomrule
\end{tabularx}
\end{center}

\promptsec{Dimension 5 --- Insight Value}
\textbf{What it evaluates}: whether the reader gains a non-trivial cognitive
update.
\begin{center}
\small
\begin{tabularx}{\linewidth}{cX}
\toprule
Score & Description \\
\midrule
\textbf{7} & Field-shaping for a domain reader. A finding or reframe a specialist would cite. Updates a prior held by people who already know the topic. The reader can articulate the change in one sentence weeks later. \\
\textbf{6} & Updates a domain prior. Counterintuitive finding, structural pattern not previously legible, or reframe that makes a familiar topic newly readable. \\
\textbf{5} & Sharpens a domain intuition meaningfully. Quantifies something a specialist suspected but had not seen pinned down at this resolution. \\
\textbf{4} & Sharpens a lay intuition. Quantifies or sharpens something the lay reader vaguely believed; goes beyond surface summary. \\
\textbf{3} & Precise common knowledge. Well-supported, but the conclusion is what an informed reader would have predicted before reading. \textbf{Typical score.} \\
\textbf{2} & Awareness, not judgment. Reader leaves knowing the data exists; no new judgment is formed. \\
\textbf{1} & No thesis. Restates conventional wisdom or has no claim at all. \\
\bottomrule
\end{tabularx}
\end{center}

\promptnote{\textbf{Quick test}: if the takeaway is just ``this dataset
exists,'' cap at 2. If the reader doesn't reach synthesis or evaluation, cap at
3. If the update is meaningful only to a lay reader (not a specialist), cap at
5.}
\end{tcolorbox}

\begin{table}[H]
\centering
\small
\renewcommand{\arraystretch}{1.25}
\caption{The Economist: The space race is dominated by new contenders}
\label{tab:caseE1-opening}
\begin{tabularx}{\linewidth}{@{}p{1.8cm} X@{}}
\toprule
\textbf{Data} & {The Great Launch Inversion — 1957 to 2018} \\
\midrule
\textbf{Category} & Audio Artifact + Visual Artifact \\
\midrule
\textbf{Human}~\href{https://www.economist.com/graphic-detail/2018/10/18/the-space-race-is-dominated-by-new-contenders}{[link]} &
Title: \emph{The space race is dominated by new contenders}
\par\smallskip
{\centering\includegraphics[width=0.5\linewidth]{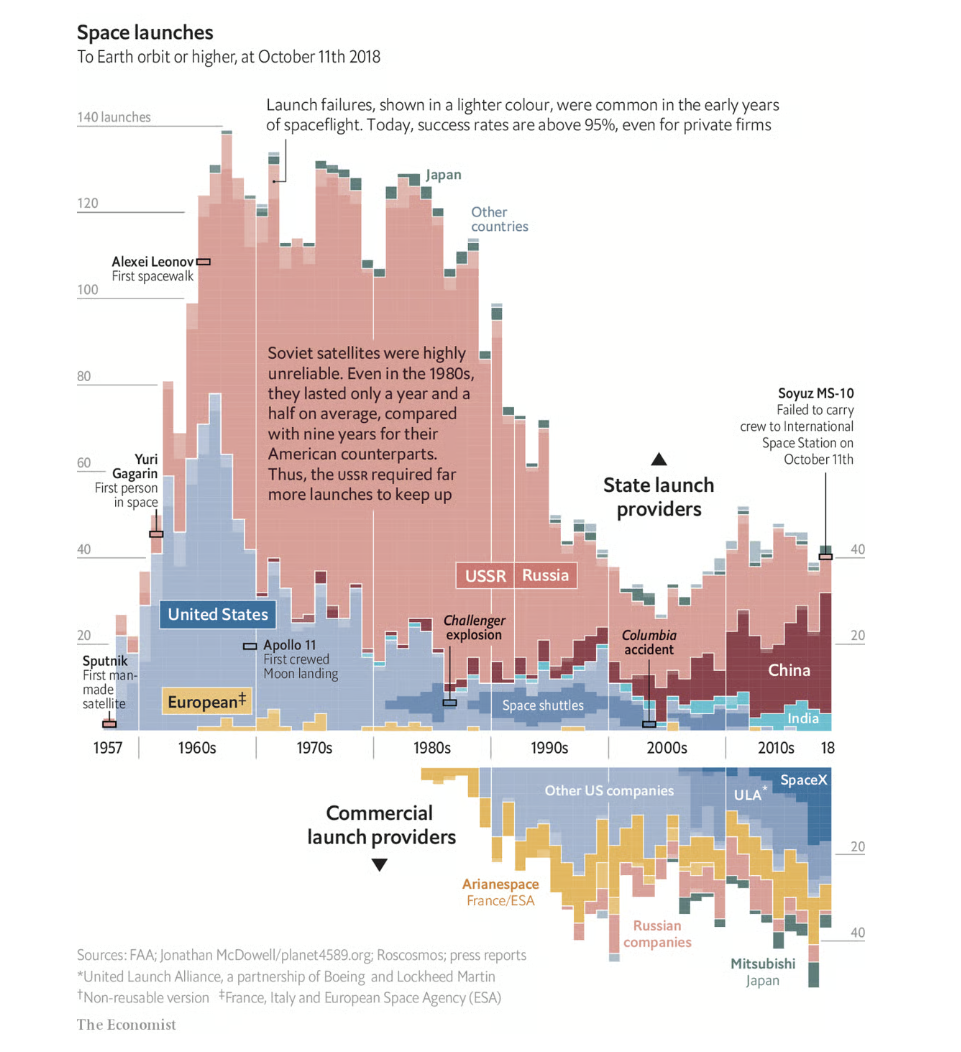}\par} \\
\midrule
\textbf{Ours}~\href{https://data2story.github.io/economist/01_space-launches/blog_opus47_0503_0236/viewer.html}{[link]} &
Title: \emph{The Great Launch Inversion}
\par\smallskip
{\centering\includegraphics[width=0.5\linewidth]{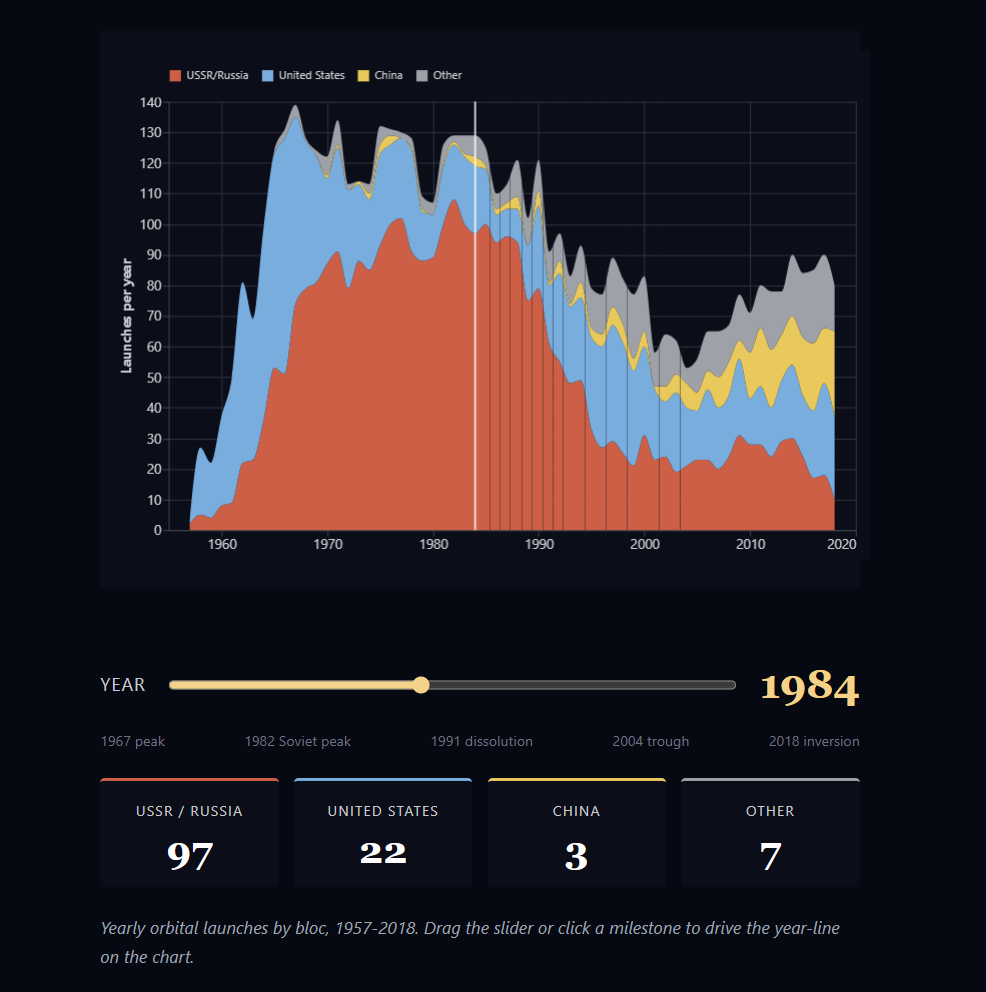}\par} \\
\midrule
\textbf{Analyze} &
In the human-written version, a large amount of information is densely packed into a single image. Key moments are annotated with descriptive text, making the chart richer in content and clearer in explanation. In contrast, the agent's version turns the image into an interactive chart, where users can slide along the year axis to view specific numbers for each year. However, it lacks the descriptive annotations found in the human version, so users can only access the raw figures without the surrounding context.
\\
\bottomrule
\end{tabularx}
\end{table}


\begin{table}[H]
\centering
\small
\renewcommand{\arraystretch}{1.25}
\caption{The Economist: Managers in football matter much less than most fans think}
\label{tab:caseE3-opening}
\begin{tabularx}{\linewidth}{@{}p{1.8cm} X@{}}
\toprule
\textbf{Data} & {Managers in football matter much less than fans think} \\
\midrule
\textbf{Category} & Audio Artifact + Visual Artifact \\
\midrule
\textbf{Human}~\href{https://www.economist.com/graphic-detail/2019/01/19/managers-in-football-matter-much-less-than-most-fans-think}{[link]} &
Title: \emph{Managers in football matter much less than most fans think}
\par\smallskip
{\centering\includegraphics[width=0.6\linewidth]{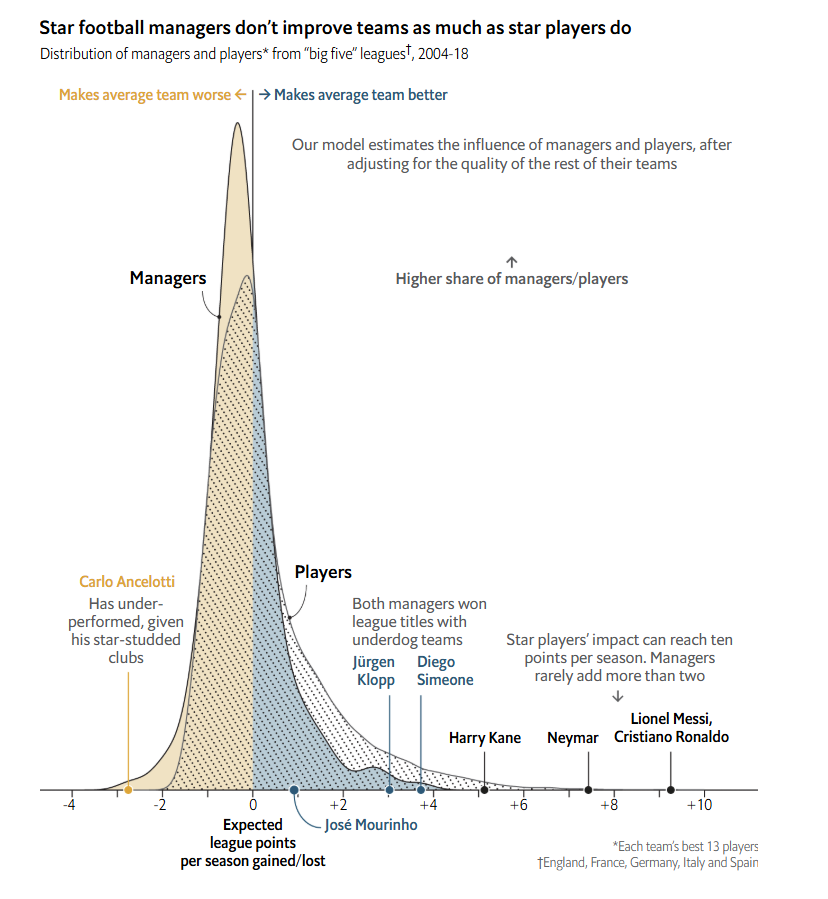}\par} \\
\midrule
\textbf{Ours}~\href{https://data2story.github.io/economist/06_football-managers/blog_opus47_0503_1137/viewer.html}{[link]} &
Title: \emph{Managers in football matter much less than fans think}
\par\smallskip
{\centering\includegraphics[width=0.6\linewidth]{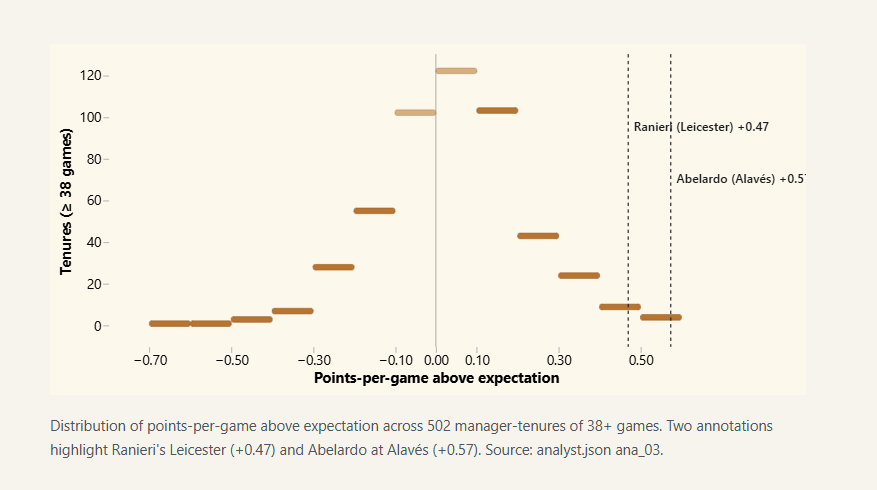}\par} \\
\midrule
\textbf{Analyze} &
In the human version, the visualization is clear and well-structured: it shows each manager's expected score, with prominent annotations marking both top-tier players and elite managers. This design makes the fine-grained information immediately readable at a glance. In contrast, the agent blog's chart does not present a clean curve; it reads more like a set of labels positioned along the y-axis of scores. The agent annotates only a few managers but omits the standout players, and the overall distribution is not visually salient.
\\
\bottomrule
\end{tabularx}
\end{table}

\begin{table}[H]
\centering
\small
\renewcommand{\arraystretch}{1.25}
\caption{The Pudding: The Structure of Stand-Up Comedy}
\label{tab:caseP1-opening}
\begin{tabularx}{\linewidth}{@{}p{1.8cm} X@{}}
\toprule
\textbf{Data} & One Ten-Second Laugh — The Architecture of Ali Wong's Baby Cobra \\
\midrule
\textbf{Category} & Video Artifact + Interactive Artifact \\
\midrule
\textbf{Human}~\href{https://pudding.cool/2018/02/stand-up/}{[link]} &
Title: \emph{The Structure of Stand-Up Comedy}
\par\smallskip
{\centering\includegraphics[width=0.6\linewidth]{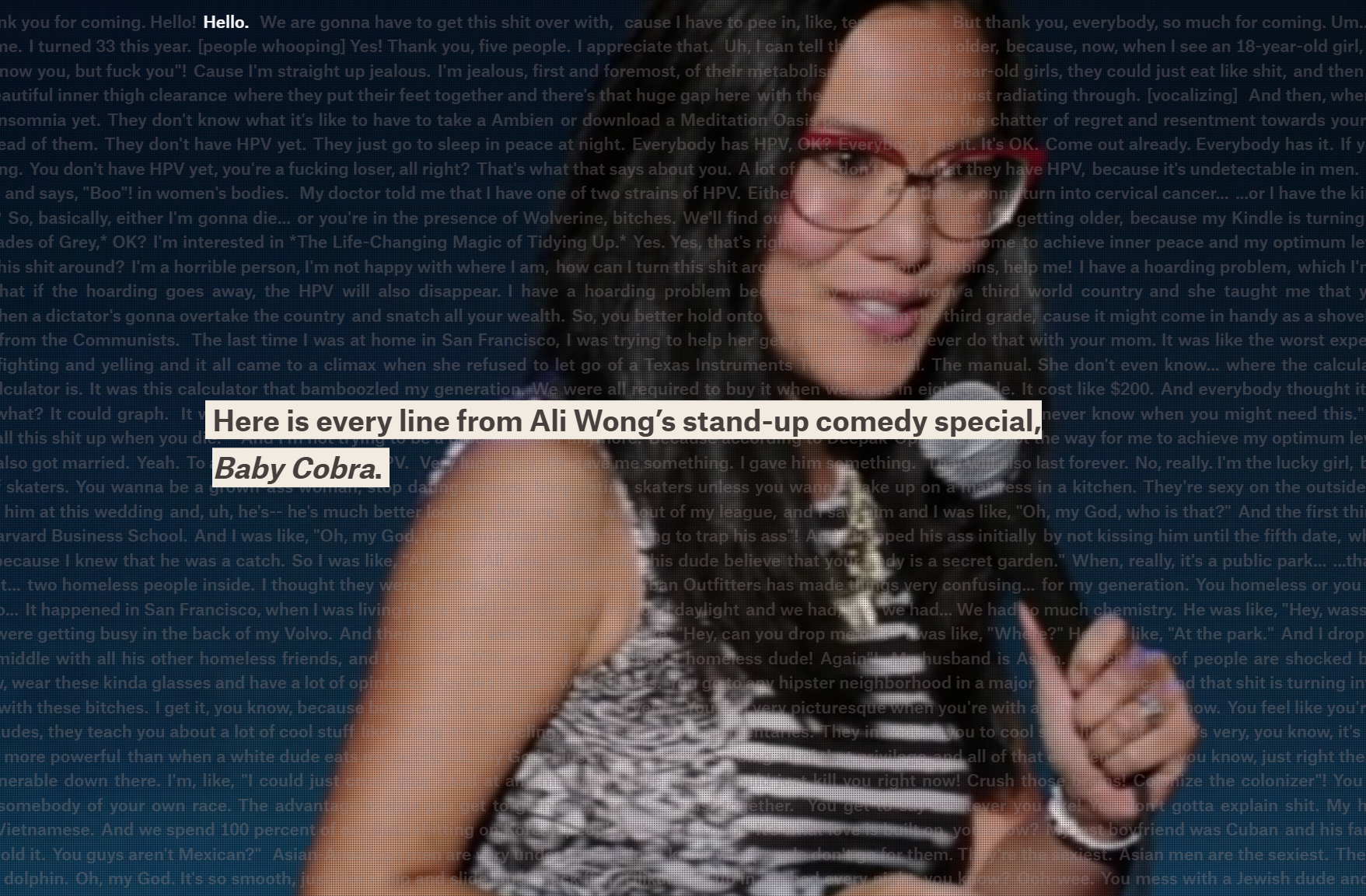}\par} \\
\midrule
\textbf{Ours}~\href{https://data2story.github.io/pudding/02_stand-up/blog_opus47_0503_0049/viewer.html}{[link]} &
Title: \emph{One ten-second laugh, and what holds it up}
\par\smallskip
{\centering\includegraphics[width=0.6\linewidth]{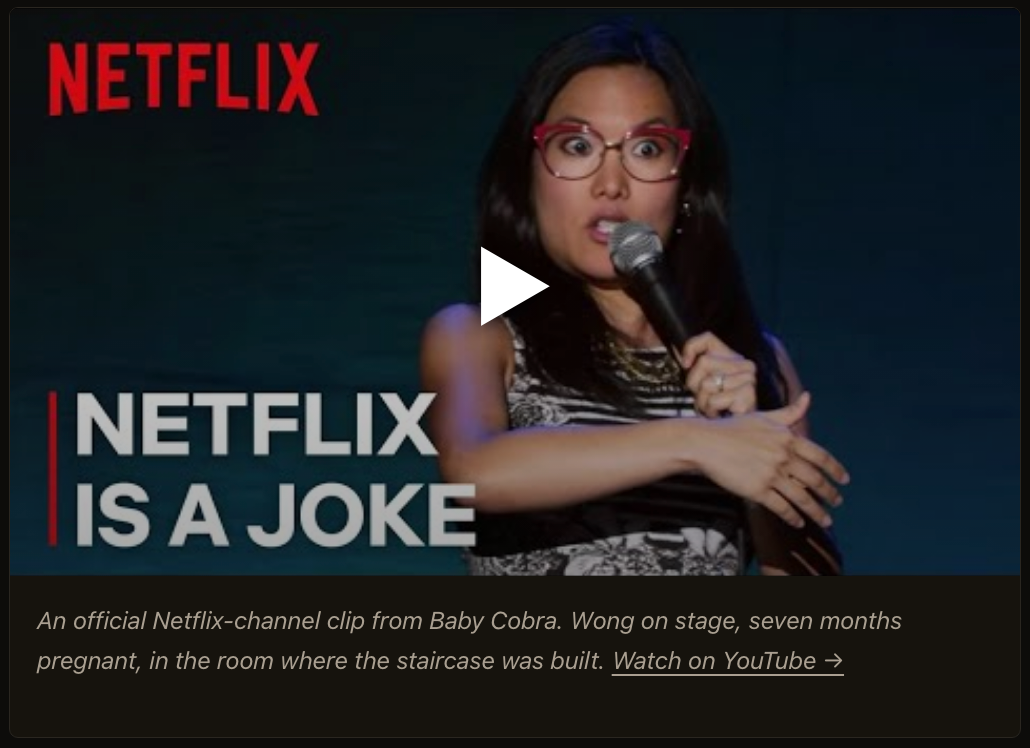}\par} \\
\midrule
\textbf{Analyze} &
In the human-authored version, the video is embedded inline within the dense surrounding text and plays automatically, with a live transcript animating alongside the playback. The effect is polished and attention-holding, reflecting careful design work. The agent-generated version, by contrast, simply embeds a static YouTube iframe that requires the reader to click through and watch the video on YouTube itself.
\\
\bottomrule
\end{tabularx}
\end{table}

\begin{table}[H]
\centering
\small
\renewcommand{\arraystretch}{1.25}
\caption{The Pudding: Internet Boy Band Database}
\label{tab:caseP4-opening}
\begin{tabularx}{\linewidth}{@{}p{1.8cm} X@{}}
\toprule
\textbf{Data} & {The look you remember is one of four — Boy bands, by the data} \\
\midrule
\textbf{Category} & Audio Artifact + Interactive Artifact \\
\midrule
\textbf{Human}~\href{https://pudding.cool/2018/11/boy-bands/}{[link]} &
Title: \emph{Internet Boy Band Database}
\par\smallskip
{\centering\includegraphics[width=0.6\linewidth]{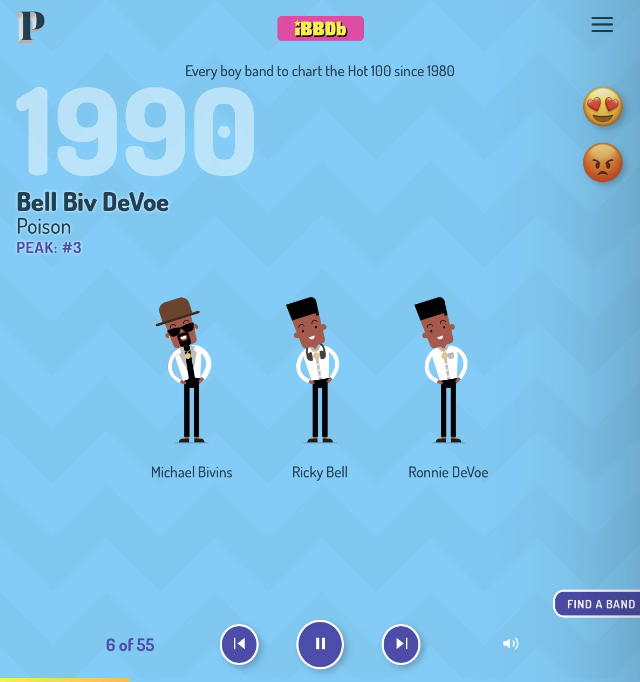}\par} \\
\midrule
\textbf{Ours}~\href{https://data2story.github.io/pudding/12_boybands/blog_opus47_0503_0109/viewer.html}{[link]} &
Title: \emph{The look you remember is one of four}
\par\smallskip
{\centering\includegraphics[width=0.6\linewidth]{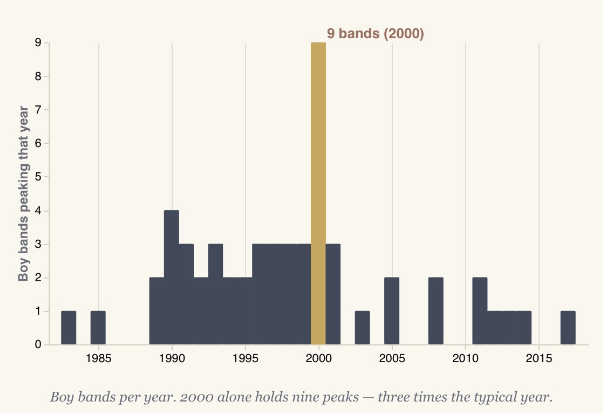}\par} \\
\midrule
\textbf{Analyze} &
The human version is a music-driven immersive page: animated avatars for each band move in sync with autoplaying audio, and switching tracks swaps both the song and the illustrated lineup. It produces strong emotional engagement but only one band is visible at a time. The agent version constructs it as a static yearly histogram, exposing the early-2000s peak and the full distribution at a glance. 
\\
\bottomrule
\end{tabularx}
\end{table}

\begin{table}[H]
\centering
\small
\renewcommand{\arraystretch}{1.25}
\caption{TidyTuesday: Moore's law: The number of transistors per microprocessor}
\label{tab:caseT1-opening}
\begin{tabularx}{\linewidth}{@{}p{1.8cm} X@{}}
\toprule
\textbf{Data} & {The forecast that aged — Moore’s Law on the data} \\
\midrule
\textbf{Category} & Interactive Artifact \\
\midrule
\textbf{Human}~\href{https://ourworldindata.org/grapher/transistors-per-microprocessor}{[link]} &
Title: \emph{Moore's law: The number of transistors per microprocessor}
\par\smallskip
{\centering\includegraphics[width=0.6\linewidth]{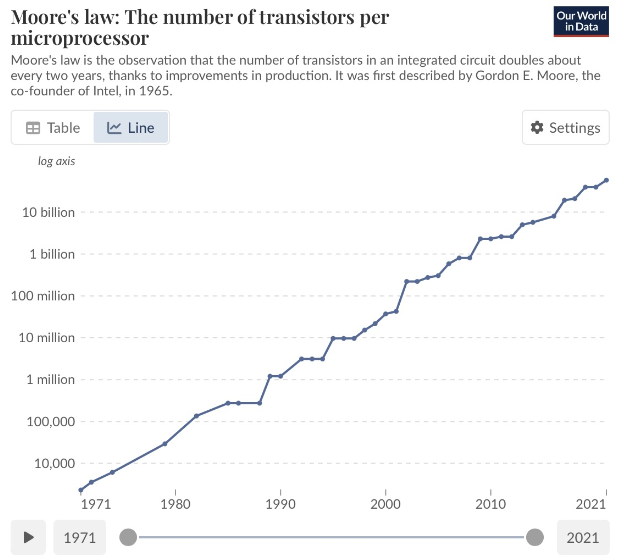}\par} \\
\midrule
\textbf{Ours}~\href{https://data2story.github.io/tidytuesday/03_moores-law/blog_opus47_0503_1218/viewer.html}{[link]} &
Title: \emph{A pencil line, drawn in 1975, that aged.}
\par\smallskip
{\centering\includegraphics[width=0.6\linewidth]{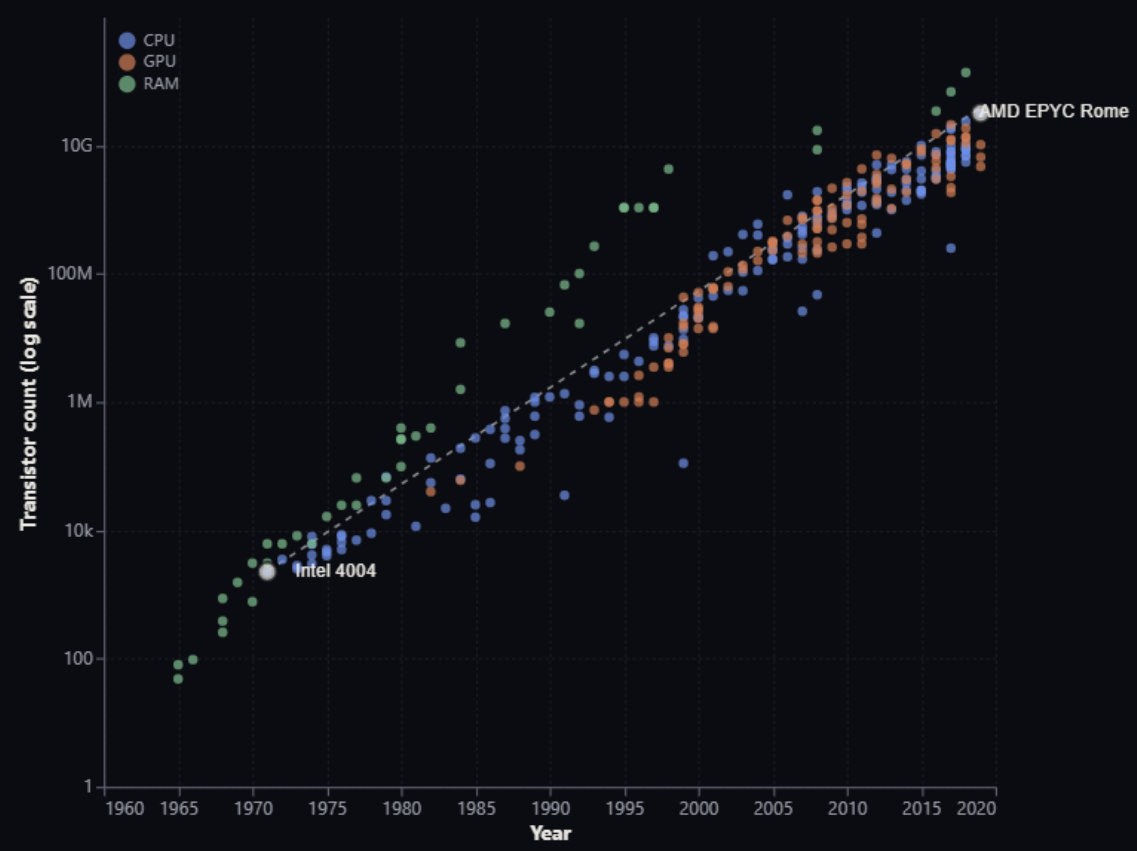}\par} \\
\midrule
\textbf{Analyze} &
The human chart distills Moore's Law to a single log-scale line from 1971 to 2021, framed with explicit prose context and a Table/Line/Settings toggle, optimizing for legibility and citation. The agent version expands the same domain into a three-class scatter (CPU, GPU, RAM) on the same log scale, annotating the Intel 4004 and AMD EPYC Rome endpoints to anchor this massive growth. The agent surfaces between-class structure that the human design intentionally hides, at the cost of denser overplotting and higher reader effort.
\\
\bottomrule
\end{tabularx}
\end{table}

\begin{table}[H]
\centering
\small
\renewcommand{\arraystretch}{1.25}
\caption{TidyTuesday: A Growing Number of `Repair Cafes' Are Popping Up Around the World to Curb Consumer Waste}
\label{tab:caseT4-opening}
\begin{tabularx}{\linewidth}{@{}p{1.8cm} X@{}}
\toprule
\textbf{Data} & {What 178,749 repair attempts say about design} \\
\midrule
\textbf{Category} & Textual Artifact \\
\midrule
\textbf{Human}~\href{https://insideclimatenews.org/news/11112025/todays-climate-repair-cafe-consumer-waste/}{[link]} &
Title: \emph{A Growing Number of `Repair Cafes' Are Popping Up Around the World to Curb Consumer Waste}
\par\smallskip
{\centering\includegraphics[width=0.6\linewidth]{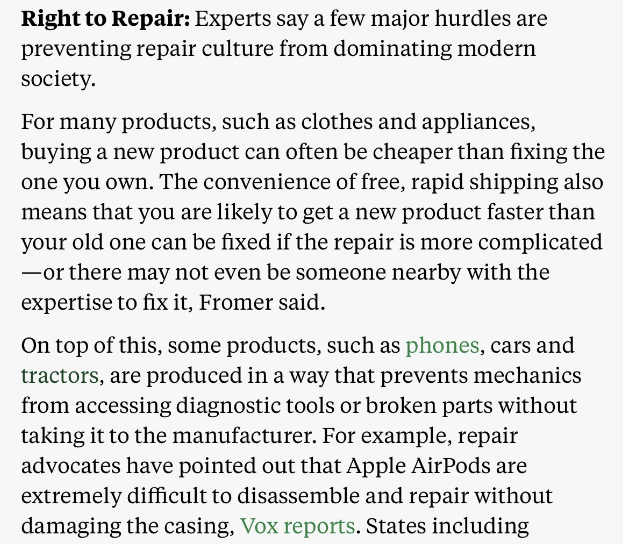}\par} \\
\midrule
\textbf{Ours}~\href{https://data2story.github.io/tidytuesday_2026/14_repair-cafes/blog_opus47_0504_0148/viewer.html}{[link]} &
Title: \emph{What 178,749 Broken Things Tell Us}
\par\smallskip
{\centering\includegraphics[width=0.6\linewidth]{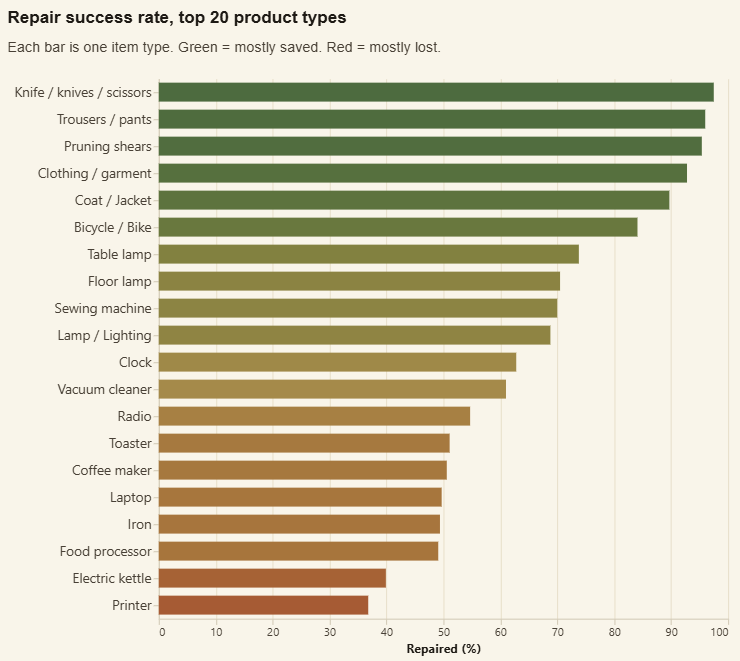}\par} \\
\midrule
\textbf{Analyze} &
The human version is qualitative and explanatory: it embeds repair culture inside a causal story where cheap replacement, fast shipping, scarce local expertise, and manufacturer lock-in act as structural barriers to repair. There is no chart; the prose is the artifact. The agent version produces a ranked bar chart of repair success rates across the top twenty product types, drawn from a large multi-country dataset. A diverging green-to-red palette encodes the outcome directly, turning textiles and hand tools into success cases and printers and electric kettles into failures.
\\
\bottomrule
\end{tabularx}
\end{table}

\section{Agent-as-Judge demonstration}
\label{sec:agent-judge-demo}

We illustrate how a computer-use agent reads a generated article and prepares its rubric judgements. The actions by computer-use agents are highlighted in red.

\begin{table}[!h]
\centering
\small
\renewcommand{\arraystretch}{1.25}
\caption{Agent-as-judge, \emph{Inspector-off} run on \emph{The Space Launches}.
}
\label{tab:vlm-judge-closed}
\begin{tabularx}{\linewidth}{@{}p{3.0cm} X@{}}
\toprule
\textbf{Initial state} &
The judge loads the article and observes the introductory animation, mirroring a human reader's first encounter with the page.
\par\smallskip
{\centering\includegraphics[width=\linewidth]{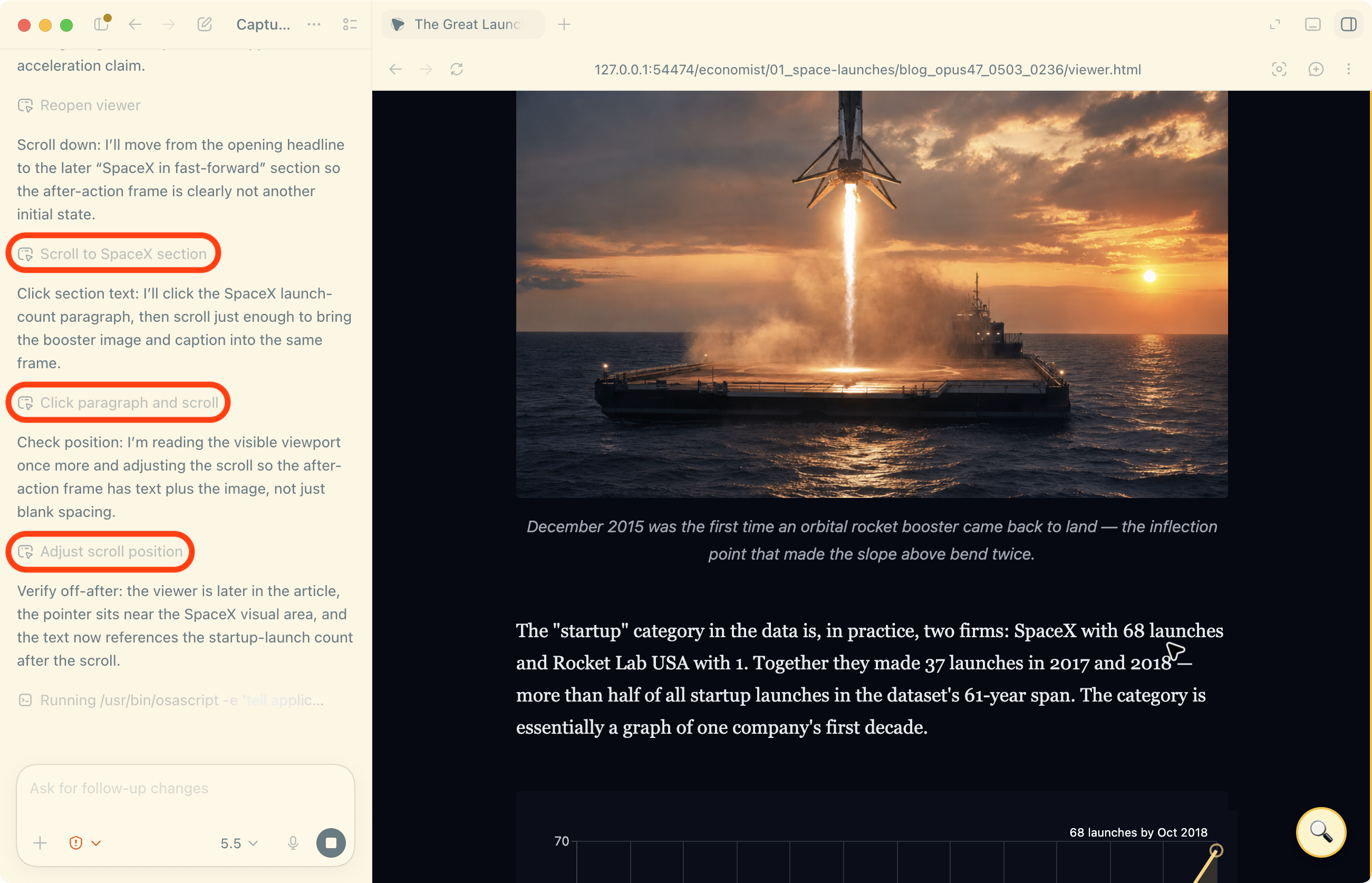}\par} \\
\midrule
\textbf{Reading the article} &
The agent then traverses the body via batched scroll-and-screenshot loops, accumulating a visual record of the prose, charts, and stat callouts in the natural reading order.
\par\smallskip
{\centering\includegraphics[width=\linewidth]{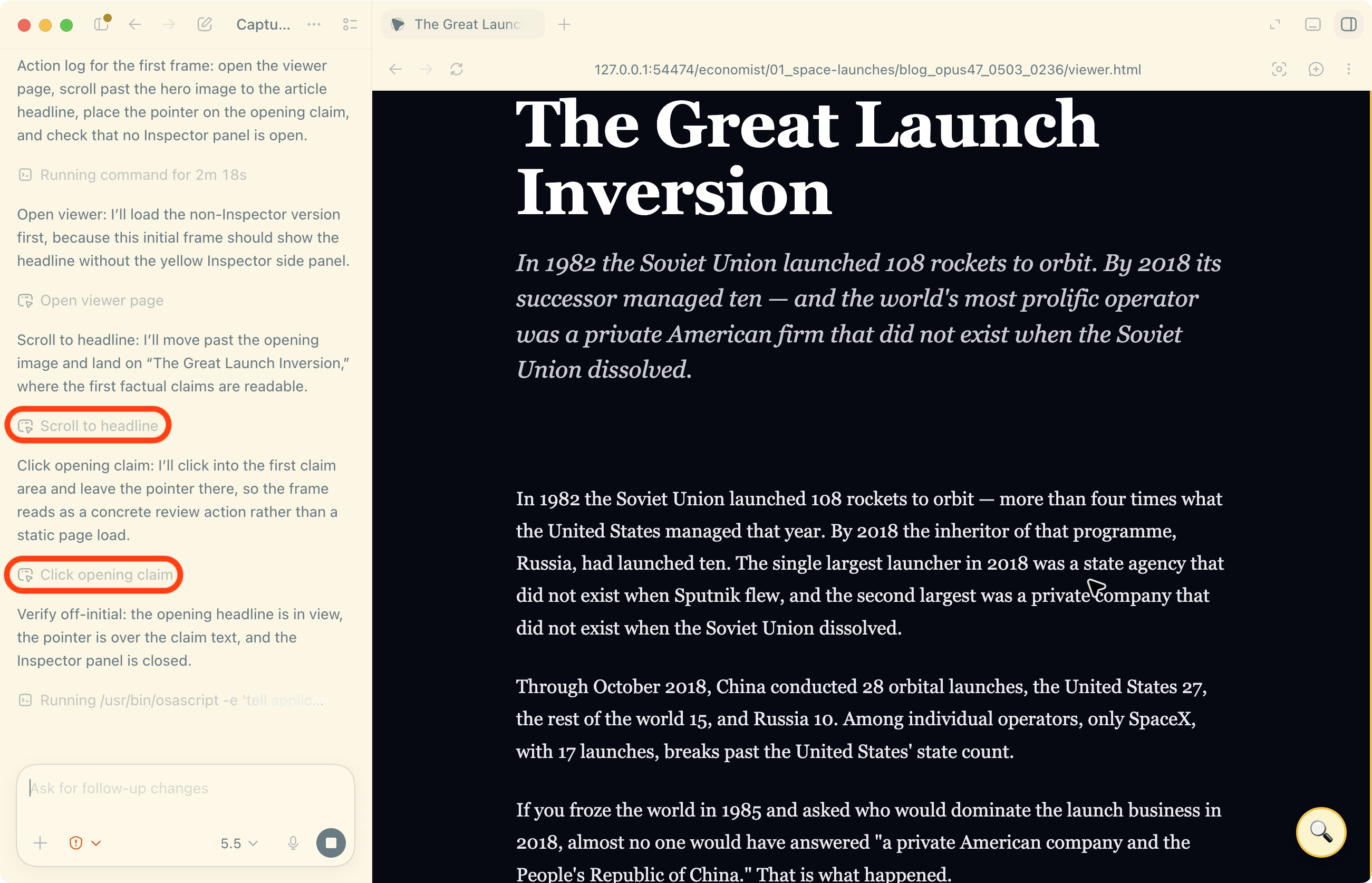}\par} \\
\bottomrule
\end{tabularx}
\end{table}

\begin{table}[!h]
\centering
\small
\renewcommand{\arraystretch}{1.25}
\caption{Agent-as-judge, \emph{Inspector-on} run on \emph{The Space Launches}.
}
\label{tab:vlm-judge-open}
\begin{tabularx}{\linewidth}{@{}p{3.0cm} X@{}}
\toprule
\textbf{Initial state} &
On arrival, the Inspector panel is already open. 
It exposes the article as two structured views: a list of every annotated sentence with its lineage badges, and a list of every named asset (chart, callout, interactive element) the article renders.
\par\smallskip
{\centering\includegraphics[width=\linewidth]{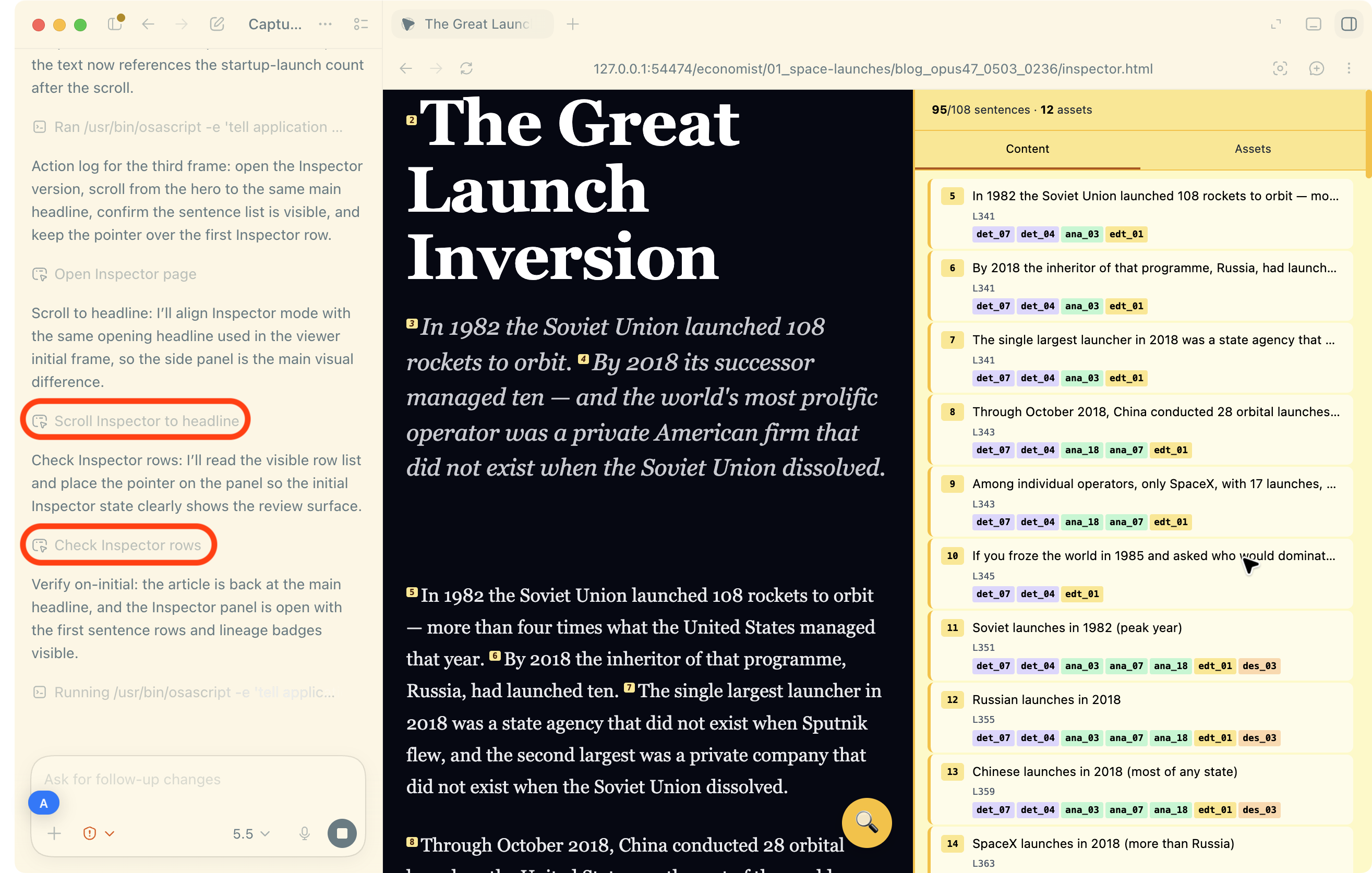}\par} \\
\midrule
\textbf{Reading the article with the Inspector} &
After reading the body, the agent navigates between the Inspector's two views (the action stream shows it locating and clicking the asset tab, then capturing what it reveals) to verify how each rendered claim and each visual asset traces back to its source (code lines, data tables, or external links) before issuing scores.
\par\smallskip
{\centering\includegraphics[width=\linewidth]{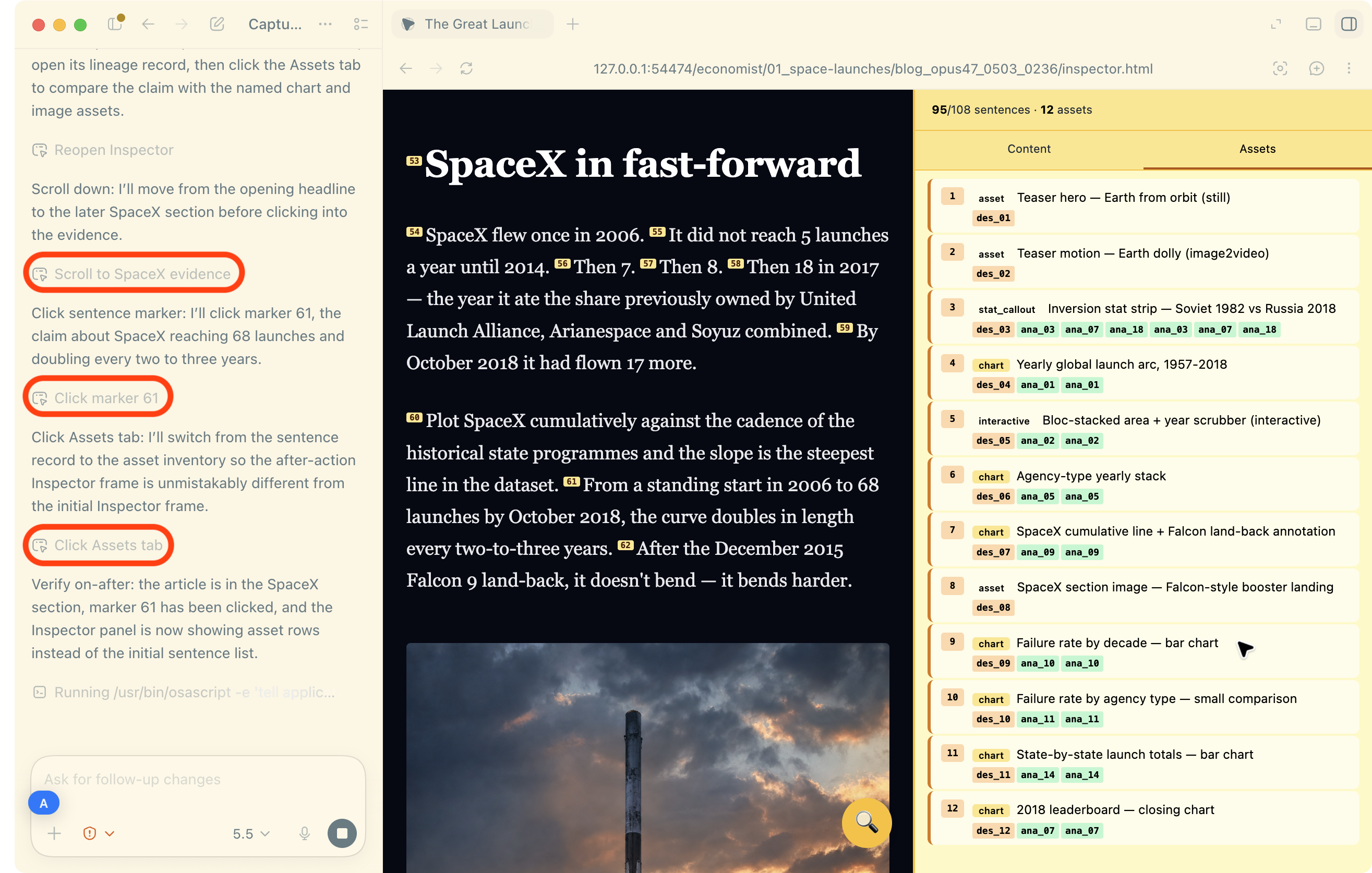}\par} \\
\bottomrule
\end{tabularx}
\end{table}

}

\end{document}